\documentclass{bmvc2k}

\title{Category-Level Pose Retrieval with Contrastive Features Learnt with Occlusion Augmentation}

\usepackage{url}
\usepackage{amsfonts}

\usepackage{multirow}

\addauthor{Georgios Kouros}{georgios.kouros@esat.kuleuven.be}{1}
\addauthor{Shubham Shrivastava}{sshriva5@ford.com}{2}
\addauthor{C\'edric Picron}{cedric.picron@esat.kuleuven.be}{1}
\addauthor{Sushruth Nagesh}{snagesh1@ford.com}{2}
\addauthor{Punarjay Chakravarty}{pchakra5@ford.com}{2}
\addauthor{Tinne Tuytelaars}{tinne.tuytelaars@esat.kuleuven.be}{1}

\addinstitution{
 PSI-VISICS \\
 Department of Electrical Engineering \\
 KU Leuven \\
 Belgium
}
\addinstitution{
 Ford Greenfield Labs\\
 Palo Alto \\
 California, USA
}

\runninghead{KOUROS ET AL.}{CATEGORY-LEVEL POSE RETRIEVAL WITH CONTRASTIVE FEATURES}

\def\etal{\emph{et al}\bmvaOneDot}
\def\@fnsymbol#1{\ensuremath{\ifcase#1\or *\or \dagger\or \ddagger\or
   \mathsection\or \mathparagraph\or \|\or **\or \dagger\dagger
   \or \ddagger\ddagger \else\@ctrerr\fi}}
\newcommand{\ssymbol}[1]{^{\@fnsymbol{#1}}}
\begin{document}

\maketitle

\begin{abstract}
Pose estimation is usually tackled as either a bin classification or a regression problem. In both cases, the idea is to directly predict the pose of an object. This is a non-trivial task due to appearance variations between similar poses and similarities between dissimilar poses. Instead, we follow the key idea that comparing two poses is easier than directly predicting one. Render-and-compare approaches have been employed to that end, however, they tend to be unstable, computationally expensive, and slow for real-time applications. We propose doing category-level pose estimation by learning an alignment metric in an embedding space using a contrastive loss with a dynamic margin and a continuous pose-label space. For efficient inference, we use a simple real-time image retrieval scheme with a pre-rendered and pre-embedded reference set of renderings. To achieve robustness to real-world conditions, we employ synthetic occlusions, bounding box perturbations, and appearance augmentations. Our approach achieves state-of-the-art performance on PASCAL3D and OccludedPASCAL3D and surpasses the competing methods on KITTI3D in a cross-dataset evaluation setting. The code is currently available at \url{https://github.com/gkouros/contrastive-pose-retrieval}.
\end{abstract}

\section{Introduction} \label{sec:introduction}

\begin{figure*}[t]
    \centering
    \includegraphics[width=\linewidth]{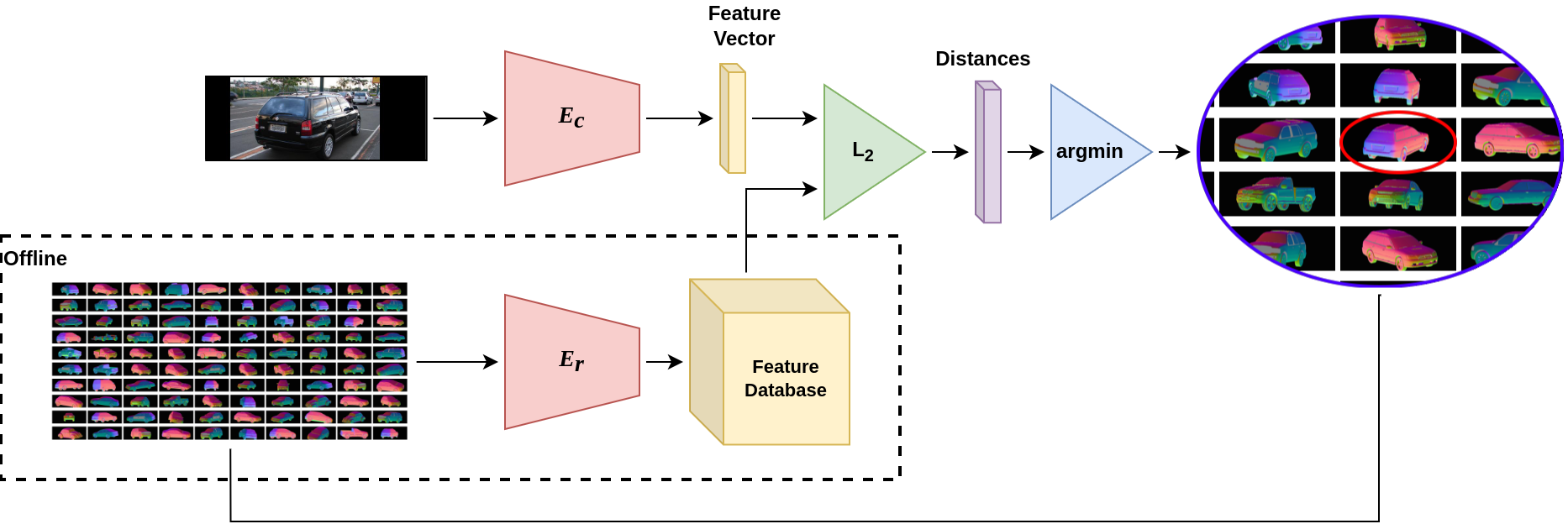}
    \caption{Using a properly learned metric, all  it takes to estimate the pose of an object  with  state-of-the-art accuracy is a simple retrieval scheme that finds the most similar encoded rendering in a database. Two ResNet-50 encoders $E_c$ and $E_r$ are jointly trained in a contrastive manner to learn the mapping of query camera images and reference renderings to a feature space where their feature distance is proportional to their geodesic/pose distance. To ensure fast online inference after training, the reference set is encoded offline.}
    \label{fig:introduction:teaser-image}
\end{figure*}


Estimating the pose of a 3D rigid object is a fundamental task in numerous computer vision applications. For instance, a self-driving vehicle must be able to estimate the pose of other road users in its surroundings in order to navigate safely without endangering itself or others. Deep learning has revolutionized such pose estimation tasks especially for challenging monocular settings \cite{he2019mono3d,beker2020monocular,wang2020NeMo,zhou2018starmap,iwase2021repose,zakharov2019dpod,grabner2020correspondence} compared to stereo \cite{wang2020directshape,chen20153dop,div2020wstereo,chen2020dsgn,li2019stereorcnn} or RGB-D \cite{Tian2020Robust6O,Saadi2021OptimizingRF} settings that leverage 3D information from their inputs. Two ongoing problems in the monocular setting are how to best extract 3D information from 2D image inputs and how to achieve real-time operation and robustness even in complex, cluttered and occluded real-world scenes.

Previous methods approach pose estimation as either a classification \cite{tulsiani2015poseinduction}, regression \cite{xiang2017posecnn,zaw2018disentangled,xiao2019posefromshape,su2015renderforcnn} or optimization problem \cite{beker2020monocular,wang2020NeMo,iwase2021repose}. Classification and regression have to directly predict the pose as either belonging to a bin or as a set of continuous values. On the other hand, optimization methods such as render-and-compare approaches iteratively optimize the pose. Comparing two images with regard to their pose can be considered a much easier task to learn. Nevertheless, such an iterative optimization approach, although accurate, may be too slow for real-time category-level pose estimation.

In this work, we propose a multimodal contrastive learning framework for extracting discriminative features from real-world images and renderings that can be used for comparing the two images with regard to their poses. Poses in this work refer to the 3D orientation of the camera with respect to an object expressed with the azimuth, elevation, and in-plane rotation angles. Rather than following a slow iterative approach similar to render-and-compare methods \cite{wang2020NeMo,beker2020monocular}, we utilize a common nearest neighbour retrieval scheme that compares the feature embedding of a query image with a reference set of embeddings from rendered objects in various poses. We also increase robustness to complex and cluttered scenes by augmenting training images with appearance variations, bounding box perturbations, and synthetic occlusions. Our main contributions can be summarized as follows:

\begin{itemize}
    \itemsep0em 
    \item We propose a simple yet effective category-level pose retrieval framework based on learning discriminative features using contrastive learning with a dynamic margin.
    \item We show that strong data augmentation can enhance a simple pose estimation architecture to outperform more complex ones.
    \item We report state-of-the-art results on PASCAL3D and OccludedPASCAL3D, as well as superior cross-dataset performance on KITTI3D against the evaluated competing methods.
\end{itemize}

\section{Related Work} \label{ch:relatedwork}

Monocular pose estimation can be described as an ill-posed problem due to the lack of 3D information despite the fact that good empirical results have been obtained. Recovering 3D information is usually accomplished through either monocular depth prediction \cite{beker2020monocular,he2019mono3d,kong2020unsupervised} or by incorporating prior hypotheses about the objects such as shape priors for template matching \cite{wang2020NeMo,iwase2021repose,zhakarov2017dynamic,zhou2018starmap} in render-and-compare or image retrieval settings. In this work, we utilize prior shape hypotheses in the form of CAD models for category-level pose estimation. We specifically target category-level methods to achieve a good trade-off between accuracy, generalization, and robustness to occlusions and clutter compared to instance-based and category-agnostic methods. 

In the scope of render-and-compare approaches, RePose \cite{iwase2021repose} runs faster than real-time by optimizing the pose of an object with learned deep textures, but is applicable only at the instance-level. Beker \etal \cite{beker2020monocular} and Wang \etal \cite{wang2020directshape} propose render-and-compare methods for estimating the pose and shape of cars using photometric, depth, or silhouette comparison, but without achieving real-time performance. NeMo \cite{wang2020NeMo}, on the other hand, uses a generative neural mesh model and contrastive learning to first learn discriminative features that distinguish objects from occlusions and background clutter before optimizing the pose through a render-and-compare scheme for approximately 8 seconds per object. To avoid the overhead of render-and-compare optimization, we choose a simple yet efficient image retrieval setting.

Retrieval-based methods rely on a good comparison metric and deep metric learning with contrastive \cite{chopra2005contrastive} or triplet-like \cite{schroff2015facenet} losses has been instrumental towards that end. Wohlhart and Lepetit \cite{wohlhart2015learningdescriptors} first proposed to use a triplet-like loss to optimize a CNN feature extractor for instance-level pose estimation tasks based on nearest neighbour retrieval. Zhakarov \etal \cite{zhakarov2017dynamic} augmented the triplet-like loss with a dynamic margin that considers both the object instance and the pose distance between anchor/positive and negative samples. All aforementioned methods, however, discretize the pose space into bins with a negative impact on accuracy. Balntas \etal \cite{Balntas2017PoseGR} incorporated a regression loss term during training that further improved performance. Papaioannidis and Pitas \cite{papaioannidis2020quaternionlearning} added a regression loss term as well that enabled direct pose regression instead of slow nearest neighbour search. PoseContrast \cite{xiao2021posecontrast} is trained with a loss function, combining classification, regression, and contrastive loss terms, on real data with pose-aware augmentations and without prior geometry knowledge. In this work, we propose a simple contrastive loss with a dynamic margin and a continuous pose space achieved by choosing positive and negative pairs and optimizing according to pose distance rather than binning. In our case, each pair consists of a real image (anchor) and a rendering of a CAD model (positive/negative), for which we jointly train two individual feature extractor CNNs.

Achieving robustness to foreground occlusions and background clutter usually requires complex architectures and loss functions. For instance, NeMo \cite{wang2020NeMo} employs contrastive learning to learn how to distinguish between object features and background clutter or foreground occlusions. In a different approach, Sarandi \etal \cite{sarandi2018synthetic} propose to augment input images with synthetic occlusions to increase robustness in human pose estimation tasks. In this work, we follow the same approach and incorporate a synthetic occlusion augmentation scheme for the pose estimation of 3D rigid objects rather than humans.
\section{Method} \label{sec:method}


\subsection{Learning Discriminative Pose Features}
The main idea in deep metric learning is to optimize a high-dimensional embedding feature space or manifold so that samples are pulled together or pushed apart depending on whether they belong to the same class or not. This task was first accomplished using the Contrastive Loss \cite{chopra2005contrastive} which is defined as
\begin{equation}\label{eq:contrastive-loss}
\mathcal{L} = \frac{1}{2N} \sum_{i=1}^N
\Big[
(1-y_i)~||f_{1,i} - f_{2,i}||^2_2 
+ y_i~max(0,~m - ||f_{1,i} - f_{2,i}||^2_2)
\Big]~,
\end{equation}
where $N$ is the batch size or number of sample pairs, $m$ is the margin, $f$ is the embedding/encoding function, and $y_i$ is a label that is $1$ if the pair is positive and $0$ if negative.

Applying deep metric learning for pose estimation requires discretizing the pose space and assigning the pose labels to bins as in \cite{wohlhart2015learningdescriptors,zhakarov2017dynamic}. However, this means that slightly different poses might fall in different bins and thus the network would be encouraged to separate them in feature space, which would negatively impact generalization to unseen poses. Consequently, similar to the Triplet-like Dynamic Margin Loss \cite{zhakarov2017dynamic}, we employ a dynamic margin that is proportional to the geodesic pose distance between two samples. In contrast to \cite{zhakarov2017dynamic}, we train our models for pose estimation on the category-level rather than the instance-level, and avoid the discretization of the pose-label space that negatively impacts accuracy. While they use the discretized pose labels to determine positive and negative sample pairs, we propose a continuous pose-label space and determine positive and negative pairs by applying a threshold on the pose distance. As a result, we redefine the Contrastive Loss from Equation~\ref{eq:contrastive-loss} to our \textit{Contrastive Pose Loss} expressed as
\begin{equation}\label{eq:contrastive-pose-loss}
\mathcal{L} = \frac{1}{2N} \sum_{i=1}^N
\Big[
(1-y_i)~max(0,~||f^c_{1,i} - f^r_{2,i}||^2_2 - m \Delta\theta)
+ y_i~max(0,~m \Delta\theta  - ||f^c_{1,i} - f^r_{2,i}||^2_2)
\Big]~,
\end{equation}
where $\Delta\theta = 2~cos^{-1}(|q_i \cdot q_j|)$ denotes the geodesic distance between two poses expressed as quaternions $q_i, q_j$. Furthermore, $f^c$ and $f^r$ denote the embedding functions of encoders $E_c$ and $E_r$, respectively.




\subsection{Sampling and Mining}
According to numerous works \cite{schroff2015facenet,wu2017sampling,hermans2017defense,harwood2017smart,wang2019multi,hong2020hard,robinson2021contrastive}, Deep Metric Learning performance is heavily influenced by the selection of samples in a mini-batch during training and thus a sophisticated scheme is essential to speeding up convergence. 

Datasets often suffer from imbalance, such as typical car datasets having more sedan cars than vans, which may result in poorer performance in the latter subcategory. This can be alleviated by a sampling scheme that weights each sample inversely proportional to the number of occurrences of its subcategory, an idea inspired by the Focal Loss \cite{Lin2020FocalLF}. In addition, for every sample already chosen for a batch, the sampling scheme aims to include $N \geq 1$ additional samples with a similar pose (e.g. less than $5^\circ$ difference) because they are more likely to have a small feature distance that needs to be optimized.

We also designed a pose-aware miner that looks for pairs of samples violating the pose margin, resulting in higher losses and thus more efficient optimization. The sampler feeds the miner a mini-batch of $N$ indices corresponding to $N$ samples composed of a camera image and its rendered counterpart. All possible positive and negative pairs are constructed based on a threshold (e.g. $5^\circ$), of which all pairs that violate the pose margin are used to calculate the loss. The rest are dropped since they do not offer any value to the optimization.

\subsection{Rendering}
\begin{figure}[t]
\centering
\begin{subfigure}{.2\textwidth}
\includegraphics[width=\linewidth]{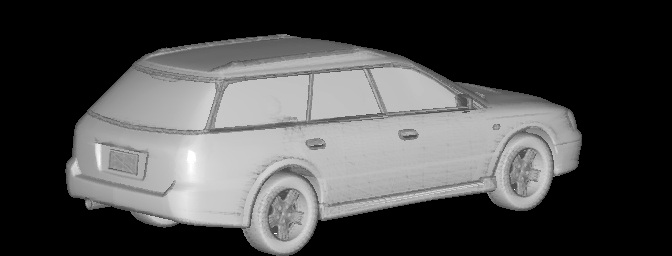}
\caption{RGB}
\label{fig:renderings:rgb}
\end{subfigure}%
\begin{subfigure}{.2\textwidth}
\includegraphics[width=\linewidth]{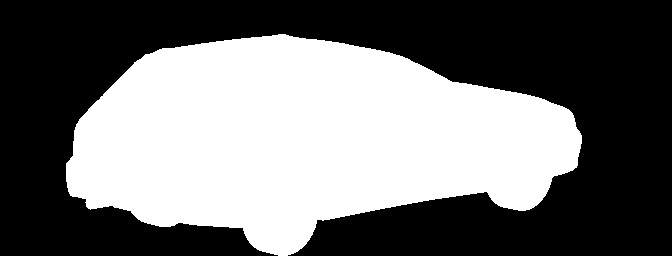}
\caption{Silhouette}
\label{fig:renderings:silhouette}
\end{subfigure}%
\begin{subfigure}{.2\textwidth}
\includegraphics[width=\linewidth]{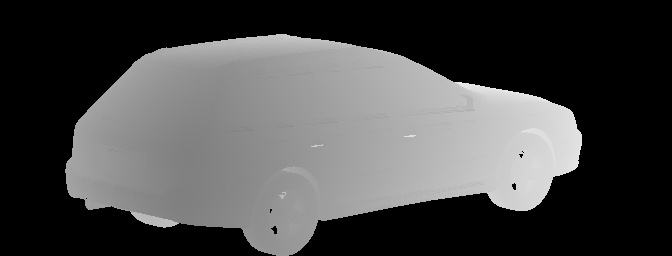}
\caption{Depth Map}
\label{fig:renderings:depth}
\end{subfigure}%
\begin{subfigure}{.2\textwidth}
\includegraphics[width=\linewidth]{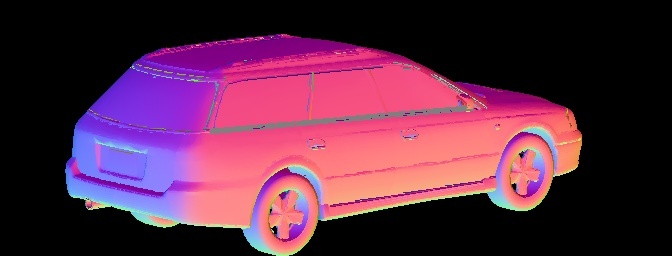}
\caption{Normals Map}
\label{fig:renderings:normals}
\end{subfigure}%
\begin{subfigure}{.2\textwidth}
\includegraphics[width=\linewidth]{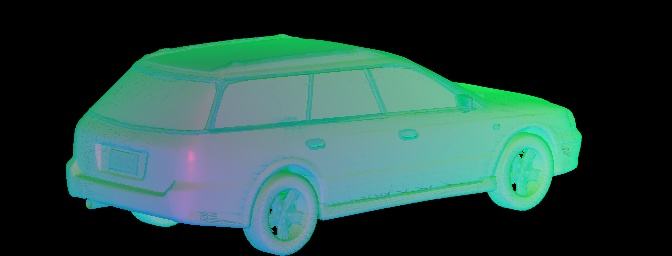}
\caption{Triplet Map}
\label{fig:renderings:combined}
\end{subfigure}%
\caption{Five types of renderings that were evaluated for pose estimation.}
\label{fig:renderings}
\end{figure}

In contrast to NeMo \cite{wang2020NeMo} and similar to \cite{zhakarov2017dynamic}, we do not use a 3D generative model for producing the renderings, but rather employ a more conventional approach of generating 2D mesh renderings, silhouettes, surface normal maps, depth maps, or even multi-channel RGB-depth-normal combinations that were inspired from \cite{grabner2020correspondence} and which we call triplets. It is our intuition that using such feature representations, as illustrated in Figure~\ref{fig:renderings}, preserves perspective information vital to pose estimation tasks as opposed to the 3D generative model used by NeMo. 
Rather than generating a set of renderings for the entire viewing sphere, we create a rendering database by generating one rendering per sample in the training set thus ensuring at least one positive per sample. This approach requires less space and less time for inference while avoiding the need to find a trade-off between discretization error, the size of the database, and inference time. Moreover, this naturally reflects the prior distribution over the viewing sphere.

\subsection{Robustness to Occlusions}
\begin{figure}[t]
\centering
\begin{subfigure}{.25\textwidth}
\includegraphics[width=\linewidth]{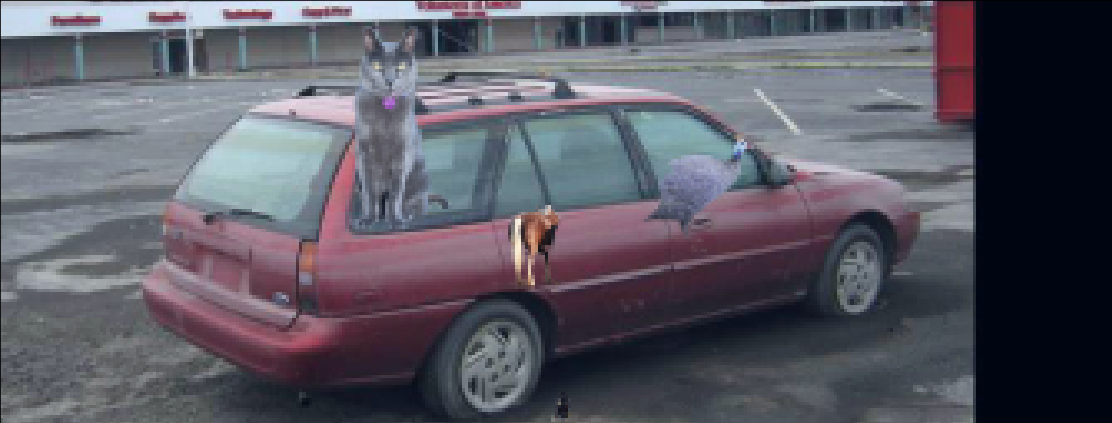}
\caption{$s_{occ}=0.25$}
\label{fig:methodology:occlusions:0.25}
\end{subfigure}%
\begin{subfigure}{.25\textwidth}
\includegraphics[width=\linewidth]{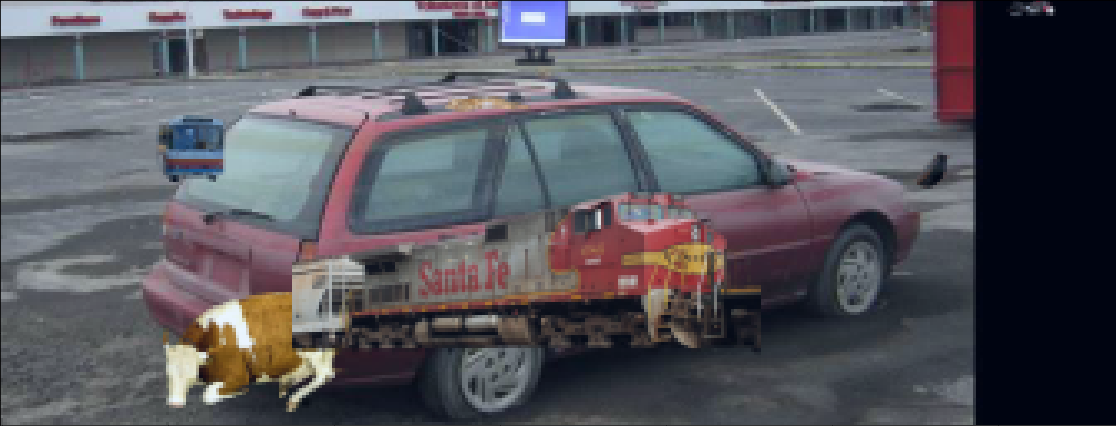}
\caption{$s_{occ}=0.5$}
\label{fig:methodology:occlusions:0.5}
\end{subfigure}%
\begin{subfigure}{.25\textwidth}
\includegraphics[width=\linewidth]{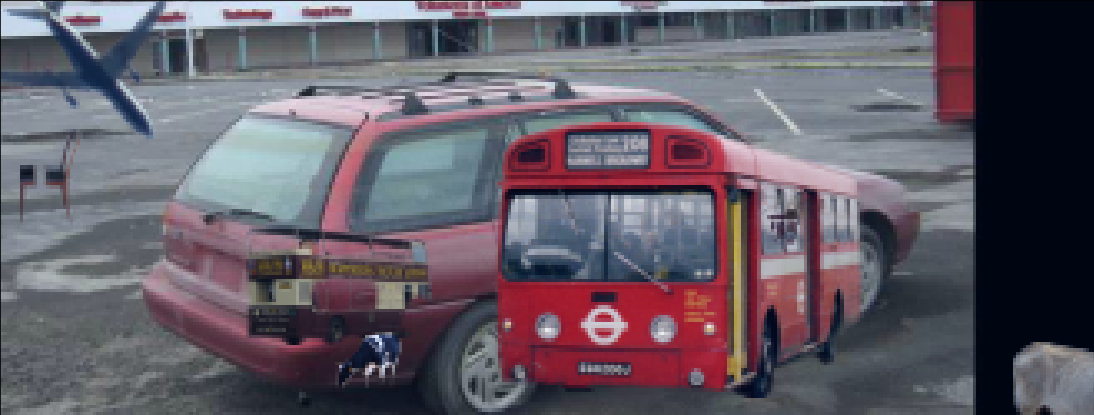}
\caption{$s_{occ}=0.75$}
\label{fig:methodology:occlusions:0.75}
\end{subfigure}%
\begin{subfigure}{.25\textwidth}
\includegraphics[width=\linewidth]{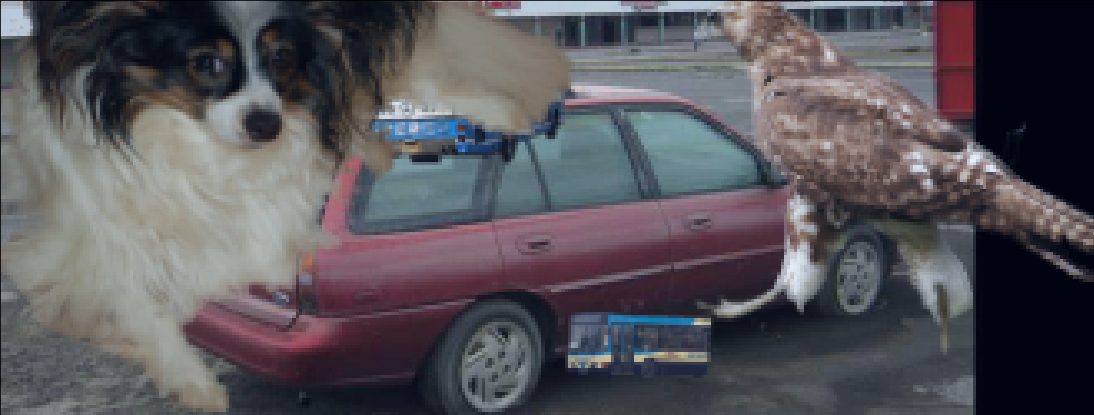}
\caption{$s_{occ}=1.0$}
\label{fig:methodology:occlusions:1.0}
\end{subfigure}%
\caption{Examples of synthetic occlusions for various scale factors. For examples of real occlusions we refer the reader to Figure \ref{appendix:data-overview}.}
\label{fig:methodology:occlusions}
\end{figure}

To increase robustness to occlusions and background clutter, we use data augmentation with synthetic occlusions \cite{sarandi2018synthetic} produced from PASCAL VOC 2012 \cite{pascal-voc-2012}. This involves segmenting objects from PASCAL VOC to create a template set of occluders from 20 object categories. An input image is then augmented with one to eight randomly selected occluders which are visually, spatially, and geometrically augmented. When training with a specific object class we naturally filter out that class from the occluder set to avoid having objects from that class occluding the actual object of interest. Finally, for tuning purposes we use a tunable occlusion scale $s_{occ}$ that is multiplied with a random resize factor $x\sim U[0, 1]$ to produce the resize factor for a random occluder
\begin{equation}
    f_x = f_y =  s_{occ}~x,
\end{equation}
where $f_x$ and $f_y$ are the resizing factors for the horizontal and vertical dimensions, respectively. Figure~\ref{fig:methodology:occlusions} illustrates occlusion augmentations and the effect of $s_{occ}$.

\subsection{Robustness to Bounding Box Noise}
\begin{figure}[t]
\centering
\begin{subfigure}{0.2\textwidth}
\includegraphics[width=\linewidth]{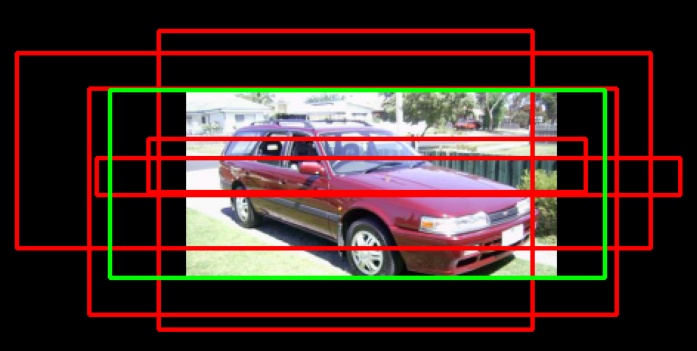}
\caption{$IoU \geq 0.0$}
\end{subfigure}%
\begin{subfigure}{0.2\textwidth}
\includegraphics[width=\linewidth]{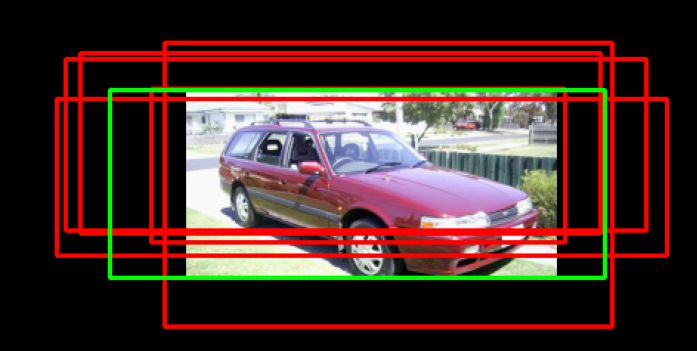}
\caption{$IoU \geq 0.25$}
\end{subfigure}%
\begin{subfigure}{0.2\textwidth}
\includegraphics[width=\linewidth]{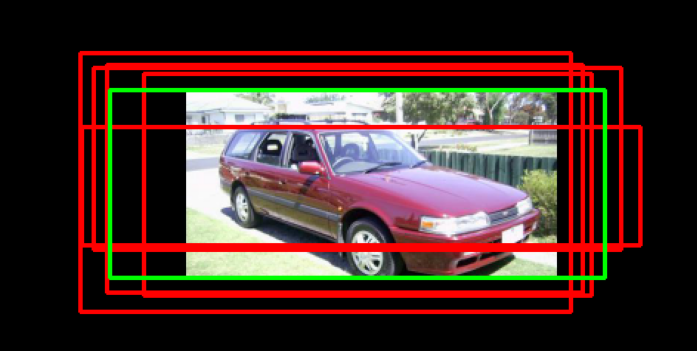}
\caption{$IoU \geq 0.5$}
\end{subfigure}%
\begin{subfigure}{0.2\textwidth}
\includegraphics[width=\linewidth]{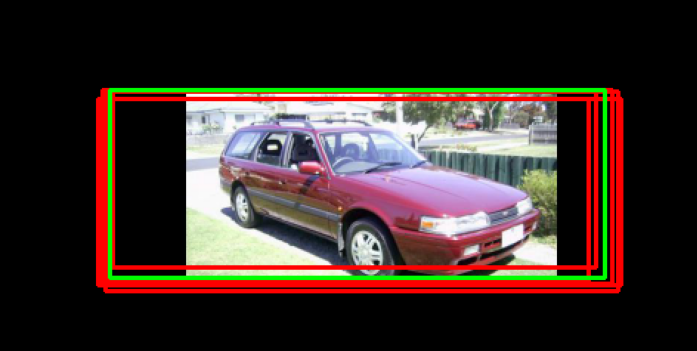}
\caption{$IoU \geq 0.75$}
\end{subfigure}%
\begin{subfigure}{0.2\textwidth}
\includegraphics[width=\linewidth]{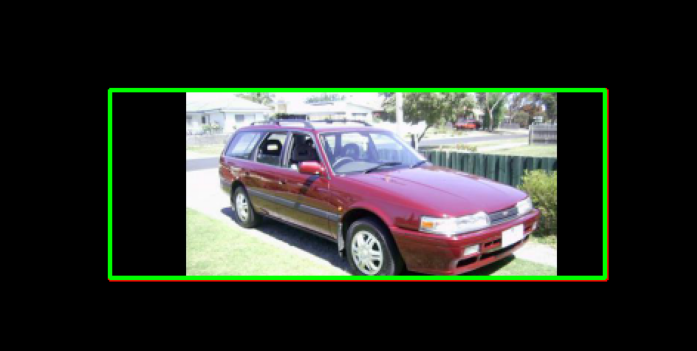}
\caption{$IoU = 1.0$}
\end{subfigure}%
\caption{Artificial bounding box noise with a lower boundary on IoU. Green denotes the original image borders and red denotes randomly perturbed bounding boxes.}
\label{fig:methodology:bbox-noise}
\end{figure}

Throughout the experiments we assume known scale and center of the objects similar to NeMo \cite{wang2020NeMo}. Practically, this is not realistic and although NeMo implies some basic tolerance to center/scale perturbations, it is designed in a way that works optimally with adequate alignment between camera images and renderings. In order to avoid this restriction we propose augmenting training samples with random bounding box noise with a lower boundary on IoU. To define this type of noise we express the deviation of the bounding box corners as a function of the lower IoU boundary. 
If $w$ and $h$ are the width and height of the bounding box and $n$ is the maximum horizontal and vertical corner deviation in pixels, then
\begin{equation}
IoU_{min} = \frac{(w - 2 n) (h - 2 n)}{w h}~.
\end{equation}
By solving the quadratic equation we can calculate the maximum pixel deviation $n$ as a function of $IoU_{min}$ via the equation
\begin{equation}
n = \frac{h+w-\sqrt{(h+w)^2 - 4wh\beta}}{4}~,
\end{equation}
where $\beta=1-IoU_{min}$ is the noise scale parameter used in the experiments. Figure~\ref{fig:methodology:bbox-noise} presents a few examples of our bounding box noise scheme for different IoU lower boundaries.

\subsection{Inference via Pose Retrieval}
After jointly training the encoders $E_c$ and $E_r$, we can predict poses through a simple image retrieval scheme, as shown in Figure \ref{fig:introduction:teaser-image}. Our inference framework requires an offline step of generating a reference set of renderings which need to be embedded using encoder $E_r$ and stored for online inference. Inferring the pose is then basically a two-step-approach composed of encoding a query image with encoder $E_c$, calculating the $L_2$ distance of the query embedding to all feature embeddings in the stored reference set, and finally finding the nearest neighbour, whose label corresponds with our predicted pose.

There are two main limitations with this approach. First, inference requires discretization to a set of sampled orientations, which can be either the orientations that are present in the training set or the orientations in a generated reference set that introduces a tradeoff between discretization and inference speed. In the end, we used the first approach to ensure fast training and inference speed. Second, comparing an encoded query image against a reference set introduces a delay not present in classification or regression approaches. Training a regression layer on top of the $E_c$ encoder, similar to \cite{xiao2021posecontrast,Balntas2017PoseGR} would potentially eliminate both issues.
\section{Experiments} \label{sec}

\subsection{Experimental Setup} \label{ssec:experiments:setup}

Our framework is developed using \textit{PyTorch} \cite{NEURIPS2019_9015}, \textit{PyTorch Metric Learning} \cite{musgrave2020pytorch}, and we also use \textit{PyTorch3D} \cite{ravi2020pytorch3d} for the generation of the renderings. We jointly train two ResNet50 \cite{he2015deep} encoders $E_c$ and $E_r$ as shown in Figure~\ref{fig:introduction:teaser-image}. MLP heads are used to further reduce the dimensionality of the feature space from 2048 to 512. Each model was trained on an NVidia Titan V GPU with 12GB of memory for approximately 2-3 days depending on the object category and subsequent dataset size. For evaluation, we use the test set images as the query set and the training renderings as the reference set. 

We train with a batch size of 32 sample pairs for 1000 epochs using the Adam Optimizer \cite{kingma2015adam} with a learning rate of $10^{-4}$ for the ResNet50 encoders and $10^{-3}$ for the MLP heads. A weight decay equal to $5 \cdot 10^{-4}$ is used for both the backbone and MLP head. The embedding size is set to 512 and the loss margin is set to $m=1$. For sampling and mining we use a positive/negative threshold $t_{\Delta\theta}=5^\circ$. Finally, unless stated otherwise, we use $\beta_{train}=0.1$ and $s_{occ}=0.5$. To make a more fair evaluation, we train with occlusions produced from PASCAL VOC 2012, and not from MS-COCO  \cite{lin2014mscoco} as in OccludedPASCAL3D. At the same time, we intentionally use smaller occluders compared to the L1-L3 sets as can be observed by comparing Figure \ref{fig:methodology:occlusions} and Figure \ref{fig:supplementary:data-overview:pascal3d:images}.

We evaluate our approach on PASCAL3D~\cite{pascal3d}, its synthetically occluded counterpart OccludedPASCAL3D~\cite{wang2020occludedpascal}, and KITTI3D \cite{Geiger2012kitti3d}.
Unless stated otherwise, we use surface normal maps in all experiments. An evaluation of the various rendering types is included in the appendix \ref{appendix:rendering-types}. Similar to \cite{wang2020NeMo} we assume known center and distance for all samples, however, we train to achieve robustness to bounding box perturbations and avoid over-reliance on 2D detection accuracy by employing training time bounding box augmentations. As a result, our models learn to disregard distance information when comparing camera images and renderings and instead solely focus on pose information. We further increase the training data variance through horizontal flipping, color jittering, gaussian blurring, bounding box perturbations, and synthetic occlusions. 

We compare the performance of our approach against a category-agnostic classifier Res-50-A and a category-specific classifier Res50-S from \cite{wang2020NeMo} as well as three state-of-the-art competing methods, namely StarMap \cite{zhou2018starmap}, NeMo \cite{wang2020NeMo}, and PoseContrast \cite{xiao2021posecontrast}. For NeMo, in particular, we compare against all three variations termed as NeMo, NeMo-MultiCuboid (NeMo-M), and NeMo-SingleCuboid (NeMo-S). We follow the exact same preprocessing and evaluation methodology as in NeMo and thus borrow their results and the results for Res50-A, Res50-S, and StarMap. PoseContrast, however, was originally trained with more data, so we had to retrain it with the same amount of data as the rest of the methods to ensure a fair comparison. Similar to StarMap and NeMo, we perform the evaluation using three metrics, namely pose accuracy with a threshold of $10^\circ$ ($ACC_\frac{\pi}{18}$) and $30^\circ$ ($ACC_\frac{\pi}{6}$), as well as \textit{Median Error}. Furthermore, we evaluate the inference speed of our approach and compare it against the best competing methods, namely NeMo and PoseContrast.

\subsection{Robust and Efficient 3D Pose Estimation} \label{ssec:experiments:results}
In Tables \ref{tab:experiments:baselines:pascal3d} and \ref{tab:results:pascal3d:per-category} we present our performance against the competing methods from \cite{zhou2018starmap,wang2020NeMo,xiao2021posecontrast}. The results in Table \ref{tab:experiments:baselines:pascal3d} are averaged over all object categories for the levels of synthetic occlusion $0\%$ (L0), $20$-$40\%$ (L1), $40$-$60\%$ (L2), and $60$-$80\%$ (L3), respectively. Similar to \cite{wang2020NeMo}, we use a weighted average that takes into account the number of samples per object category. Overall, our approach outperforms the competing methods across all occlusion levels showcasing the benefit of strong data augmentation compared to complex and specialized architectures.

\begin{table*}[t]
\setlength{\tabcolsep}{3.1pt}
\centering
\begin{tabular}{| l | c |  c c c c | c c c c | c c c c |}
\hline
& Categ. & \multicolumn{4}{c|}{$ACC_\frac{\pi}{6} \uparrow$} & \multicolumn{4}{c|}{$ACC_\frac{\pi}{18} \uparrow$} & \multicolumn{4}{c|}{$MedErr \downarrow$} \\
& aware & L0 & L1 & L2 & L3 & L0 & L1 & L2 & L3 & L0 & L1 & L2 & L3\\
\hline
Res50-A $\ssymbol{2}$ &  & 88.1 & 70.4 & 52.8 & 37.8 & 44.6 & 25.3 & 14.5 & 6.7 & 11.7 & 17.9 & 30.4 & 46.4 \\
Res50-S $\ssymbol{2}$ & $\checkmark$ & 87.6 & 73.2 & 58.4 & 43.1 & 43.9 & 28.1 & 18.6 & 9.9 & 11.8 & 17.3 & 26.1 & 44.0 \\
StarMap $\ssymbol{2}$ &  & 89.4 & 71.1 & 47.2 & 22.9 & 59.5 & 34.4 & 13.9 & 3.7 & 9.0 & 17.6 & 34.1 & 63.0\\
NeMo $\ssymbol{2}$ & $\checkmark$ & 84.1 & 73.1 & 59.9 & 41.3 & 60.4 & 45.1 & 30.2 & 14.5 & 9.3 & 15.6 & 24.1 & 41.8 \\
NeMo-M $\ssymbol{2}$ & $\checkmark$ & 86.7 & 77.2 & 65.2 & 47.1 & 63.2 & 49.9 & 34.5 & 17.8 & 8.2 & 13.0 & 20.2 & \textbf{36.1} \\
NeMo-S $\ssymbol{2}$ & $\checkmark$ & 86.1 & 76.0 & 63.9 & 46.8 & 61.0 & 46.3 & 32.0 & 17.1 & 8.8 & 13.6 & 20.2 & 36.5 \\
PoseCon &  & 90.8 & 76.2 & 59.3 & 39.7 & 67.2 & 46.4 & 28.1 & 12.7 & 7.1 & 12.6 & 23.1 & 45.5 \\
Ours & $\checkmark$ & \textbf{92.3} & \textbf{85.7} & \textbf{72.7} & \textbf{49.8} & \textbf{72.2} & \textbf{56.7} & \textbf{38.9} & \textbf{17.9} & \textbf{6.6} & \textbf{9.7} & \textbf{16.0} & 37.9\\
\hline
\end{tabular}
\caption{Evaluation against the state-of-the-art on PASCAL3D (L0) and OccludedPASCAL3D (L1-L3). The results are averaged across the 12 object categories and the symbol $\ssymbol{2}$ denotes results taken from \cite{wang2020NeMo}.} 
\label{tab:experiments:baselines:pascal3d}
\end{table*}

\begin{table}[t]
\setlength{\tabcolsep}{1.8pt}
\centering
\begin{tabular}{|c| l | c c c c c c c c c c c c | c |}
\hline
Occl. & Method & aero & bike & boat & bottle & bus & car & chair & table & mbike & sofa & train & tv & Mean\\
\hline
& NeMo-M & 76.9 & 82.2 & 66.5 & 87.1 & 93.0 & 98.0 & 90.1 & 80.5 & 81.8 & 96.0 & 89.3 & 87.1 & 86.7 \\
L0 & PoseCon. & 83.7 & 84.0 & \textbf{82.5} & 88.9 & 97.7 & 96.7 & 95.3 & 86.9 & \textbf{87.2} & \textbf{97.1} & 96.7 & 87.8 & 90.8 \\
& Ours & \textbf{84.4} & \textbf{88.1} & \textbf{82.5} & \textbf{91.7} & \textbf{98.7} & \textbf{99.2} & \textbf{95.9} & \textbf{88.8} & 85.6 & 97.0 & \textbf{98.0} & \textbf{90.0} & \textbf{92.3}\\
\hline
& NeMo-M & 58.1 & 68.8 & 53.4 & 78.8 & 86.9 & 94.0 & 76.0 & 70.0 & 61.8 & 87.3 & 82.8 & 82.8 & 77.2\\
L1 & PoseCon. & 57.7 & 66.6 & 56.9 & \textbf{86.7} & 87.1 & 83.6 & 66.9 & 74.2 & 72.3 & 90.6 & 89.4 & 78.2 & 76.2 \\
& Ours & \textbf{71.4} & \textbf{79.2} & \textbf{70.6} & 85.2 & \textbf{87.7} & \textbf{97.4} & \textbf{87.2} & \textbf{81.9} & \textbf{78.4} & \textbf{94.1} & \textbf{96.5} & \textbf{80.0} & \textbf{85.7}\\
\hline
& NeMo-M & 43.1 & \textbf{55.7} & 43.3 & 69.1 &\textbf{ 79.8} & 84.5 & 58.8 & 58.4 & 43.9 & 76.4 & 64.3 & \textbf{70.3} & 65.2\\
L2 & PoseCon. & 38.5 & 51.2 & 39.2 & \textbf{81.8} & 69.5 & 61.8 & 49.3 & 57.6 & 56.1 & 74.1 & 82.4 & 61.0 & 59.3\\
& Ours & \textbf{54.6} & 54.6 & \textbf{55.4} & 68.8 & 71.0 & \textbf{91.5} & \textbf{66.5} & \textbf{67.8} &\textbf{ 57.9} & \textbf{84.4} & \textbf{93.1} & 67.3 & \textbf{72.7} \\
\hline
& NeMo-M & 23.8 & \textbf{34.3} & 29.5 & 53.9 & \textbf{56.0} & 65.5 & \textbf{43.4} & 41.5 & 25.4 & 58.2 & 43.2 & \textbf{54.1} & 47.1\\
L3 & PoseCon. & 19.2 & 30.6 & 27.4 & \textbf{73.5} & 47 & 35.2 & \textbf{33.3} & 38.0 & 33.3 & 52.1 & 70.7 & 44.4 & 39.7 \\
& Ours & \textbf{27.4} & 28.8 & \textbf{31.8} & 43.3 & 41.3 & \textbf{69.6} & 40.9 & \textbf{45.6} & 32.1 & \textbf{62.1} & \textbf{85.2} & 47.8 & \textbf{49.8}\\
\hline
\end{tabular}
\caption{$ACC_\frac{\pi}{6}$ per PASCAL3D category against the two best competing methods \cite{wang2020NeMo,xiao2021posecontrast}.}
\label{tab:results:pascal3d:per-category}
\end{table}

In Table~\ref{tab:experiments:baselines:kitti3d}, we present our results on KITTI3D. Since the test set of KITTI3D does not provide labels, we split the training set based on the 50-50 split proposed by \cite{chen20153dop}. NeMo-MultiCuboid requires car type labels and StarMap requires object keypoints which are not provided by KITTI3D, so we evaluate solely against NeMo-SingleCuboid and PoseContrast. Even without retraining or fine-tuning on KITTI3D, our approach exhibits similar performance as in PASCAL3D, outperforming the competing methods in the Fully-Visible (FV), Partly-Occluded (PO) and Largely-Occluded (LO) evaluation categories. At the same time, we demonstrate robustness to out-of-distribution occlusions considering that we trained on cars without same-category occlusions.


\begin{table*}[t]
\setlength{\tabcolsep}{3.5pt}
\centering
\begin{tabular}{| l |  c c c c | c c c c | c c c c |}
\hline
& \multicolumn{4}{c|}{$ACC_\frac{\pi}{6} \uparrow$} & \multicolumn{4}{c|}{$ACC_\frac{\pi}{18} \uparrow$} & \multicolumn{4}{c|}{$MedErr \downarrow$} \\
& FV & PO & LO & All & FV & PO & LO & All & FV & PO & LO & All \\
\hline
NeMo-S & 88.1 & 72.4 & 34.9 & 67.9 & 70.3 & 40.4 & 7.5 & 43.7 & 7.3 & 11.6 & 46.1 & 20.0 \\
PoseContrast & 97.8 & 88.5 & 48.6 & 80.6 & 81.6 & 62.4 & 18.6 & 57.5 & 6.6 & 8.6 & 33.0 & 15.0 \\
Ours & \textbf{98.1} & {90.0} & {56.1} & {83.4} & {92.8} & {70.6} & {21.0} & {65.3} & {3.2} & {5.4} & {24.8} & {10.2} \\
Ours-2 & 97.9 & \textbf{90.6} & \textbf{66.5} & \textbf{86.5} & \textbf{94.2} & \textbf{74.4} & \textbf{34.4} & \textbf{70.9} & \textbf{2.9} & \textbf{5.3} & \textbf{15.5} & \textbf{7.3} \\
\hline
\end{tabular}
\caption{Evaluation on cars of KITTI3D without retraining or fine-tuning. Ours-2 was trained with higher bounding box noise $\beta_{train}=0.75$ showcasing the benefit of this augmentation technique to cross-dataset performance.}
\label{tab:experiments:baselines:kitti3d}
\end{table*}

\begin{figure}
\begin{minipage}[t]{0.47\textwidth}
\centering
\includegraphics[width=\textwidth]{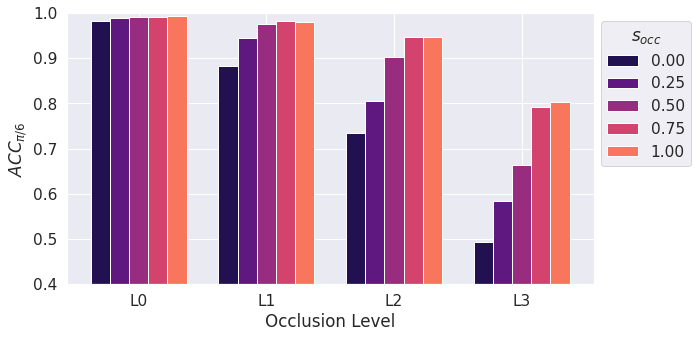} 
\vspace{-0.8cm}
\caption{Comparison of models trained with different levels $s_{occ}$ of synthetic occlusion and evaluated on L0-L3.}
\label{fig:experiments:occlusions}
\end{minipage}\hfill
\begin{minipage}[t]{0.47\textwidth}
\centering
\includegraphics[width=\textwidth]{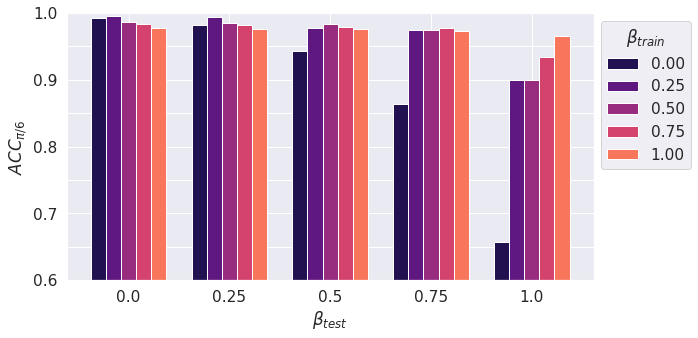}
\vspace{-0.8cm}
\caption{Comparison of models trained with different bounding box noise levels $\beta_{train}$ on perturbed L0 by $\beta_{test}$.}
\label{fig:experiments:bbox-noise}
\end{minipage}
\end{figure}

To evaluate the effect of occlusion augmentation in our approach, we trained five models with different occlusion scales $s_{occ}$ and evaluated them on $L0$-$L3$. As shown in Figure~\ref{fig:experiments:occlusions}, the higher the occlusion scale $s_{occ}$ during training, the more robust the model becomes to higher levels of occlusions. We also note that even on $L0$ the models trained with synthetic occlusions outperform the ones without demonstrating robustness to real occlusions and clutter. 

To evaluate the effect of bounding box augmentation we train five models with different levels of bounding box noise and evaluate each one on increasing levels of test time augmentation. Based on the graphs in Figure~\ref{fig:experiments:bbox-noise}, training with bounding box augmentations leads to increased robustness to higher $\beta_{test}$ noise values. However, performance drops slightly for larger $\beta_{train}$ values when evaluating on the unperturbed datasets ($\beta_{test}=0$). 

\begin{table*}[t]
\setlength{\tabcolsep}{2.9pt}
\centering
\begin{tabular}{| l |  c c c c | c c c c | c c c c |}
\hline
& \multicolumn{4}{c|}{$ACC_\frac{\pi}{6} \uparrow$} & \multicolumn{4}{c|}{$ACC_\frac{\pi}{18} \uparrow$} & \multicolumn{4}{c|}{$MedErr \downarrow$} \\
& L0 & L1 & L2 & L3 & L0 & L1 & L2 & L3 & L0 & L1 & L2 & L3\\
\hline
Ours & \textbf{99.2} & \textbf{97.4} & 91.5 & 69.6 & 95.9 & \textbf{89.3} & \textbf{68.8} & \textbf{32.6} & 3.1 & \textbf{4.2} & \textbf{6.7} & \textbf{16.1} \\
w/o multi-CAD & 99.0 & 97.3 & \textbf{91.8} & 70.1 & \textbf{96.2} & 88.1 & 65.4 & 30.2 & \textbf{3.0} & 4.3 & 7.3 & 16.2 \\
w/o data augment. & \textbf{99.2} & 97.2 & 90.3 & \textbf{70.7} & 94.9 & 85.0 & 62.9 & 29.9 & 3.5 & 4.7 & 7.8 & 16.2 \\
w/ same cat. occl. & 99.0 & 97.1 & 90.8 & 68.2 & 95.8 & 87.4 & 64.5 & 28.6 & 3.6 & 4.7 & 7.5 & 17.3\\
w/o contin. labels & 98.6 & 95.5 & 86.4 & 59.7 & 90.4 & 78.0 & 54.0 & 23.0 & 5.2 & 6.3 & 9.3 & 22.3 \\
w/ Triplet Loss \cite{zhakarov2017dynamic} & 98.3 & 94.8 & 84.6 & 60.6 & 93.6 & 83.6 & 62.5 & 31.6 & 4.2 & 5.3 & 7.7 & 17.1 \\
w/o syn. occlusions & 97.8 & 88.3 & 74.9 & 54.6 & 94.7 & 67.8 & 37.7 & 13.8 & 3.2 & 6.9 & 13.3 & 25.9  \\
w/o dyn. margin & 36.9 & 36.5 & 35.8 & 33.4 & 25.5 & 19.2 & 13.7 & 7.4 & 57.9 & 56.7 & 53.7 & 50.0 \\
\hline
\end{tabular}
\caption{Ablation study results on cars of PASCAL3D L0-L3}
\label{tab:experiments:ablation}
\end{table*}

To evaluate the inference speed of our approach we count the duration of embedding a query image and retrieving the closest neighbour. We ran this experiment on a consumer-grade GPU, namely NVidia GTX 1050, and averaged the measurements over one thousand runs. Our approach runs at $35 fps$ or requires approximately $29ms$ per object instance, significantly faster than the 8-second inference of NeMo, but still almost double the $15ms$ inference time of PoseContrast that uses a mix of classification and regression.

\begin{figure}[t]
\centering
\begin{subfigure}{0.48\textwidth}
\centering
\includegraphics[height=2.2cm, width=\linewidth]{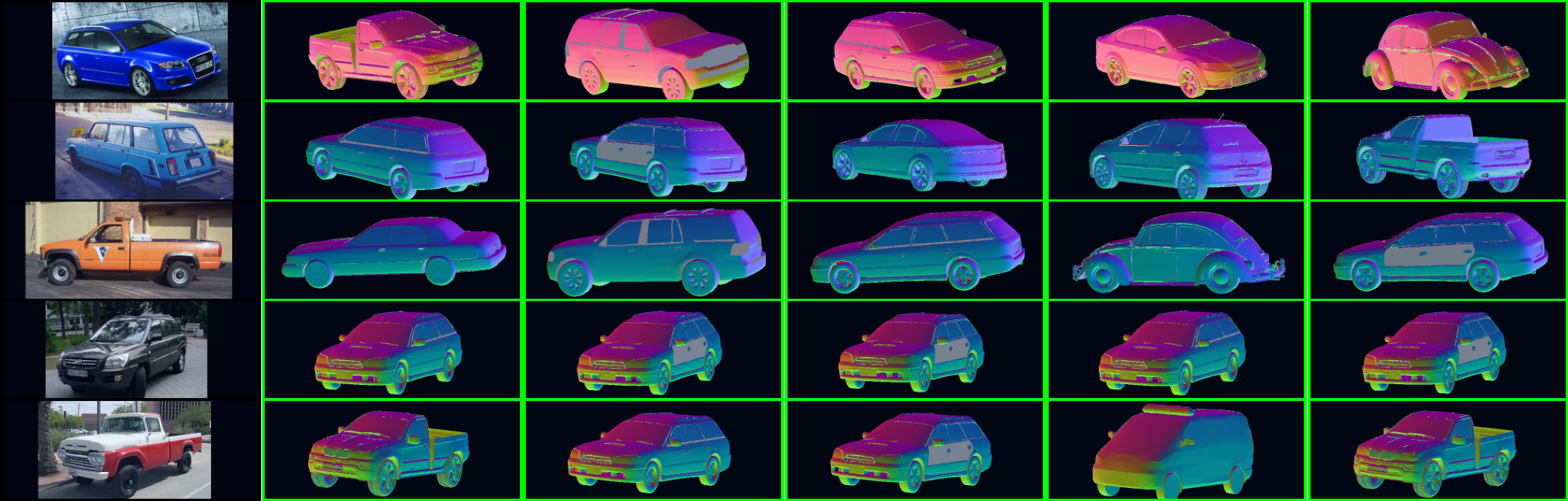}
\caption{Successful retrievals ($\Delta\theta<10^\circ$).}
\label{fig:experiments:retrievals:successes}
\end{subfigure}\hfill%
\begin{subfigure}{0.48\textwidth}
\centering
\includegraphics[height=2.2cm, width=\linewidth]{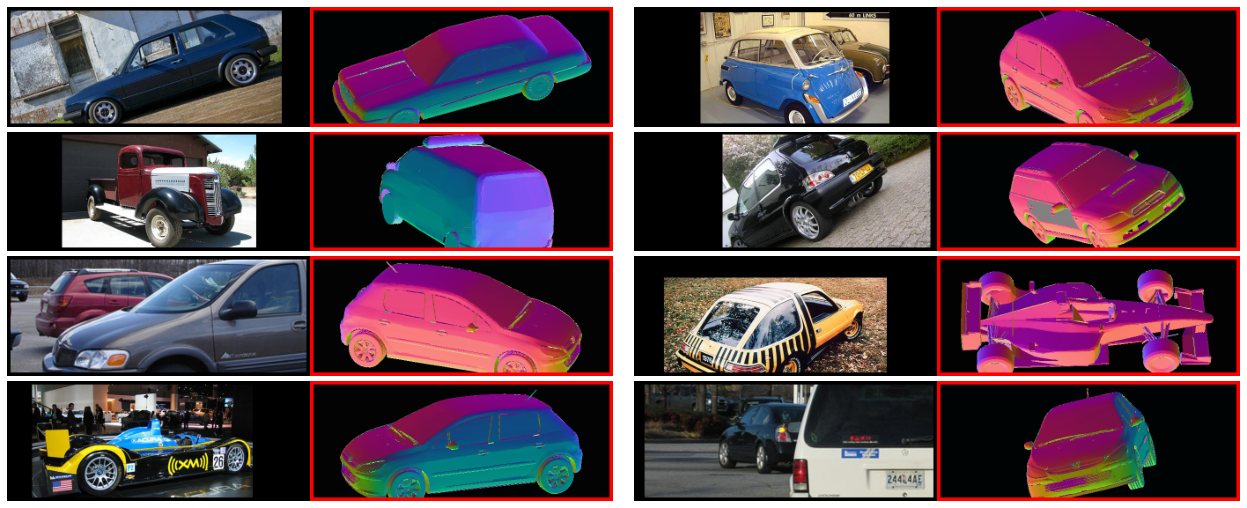}
\caption{Failed retrievals ($\Delta\theta>30^\circ$).}
\label{fig:experiments:retrievals:failures}
\end{subfigure}%
\caption{Examples of nearest neighbour retrievals.}
\end{figure}

In Figure~\ref{fig:experiments:retrievals:successes}, we present examples of queries with their corresponding rendered retrievals demonstrating pose estimation accuracy and feature extraction invariant to specific CAD models. 
In Figure~\ref{fig:experiments:retrievals:failures} we present failure cases of queries with wrong retrievals which have an angle error higher than $30^\circ$. We observe that most failure cases are due to confusion between opposite directions, same-category occlusions, rarely-seen poses, or atypical vehicles. More examples are included in appendix \ref{appendix:qualitative-results}. 



\subsection{Ablation Study} \label{ssec:experiments:ablation}
We conducted an ablation study in which we evaluated our approach for seven distinct cases, as presented in Table~\ref{tab:experiments:ablation}. First, we trained with only a single CAD model arbitrarily chosen as the sedan for the car category. Second, we trained without appearance augmentations (color jitter, gaussian blur, horizontal flipping). Third, we evaluated our approach with same-category occlusions. Fourth, we used discretized pose labels. Fifth, we used a Triplet loss with a dynamic margin rather than our proposed contrastive loss. Sixth, we trained without synthetic occlusions and in the final case, we removed the dynamic margin. In all cases we observed a non-negligible decrease in performance. 

\section{Conclusion} \label{sec:conclusion}

Object Pose estimation in a monocular setting is a non-trivial task, especially when dealing with occlusions, clutter, and appearance variations that make handcrafted approaches more prone to error or lack of accuracy. Therefore, we propose learning a pose alignment metric using a contrastive loss with a dynamic margin for comparing object images and renderings with regard to their pose. We reinforce the robustness of the metric using synthetic occlusions and other appearance augmentations. The metric learnt with our Contrastive Pose Loss can be used for pose estimation in an efficient real-time image retrieval setting and achieves state-of-the-art performance on PASCAL3D and OccludedPASCAL3D, as well as high cross-dataset performance on KITTI3D.

\section*{Acknowledgements}
We gratefully acknowledge funding support from the Sim2Real project, in the context of the Ford-KU Leuven alliance program.

\bibliography{egbib}

\begin{thebibliography}{45}
\providecommand{\natexlab}[1]{#1}
\providecommand{\url}[1]{\texttt{#1}}
\expandafter\ifx\csname urlstyle\endcsname\relax
  \providecommand{\doi}[1]{doi: #1}\else
  \providecommand{\doi}{doi: \begingroup \urlstyle{rm}\Url}\fi

\bibitem[Angtian et~al.(2021)Angtian, Kortylewski, and Yuille]{wang2020NeMo}
Wang Angtian, Adam Kortylewski, and Alan Yuille.
\newblock Nemo: Neural mesh models of contrastive features for robust 3d pose
  estimation.
\newblock In \emph{Proceedings International Conference on Learning
  Representations (ICLR)}, 2021.

\bibitem[Balntas et~al.(2017)Balntas, Doumanoglou, Sahin, Sock, Kouskouridas,
  and Kim]{Balntas2017PoseGR}
Vassileios Balntas, Andreas Doumanoglou, Caner Sahin, Juil Sock, Rigas
  Kouskouridas, and Tae-Kyun Kim.
\newblock Pose guided rgbd feature learning for 3d object pose estimation.
\newblock \emph{2017 IEEE International Conference on Computer Vision (ICCV)},
  pages 3876--3884, 2017.

\bibitem[Beker et~al.(2020)Beker, Kato, Morariu, Ando, Matsuoka, Kehl, and
  Gaidon]{beker2020monocular}
Deniz Beker, Hiroharu Kato, Mihai~Adrian Morariu, Takahiro Ando, Toru Matsuoka,
  Wadim Kehl, and Adrien Gaidon.
\newblock Monocular differentiable rendering for self-supervised 3d object
  detection.
\newblock In Andrea Vedaldi, Horst Bischof, Thomas Brox, and Jan-Michael Frahm,
  editors, \emph{Computer Vision -- ECCV 2020}, Cham, 2020. Springer
  International Publishing.
\newblock ISBN 978-3-030-58589-1.

\bibitem[Chen et~al.(2015)Chen, Kundu, Zhu, Berneshawi, Ma, Fidler, and
  Urtasun]{chen20153dop}
Xiaozhi Chen, Kaustav Kundu, Yukun Zhu, Andrew~G Berneshawi, Huimin Ma, Sanja
  Fidler, and Raquel Urtasun.
\newblock 3d object proposals for accurate object class detection.
\newblock In C.~Cortes, N.~Lawrence, D.~Lee, M.~Sugiyama, and R.~Garnett,
  editors, \emph{Advances in Neural Information Processing Systems}, volume~28.
  Curran Associates, Inc., 2015.
\newblock URL
  \url{https://proceedings.neurips.cc/paper/2015/file/6da37dd3139aa4d9aa55b8d237ec5d4a-Paper.pdf}.

\bibitem[Chen et~al.(2020)Chen, Liu, Shen, and Jia]{chen2020dsgn}
Yilun Chen, Shu Liu, Xiaoyong Shen, and Jiaya Jia.
\newblock Dsgn: Deep stereo geometry network for 3d object detection.
\newblock In \emph{2020 IEEE/CVF Conference on Computer Vision and Pattern
  Recognition (CVPR)}, pages 12533--12542, 2020.
\newblock \doi{10.1109/CVPR42600.2020.01255}.

\bibitem[Chopra et~al.(2005)Chopra, Hadsell, and LeCun]{chopra2005contrastive}
S.~Chopra, R.~Hadsell, and Y.~LeCun.
\newblock Learning a similarity metric discriminatively, with application to
  face verification.
\newblock In \emph{2005 IEEE Computer Society Conference on Computer Vision and
  Pattern Recognition (CVPR'05)}, volume~1, pages 539--546 vol. 1, 2005.
\newblock \doi{10.1109/CVPR.2005.202}.

\bibitem[Everingham et~al.(2012)Everingham, Van~Gool, Williams, Winn, and
  Zisserman]{pascal-voc-2012}
M.~Everingham, L.~Van~Gool, C.~K.~I. Williams, J.~Winn, and A.~Zisserman.
\newblock {T}he {PASCAL} {V}isual {O}bject {C}lasses {C}hallenge 2012
  ({VOC2012}) {R}esults, 2012.
\newblock URL
  \url{http://www.pascal-network.org/challenges/VOC/voc2012/workshop/index.html}.

\bibitem[Garg et~al.(2020)Garg, Wang, Hariharan, Campbell, Weinberger, and
  Chao]{div2020wstereo}
Divyansh Garg, Yan Wang, Bharath Hariharan, Mark Campbell, Kilian~Q.
  Weinberger, and Wei-Lun Chao.
\newblock Wasserstein distances for stereo disparity estimation.
\newblock In \emph{Proceedings of the 34th International Conference on Neural
  Information Processing Systems}, NIPS'20, Red Hook, NY, USA, 2020. Curran
  Associates Inc.
\newblock ISBN 9781713829546.

\bibitem[Geiger et~al.(2012)Geiger, Lenz, and Urtasun]{Geiger2012kitti3d}
Andreas Geiger, Philip Lenz, and Raquel Urtasun.
\newblock Are we ready for autonomous driving? the kitti vision benchmark
  suite.
\newblock In \emph{2012 IEEE Conference on Computer Vision and Pattern
  Recognition}, pages 3354--3361, 2012.
\newblock \doi{10.1109/CVPR.2012.6248074}.

\bibitem[Grabner et~al.(2020)Grabner, Wang, Zhang, Guo, Xiao, Vajda, Roth, and
  Lepetit]{grabner2020correspondence}
Alexander Grabner, Yaming Wang, Peizhao Zhang, Peihong Guo, Tong Xiao, Peter
  Vajda, Peter~M. Roth, and Vincent Lepetit.
\newblock Geometric correspondence fields: Learned differentiable rendering for
  3d pose refinement in the wild.
\newblock In Andrea Vedaldi, Horst Bischof, Thomas Brox, and Jan-Michael Frahm,
  editors, \emph{Computer Vision -- ECCV 2020}, pages 102--119, Cham, 2020.
  Springer International Publishing.
\newblock ISBN 978-3-030-58517-4.

\bibitem[Harwood et~al.(2017)Harwood, B.G., Carneiro, Reid, and
  Drummond]{harwood2017smart}
B.~Harwood, V.~Kumar B.G., G.~Carneiro, I.~Reid, and T.~Drummond.
\newblock Smart mining for deep metric learning.
\newblock In \emph{2017 IEEE International Conference on Computer Vision
  (ICCV)}, pages 2840--2848, Los Alamitos, CA, USA, oct 2017. IEEE Computer
  Society.
\newblock \doi{10.1109/ICCV.2017.307}.
\newblock URL \url{https://doi.ieeecomputersociety.org/10.1109/ICCV.2017.307}.

\bibitem[He et~al.(2016)He, Zhang, Ren, and Sun]{he2015deep}
Kaiming He, Xiangyu Zhang, Shaoqing Ren, and Jian Sun.
\newblock Deep residual learning for image recognition.
\newblock In \emph{2016 IEEE Conference on Computer Vision and Pattern
  Recognition (CVPR)}, pages 770--778, 2016.
\newblock \doi{10.1109/CVPR.2016.90}.

\bibitem[He and Soatto(2019)]{he2019mono3d}
Tong He and Stefano Soatto.
\newblock Mono3d++: Monocular 3d vehicle detection with two-scale 3d hypotheses
  and task priors.
\newblock \emph{Proceedings of the AAAI Conference on Artificial Intelligence},
  33:\penalty0 8409--8416, 07 2019.
\newblock \doi{10.1609/aaai.v33i01.33018409}.

\bibitem[Hermans et~al.(2017)Hermans, Beyer, and Leibe]{hermans2017defense}
Alexander Hermans, Lucas Beyer, and Bastian Leibe.
\newblock In defense of the triplet loss for person re-identification.
\newblock \emph{CoRR}, abs/1703.07737, 2017.
\newblock URL \url{http://arxiv.org/abs/1703.07737}.

\bibitem[Iwase et~al.(2021)Iwase, Liu, Khirodkar, Yokota, and
  Kitani]{iwase2021repose}
Shun Iwase, Xingyu Liu, Rawal Khirodkar, Rio Yokota, and Kris~M. Kitani.
\newblock Repose: Fast 6d object pose refinement via deep texture rendering.
\newblock In \emph{Proceedings of the IEEE/CVF International Conference on
  Computer Vision (ICCV)}, pages 3303--3312, October 2021.

\bibitem[Johnson et~al.(2020)Johnson, Ravi, Reizenstein, Novotny, Tulsiani,
  Lassner, and Branson]{ravi2020pytorch3d}
Justin Johnson, Nikhila Ravi, Jeremy Reizenstein, David Novotny, Shubham
  Tulsiani, Christoph Lassner, and Steve Branson.
\newblock Accelerating 3d deep learning with pytorch3d.
\newblock In \emph{SIGGRAPH Asia 2020 Courses}, SA '20, New York, NY, USA,
  2020. Association for Computing Machinery.
\newblock ISBN 9781450381123.
\newblock \doi{10.1145/3415263.3419160}.
\newblock URL \url{https://doi.org/10.1145/3415263.3419160}.

\bibitem[Kingma and Ba(2015)]{kingma2015adam}
Diederik~P. Kingma and Jimmy Ba.
\newblock Adam: {A} method for stochastic optimization.
\newblock In Yoshua Bengio and Yann LeCun, editors, \emph{3rd International
  Conference on Learning Representations, {ICLR} 2015, San Diego, CA, USA, May
  7-9, 2015, Conference Track Proceedings}, 2015.
\newblock URL \url{http://arxiv.org/abs/1412.6980}.

\bibitem[Kong et~al.(2020)Kong, Liu, Hu, Fang, and Sun]{kong2020unsupervised}
Huifang Kong, Tiankuo Liu, Jie Hu, Yao Fang, and Jixing Sun.
\newblock Unsupervised monocular depth and pose estimation using multiple masks
  based on photometric and geometric consistency.
\newblock In \emph{2020 Chinese Automation Congress (CAC)}, pages 3558--3563,
  2020.
\newblock \doi{10.1109/CAC51589.2020.9326951}.

\bibitem[Li et~al.(2019)Li, Chen, and Shen]{li2019stereorcnn}
Peiliang Li, Xiaozhi Chen, and Shaojie Shen.
\newblock Stereo r-cnn based 3d object detection for autonomous driving.
\newblock In \emph{2019 IEEE/CVF Conference on Computer Vision and Pattern
  Recognition (CVPR)}, pages 7636--7644, 2019.
\newblock \doi{10.1109/CVPR.2019.00783}.

\bibitem[Lin et~al.(2018)Lin, Xu, Sun, Theobalt, and Chua]{zaw2018disentangled}
Kyaw~Zaw Lin, Weipeng Xu, Qianru Sun, Christian Theobalt, and Tat{-}Seng Chua.
\newblock Learning a disentangled embedding for monocular 3d shape retrieval
  and pose estimation.
\newblock \emph{CoRR}, abs/1812.09899, 2018.
\newblock URL \url{http://arxiv.org/abs/1812.09899}.

\bibitem[Lin et~al.(2020)Lin, Goyal, Girshick, He, and Dollar]{Lin2020FocalLF}
T.~Lin, P.~Goyal, R.~Girshick, K.~He, and P.~Dollar.
\newblock Focal loss for dense object detection.
\newblock \emph{IEEE Transactions on Pattern Analysis \& Machine Intelligence},
  42\penalty0 (02):\penalty0 318--327, feb 2020.
\newblock ISSN 1939-3539.
\newblock \doi{10.1109/TPAMI.2018.2858826}.

\bibitem[Lin et~al.(2014)Lin, Maire, Belongie, Hays, Perona, Ramanan,
  Doll{\'a}r, and Zitnick]{lin2014mscoco}
Tsung-Yi Lin, Michael Maire, Serge Belongie, James Hays, Pietro Perona, Deva
  Ramanan, Piotr Doll{\'a}r, and C.~Lawrence Zitnick.
\newblock Microsoft coco: Common objects in context.
\newblock In David Fleet, Tomas Pajdla, Bernt Schiele, and Tinne Tuytelaars,
  editors, \emph{Computer Vision -- ECCV 2014}, pages 740--755, Cham, 2014.
  Springer International Publishing.
\newblock ISBN 978-3-319-10602-1.

\bibitem[Manmatha et~al.(2017)Manmatha, Wu, Smola, and
  Krahenbuhl]{wu2017sampling}
R.~Manmatha, Chao-Yuan Wu, Alexander Smola, and Philipp Krahenbuhl.
\newblock Sampling matters in deep embedding learning.
\newblock In \emph{2017 IEEE International Conference on Computer Vision
  (ICCV)}, pages 2859--2867, 10 2017.
\newblock \doi{10.1109/ICCV.2017.309}.

\bibitem[Musgrave et~al.(2020)Musgrave, Belongie, and Lim]{musgrave2020pytorch}
Kevin Musgrave, Serge Belongie, and Ser-Nam Lim.
\newblock Pytorch metric learning, 2020.
\newblock URL \url{https://github.com/KevinMusgrave/pytorch-metric-learning}.

\bibitem[Papaioannidis and Pitas(2020)]{papaioannidis2020quaternionlearning}
Christos Papaioannidis and Ioannis Pitas.
\newblock 3d object pose estimation using multi-objective quaternion learning.
\newblock \emph{IEEE Transactions on Circuits and Systems for Video
  Technology}, 30\penalty0 (8):\penalty0 2683--2693, 2020.
\newblock \doi{10.1109/TCSVT.2019.2929600}.

\bibitem[Paszke et~al.(2019)Paszke, Gross, Massa, Lerer, Bradbury, Chanan,
  Killeen, Lin, Gimelshein, Antiga, Desmaison, Kopf, Yang, DeVito, Raison,
  Tejani, Chilamkurthy, Steiner, Fang, Bai, and Chintala]{NEURIPS2019_9015}
Adam Paszke, Sam Gross, Francisco Massa, Adam Lerer, James Bradbury, Gregory
  Chanan, Trevor Killeen, Zeming Lin, Natalia Gimelshein, Luca Antiga, Alban
  Desmaison, Andreas Kopf, Edward Yang, Zachary DeVito, Martin Raison, Alykhan
  Tejani, Sasank Chilamkurthy, Benoit Steiner, Lu~Fang, Junjie Bai, and Soumith
  Chintala.
\newblock Pytorch: An imperative style, high-performance deep learning library.
\newblock In H.~Wallach, H.~Larochelle, A.~Beygelzimer, F.~d\textquotesingle
  Alch\'{e}-Buc, E.~Fox, and R.~Garnett, editors, \emph{Advances in Neural
  Information Processing Systems 32}, pages 8024--8035. Curran Associates,
  Inc., 2019.
\newblock URL
  \url{http://papers.neurips.cc/paper/9015-pytorch-an-imperative-style-high-performance-deep-learning-library.pdf}.

\bibitem[Robinson et~al.(2021)Robinson, Chuang, Sra, and
  Jegelka]{robinson2021contrastive}
Joshua~David Robinson, Ching-Yao Chuang, Suvrit Sra, and Stefanie Jegelka.
\newblock Contrastive learning with hard negative samples.
\newblock In \emph{International Conference on Learning Representations}, 2021.
\newblock URL \url{https://openreview.net/forum?id=CR1XOQ0UTh-}.

\bibitem[Saadi et~al.(2021)Saadi, Besbes, Kramm, and
  Bensrhair]{Saadi2021OptimizingRF}
Loun{\`e}s Saadi, Bassem Besbes, S{\'e}bastien Kramm, and Abdelaziz Bensrhair.
\newblock Optimizing rgb-d fusion for accurate 6dof pose estimation.
\newblock \emph{IEEE Robotics and Automation Letters}, 6\penalty0 (2):\penalty0
  2413--2420, 2021.
\newblock \doi{10.1109/LRA.2021.3061347}.

\bibitem[Schroff et~al.(2015)Schroff, Kalenichenko, and
  Philbin]{schroff2015facenet}
Florian Schroff, Dmitry Kalenichenko, and James Philbin.
\newblock Facenet: A unified embedding for face recognition and clustering.
\newblock \emph{2015 IEEE Conference on Computer Vision and Pattern Recognition
  (CVPR)}, Jun 2015.
\newblock \doi{10.1109/cvpr.2015.7298682}.
\newblock URL \url{http://dx.doi.org/10.1109/CVPR.2015.7298682}.

\bibitem[Su et~al.(2015)Su, Qi, Li, and Guibas]{su2015renderforcnn}
Hao Su, Charles~R. Qi, Yangyan Li, and Leonidas~J. Guibas.
\newblock Render for cnn: Viewpoint estimation in images using cnns trained
  with rendered 3d model views.
\newblock In \emph{2015 IEEE International Conference on Computer Vision
  (ICCV)}, pages 2686--2694, 2015.
\newblock \doi{10.1109/ICCV.2015.308}.

\bibitem[Sárándi et~al.(2018)Sárándi, Linder, Arras, and
  Leibe]{sarandi2018synthetic}
I.~Sárándi, T.~Linder, K.~O. Arras, and B.~Leibe.
\newblock How robust is 3d human pose estimation to occlusion?
\newblock In \emph{IEEE/RSJ Int. Conference on Intelligent Robots and Systems
  (IROS) Workshops}, 2018.

\bibitem[Tian et~al.(2020)Tian, Pan, Ang~Jr, and Lee]{Tian2020Robust6O}
Meng Tian, Liang Pan, Marcelo~H Ang~Jr, and Gim~Hee Lee.
\newblock Robust 6d object pose estimation by learning rgb-d features.
\newblock In \emph{International Conference on Robotics and Automation (ICRA)},
  2020.

\bibitem[Tulsiani et~al.(2015)Tulsiani, Carreira, and
  Malik]{tulsiani2015poseinduction}
S.~Tulsiani, J.~Carreira, and J.~Malik.
\newblock Pose induction for novel object categories.
\newblock In \emph{2015 IEEE International Conference on Computer Vision
  (ICCV)}, pages 64--72, Los Alamitos, CA, USA, dec 2015. IEEE Computer
  Society.
\newblock \doi{10.1109/ICCV.2015.16}.
\newblock URL \url{https://doi.ieeecomputersociety.org/10.1109/ICCV.2015.16}.

\bibitem[Wang et~al.(2020{\natexlab{a}})Wang, Sun, Kortylewski, and
  Yuille]{wang2020occludedpascal}
Angtian Wang, Yihong Sun, Adam Kortylewski, and Alan~L Yuille.
\newblock Robust object detection under occlusion with context-aware
  compositionalnets.
\newblock In \emph{Proceedings of the IEEE/CVF Conference on Computer Vision
  and Pattern Recognition}, pages 12645--12654, 2020{\natexlab{a}}.

\bibitem[Wang et~al.(2020{\natexlab{b}})Wang, Yang, Stückler, and
  Cremers]{wang2020directshape}
Rui Wang, Nan Yang, Jörg Stückler, and Daniel Cremers.
\newblock Directshape: Direct photometric alignment of shape priors for visual
  vehicle pose and shape estimation.
\newblock In \emph{2020 IEEE International Conference on Robotics and
  Automation (ICRA)}, pages 11067--11073, 2020{\natexlab{b}}.
\newblock \doi{10.1109/ICRA40945.2020.9197095}.

\bibitem[Wang et~al.(2019)Wang, Han, Huang, Dong, and Scott]{wang2019multi}
Xun Wang, Xintong Han, Weilin Huang, Dengke Dong, and Matthew~R Scott.
\newblock Multi-similarity loss with general pair weighting for deep metric
  learning.
\newblock In \emph{Proceedings of the IEEE Conference on Computer Vision and
  Pattern Recognition (CVPR)}, pages 5022--5030, 2019.

\bibitem[Wohlhart and Lepetit(2015)]{wohlhart2015learningdescriptors}
Paul Wohlhart and Vincent Lepetit.
\newblock Learning descriptors for object recognition and 3d pose estimation.
\newblock In \emph{2015 IEEE Conference on Computer Vision and Pattern
  Recognition (CVPR)}, pages 3109--3118, 2015.
\newblock \doi{10.1109/CVPR.2015.7298930}.

\bibitem[Xiang et~al.(2014)Xiang, Mottaghi, and Savarese]{pascal3d}
Yu~Xiang, Roozbeh Mottaghi, and Silvio Savarese.
\newblock Beyond pascal: A benchmark for 3d object detection in the wild.
\newblock In \emph{IEEE Winter Conference on Applications of Computer Vision},
  pages 75--82, 2014.
\newblock \doi{10.1109/WACV.2014.6836101}.

\bibitem[Xiang et~al.(2017)Xiang, Schmidt, Narayanan, and
  Fox]{xiang2017posecnn}
Yu~Xiang, Tanner Schmidt, Venkatraman Narayanan, and Dieter Fox.
\newblock Posecnn: {A} convolutional neural network for 6d object pose
  estimation in cluttered scenes.
\newblock \emph{CoRR}, abs/1711.00199, 2017.
\newblock URL \url{http://arxiv.org/abs/1711.00199}.

\bibitem[Xiao et~al.(2021)Xiao, Du, and Marlet]{xiao2021posecontrast}
Y.~Xiao, Y.~Du, and R.~Marlet.
\newblock Posecontrast: Class-agnostic object viewpoint estimation in the wild
  with pose-aware contrastive learning.
\newblock In \emph{2021 International Conference on 3D Vision (3DV)}, pages
  74--84, Los Alamitos, CA, USA, dec 2021. IEEE Computer Society.
\newblock \doi{10.1109/3DV53792.2021.00018}.
\newblock URL
  \url{https://doi.ieeecomputersociety.org/10.1109/3DV53792.2021.00018}.

\bibitem[Xiao et~al.(2019)Xiao, Qiu, Langlois, Aubry, and
  Marlet]{xiao2019posefromshape}
Yang Xiao, Xuchong Qiu, Pierre{-}Alain Langlois, Mathieu Aubry, and Renaud
  Marlet.
\newblock Pose from shape: Deep pose estimation for arbitrary {3D} objects.
\newblock In \emph{British Machine Vision Conference (BMVC)}, 2019.

\bibitem[Xuan et~al.(2020)Xuan, Stylianou, Liu, and Pless]{hong2020hard}
Hong Xuan, Abby Stylianou, Xiaotong Liu, and Robert Pless.
\newblock Hard negative examples are hard, but useful.
\newblock In Andrea Vedaldi, Horst Bischof, Thomas Brox, and Jan-Michael Frahm,
  editors, \emph{Computer Vision -- ECCV 2020}, pages 126--142, Cham, 2020.
  Springer International Publishing.
\newblock ISBN 978-3-030-58568-6.

\bibitem[Zakharov et~al.(2017)Zakharov, Kehl, Planche, Hutter, and
  Ilic]{zhakarov2017dynamic}
Sergey Zakharov, Wadim Kehl, Benjamin Planche, Andreas Hutter, and Slobodan
  Ilic.
\newblock 3d object instance recognition and pose estimation using triplet loss
  with dynamic margin.
\newblock \emph{2017 IEEE/RSJ International Conference on Intelligent Robots
  and Systems (IROS)}, Sep 2017.
\newblock \doi{10.1109/iros.2017.8202207}.
\newblock URL \url{http://dx.doi.org/10.1109/IROS.2017.8202207}.

\bibitem[Zakharov et~al.(2019)Zakharov, Shugurov, and Ilic]{zakharov2019dpod}
Sergey Zakharov, Ivan~S. Shugurov, and Slobodan Ilic.
\newblock Dpod: 6d pose object detector and refiner.
\newblock \emph{2019 IEEE/CVF International Conference on Computer Vision
  (ICCV)}, pages 1941--1950, 2019.

\bibitem[Zhou et~al.(2018)Zhou, Karpur, Luo, and Huang]{zhou2018starmap}
Xingyi Zhou, Arjun Karpur, Linjie Luo, and Qixing Huang.
\newblock Starmap for category-agnostic keypoint and viewpoint estimation.
\newblock \emph{European Conference on Computer Vision (ECCV)}, 2018.

\end{thebibliography}
\newpage
\appendix


\section{Data Overview} \label{appendix:data-overview}
\begin{figure}[t]
\begin{subfigure}{\textwidth}
\centering
\includegraphics[height=1.2cm]{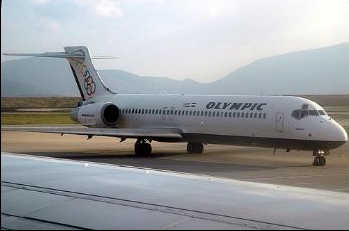}
\includegraphics[height=1.2cm]{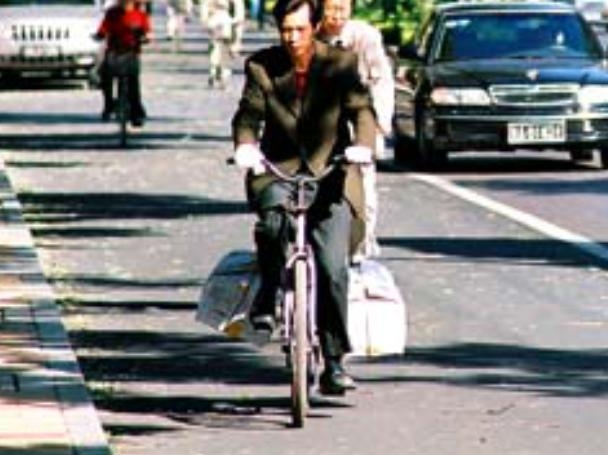}
\includegraphics[height=1.2cm]{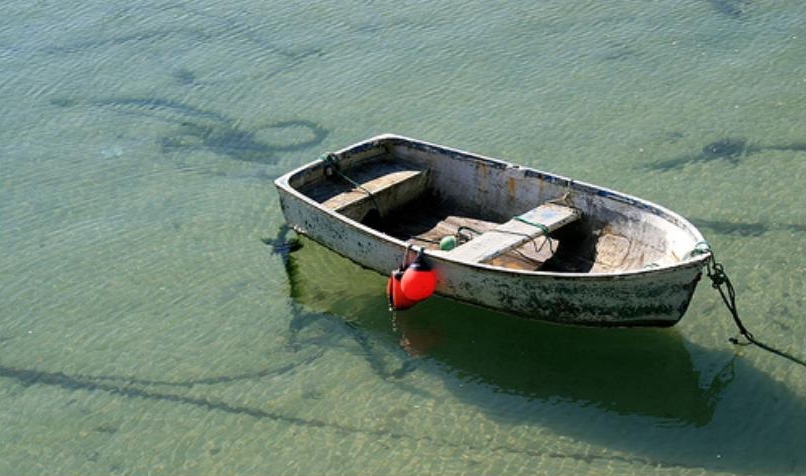}
\includegraphics[height=1.2cm]{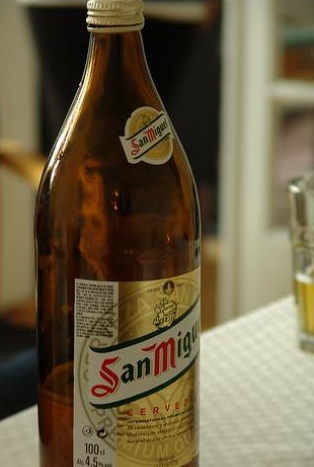}
\includegraphics[height=1.2cm]{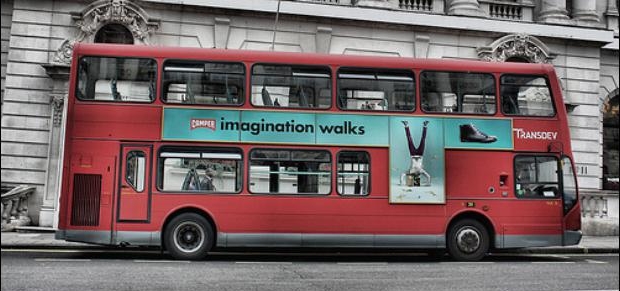}
\includegraphics[height=1.2cm]{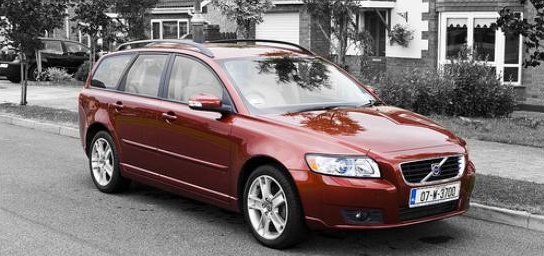}
\includegraphics[height=1.2cm]{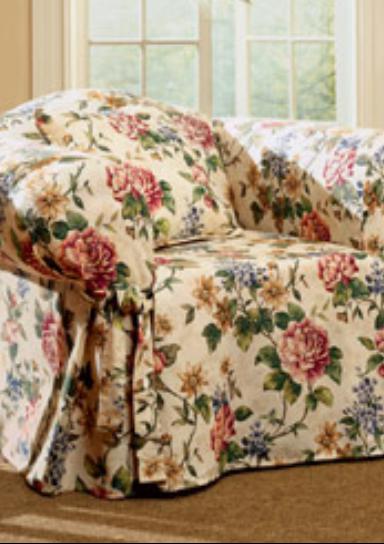}
\includegraphics[height=1.2cm]{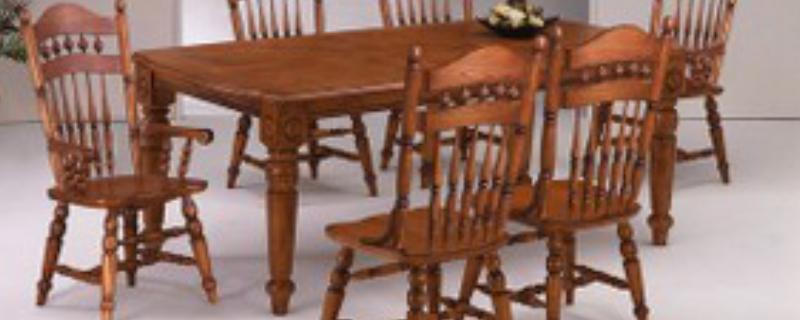}
\includegraphics[height=1.2cm]{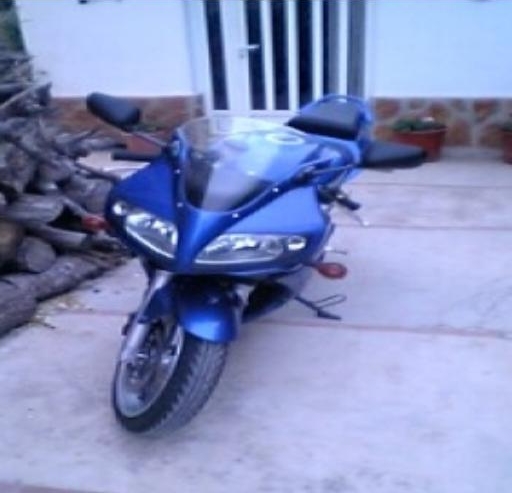}
\includegraphics[height=1.2cm]{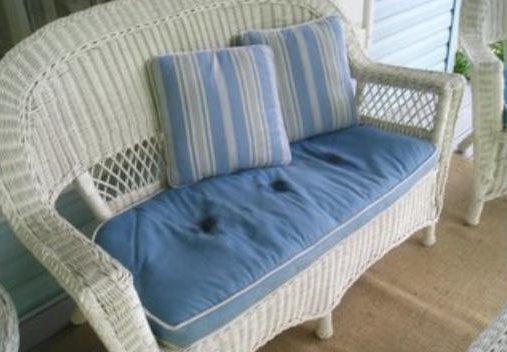}
\includegraphics[height=1.2cm]{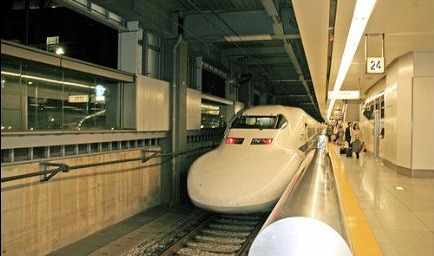}
\includegraphics[height=1.2cm]{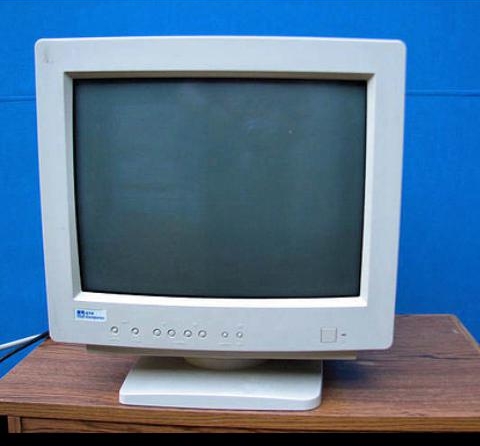}
\caption{L0}
\end{subfigure}\\[1cm]
\begin{subfigure}{\textwidth}
\centering
\includegraphics[height=1.2cm]{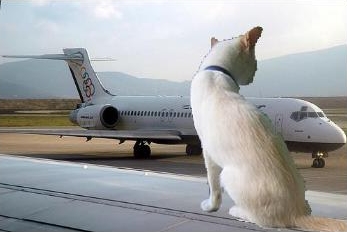}
\includegraphics[height=1.2cm]{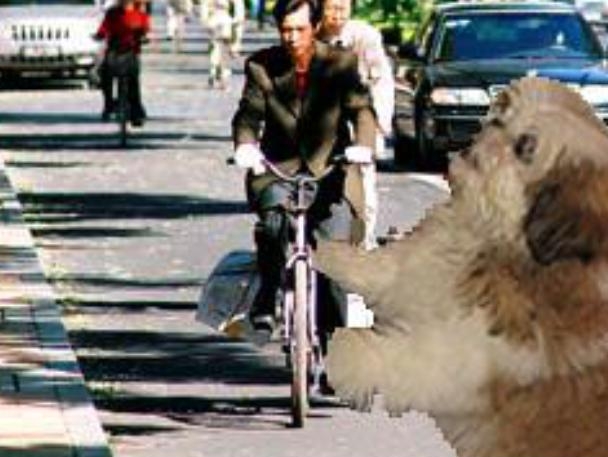}
\includegraphics[height=1.2cm]{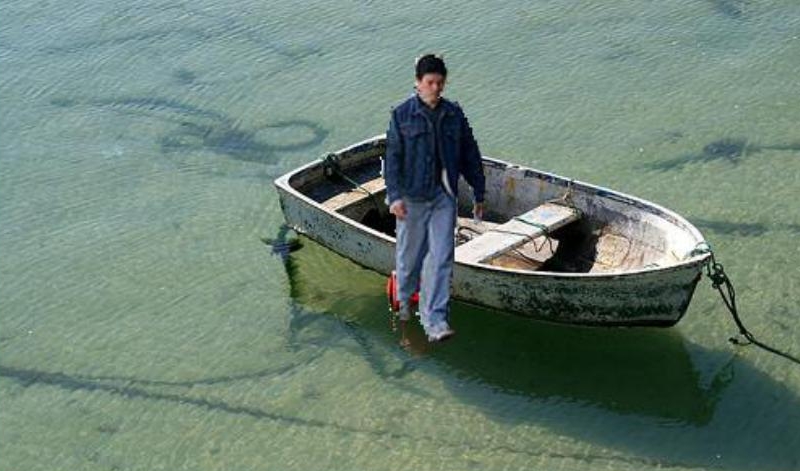}
\includegraphics[height=1.2cm]{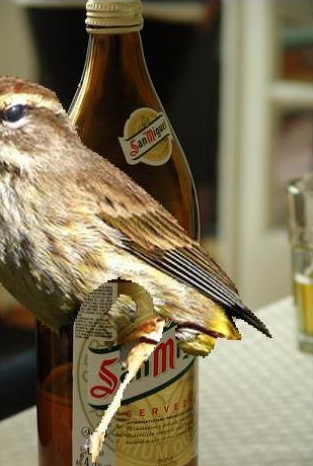}
\includegraphics[height=1.2cm]{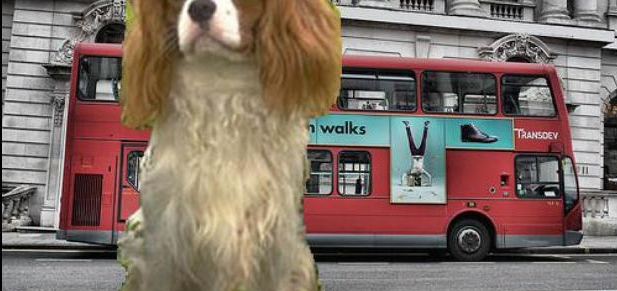}
\includegraphics[height=1.2cm]{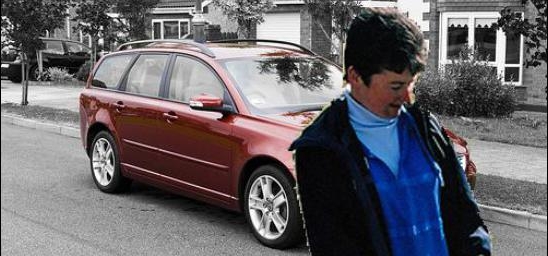}
\includegraphics[height=1.2cm]{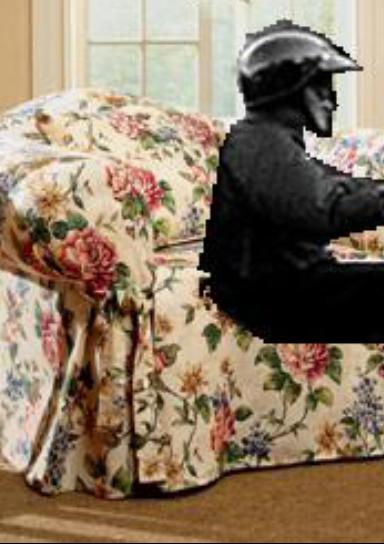}
\includegraphics[height=1.2cm]{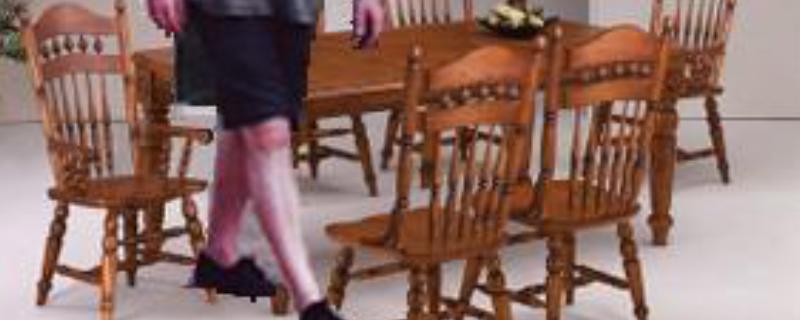}
\includegraphics[height=1.2cm]{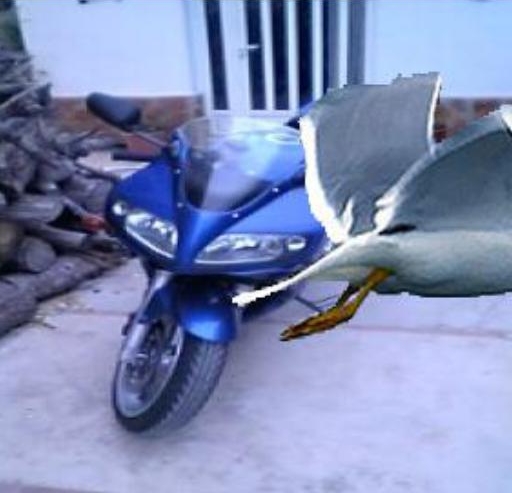}
\includegraphics[height=1.2cm]{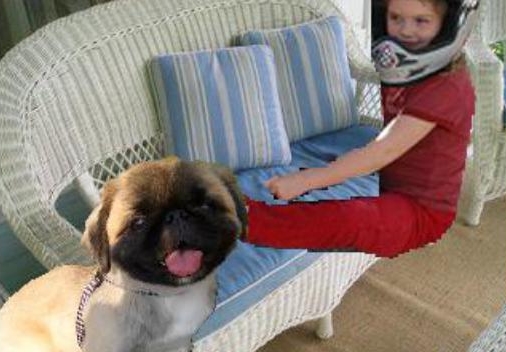}
\includegraphics[height=1.2cm]{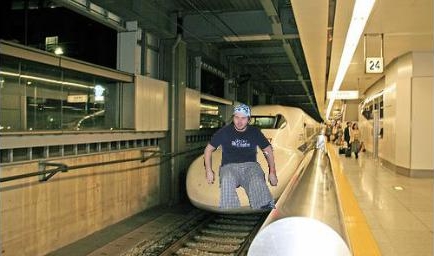}
\includegraphics[height=1.2cm]{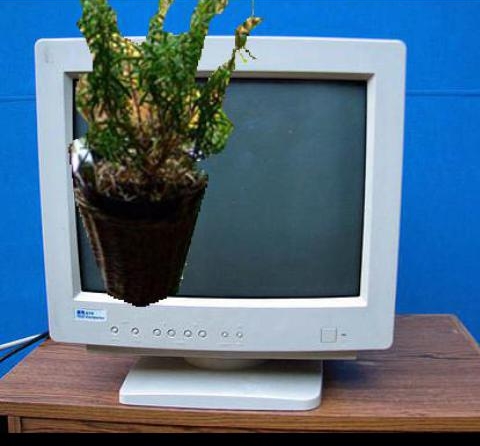}
\caption{L1}
\end{subfigure}\\[1cm]
\begin{subfigure}{\textwidth}
\centering
\includegraphics[height=1.2cm]{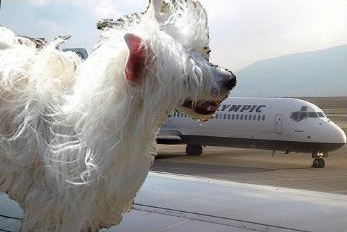}
\includegraphics[height=1.2cm]{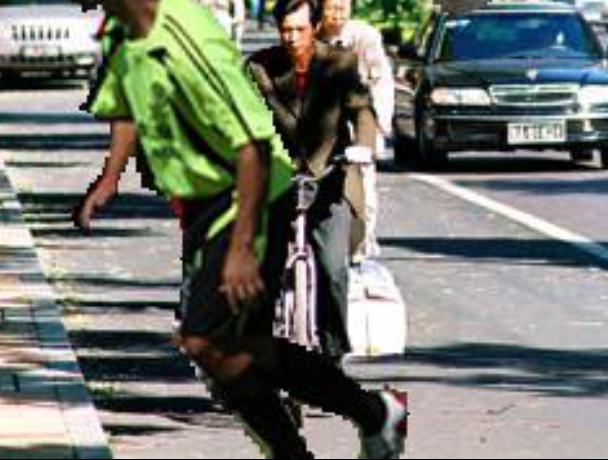}
\includegraphics[height=1.2cm]{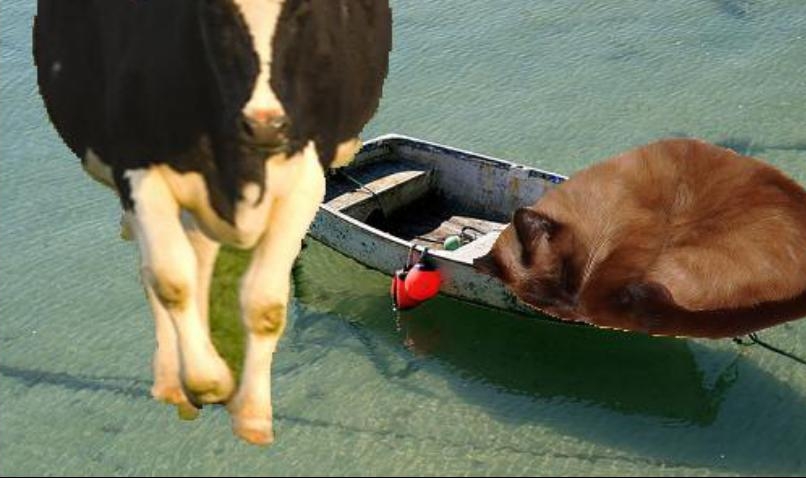}
\includegraphics[height=1.2cm]{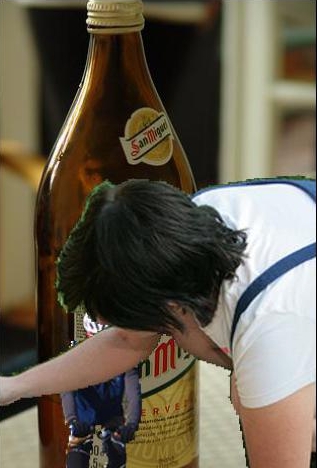}
\includegraphics[height=1.2cm]{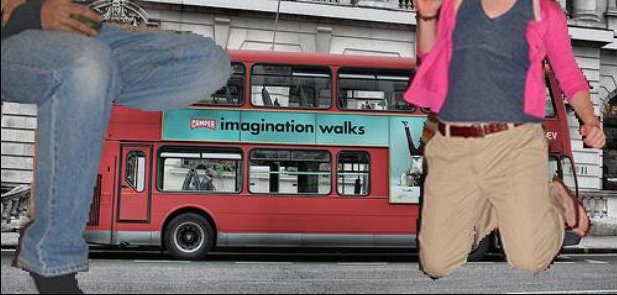}
\includegraphics[height=1.2cm]{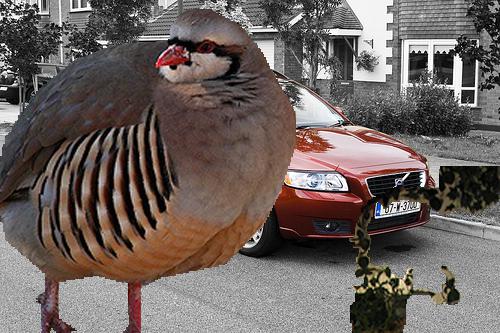}
\includegraphics[height=1.2cm]{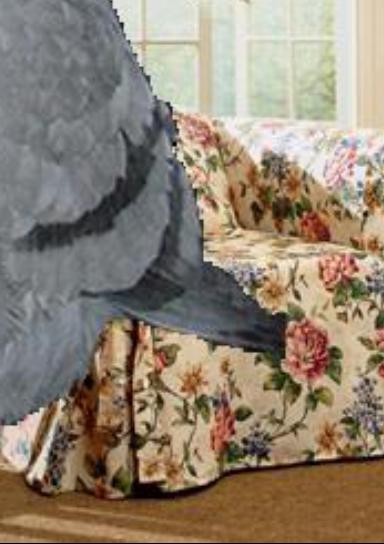}
\includegraphics[height=1.2cm]{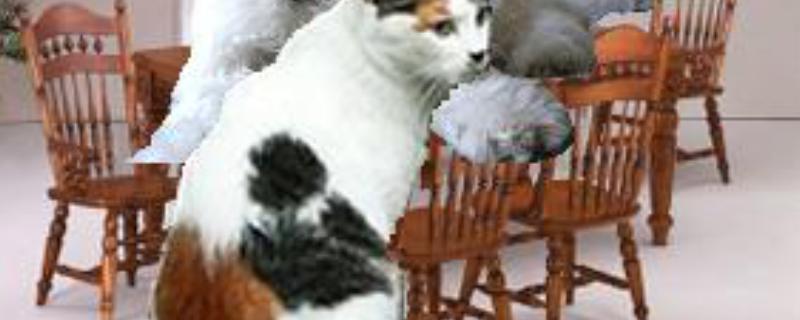}
\includegraphics[height=1.2cm]{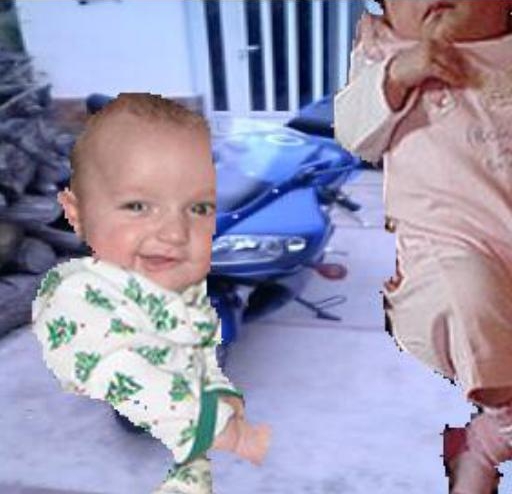}
\includegraphics[height=1.2cm]{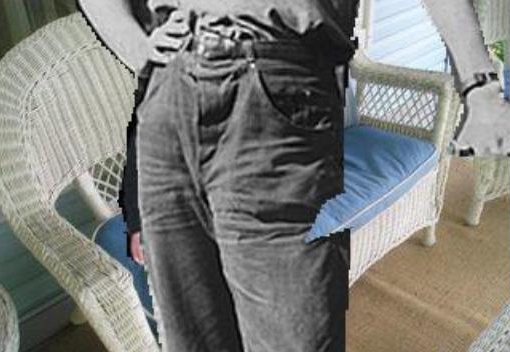}
\includegraphics[height=1.2cm]{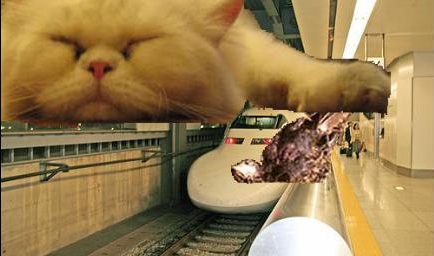}
\includegraphics[height=1.2cm]{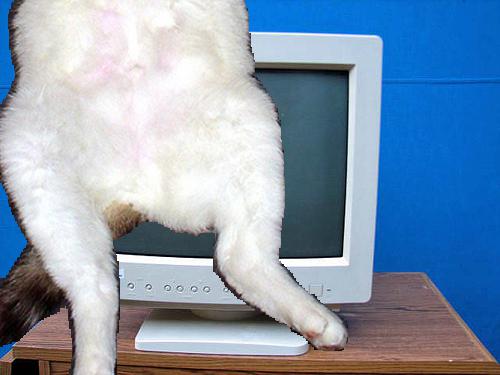}
\caption{L2}
\end{subfigure}\\[1cm]
\begin{subfigure}{\textwidth}
\centering
\includegraphics[height=1.2cm]{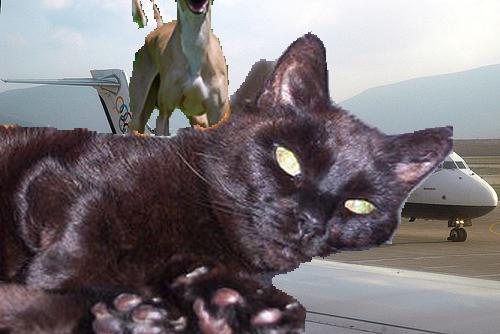}
\includegraphics[height=1.2cm]{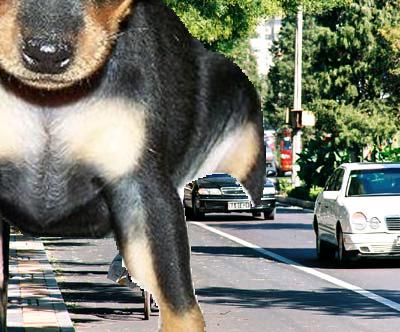}
\includegraphics[height=1.2cm]{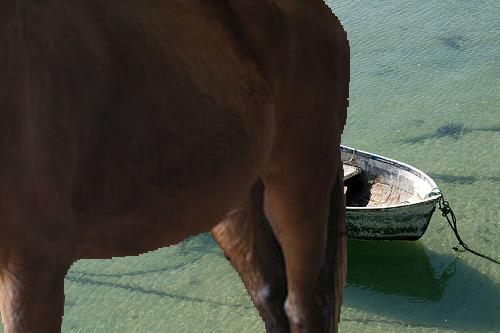}
\includegraphics[height=1.2cm]{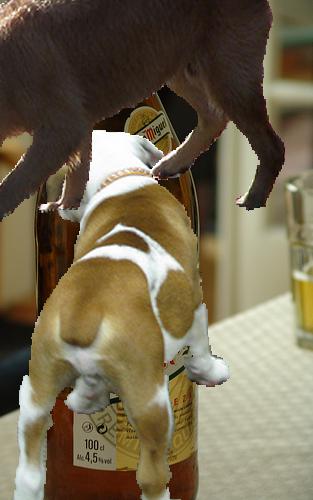}
\includegraphics[height=1.2cm]{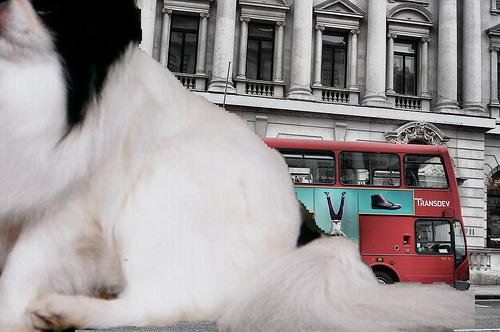}
\includegraphics[height=1.2cm]{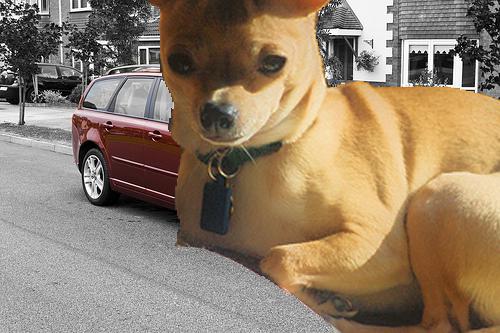}
\includegraphics[height=1.2cm]{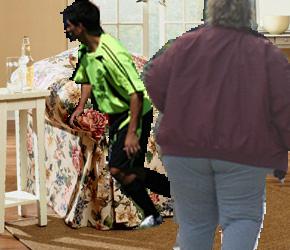}
\includegraphics[height=1.2cm]{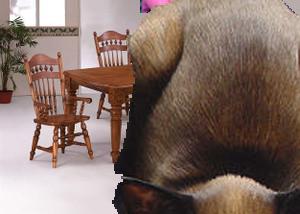}
\includegraphics[height=1.2cm]{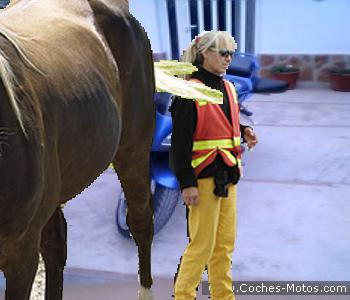}
\includegraphics[height=1.2cm]{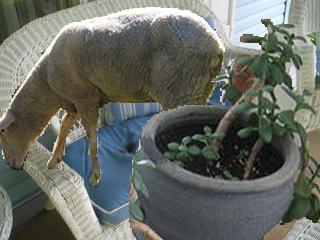}
\includegraphics[height=1.2cm]{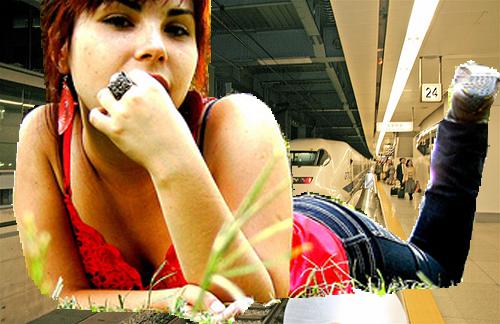}
\includegraphics[height=1.2cm]{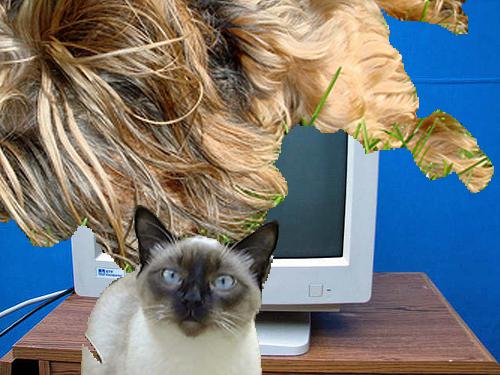}
\caption{L3}
\end{subfigure}\\[0.5cm]
\caption{Sample images from the 12 categories of PASCAL3D (L0) and OccludedPASCAL3D (L1-L3) demonstrating the level of occlusion per occlusion category.}
\label{fig:supplementary:data-overview:pascal3d:images}
\end{figure}

\begin{figure}[t]
\centering
\begin{subfigure}{\textwidth}
\centering
\includegraphics[height=0.6cm]{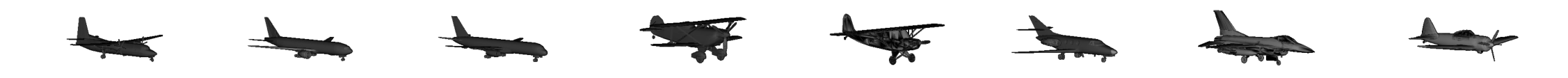}
\end{subfigure}\\[0.3cm]
\begin{subfigure}{\textwidth}
\centering
\includegraphics[height=2cm]{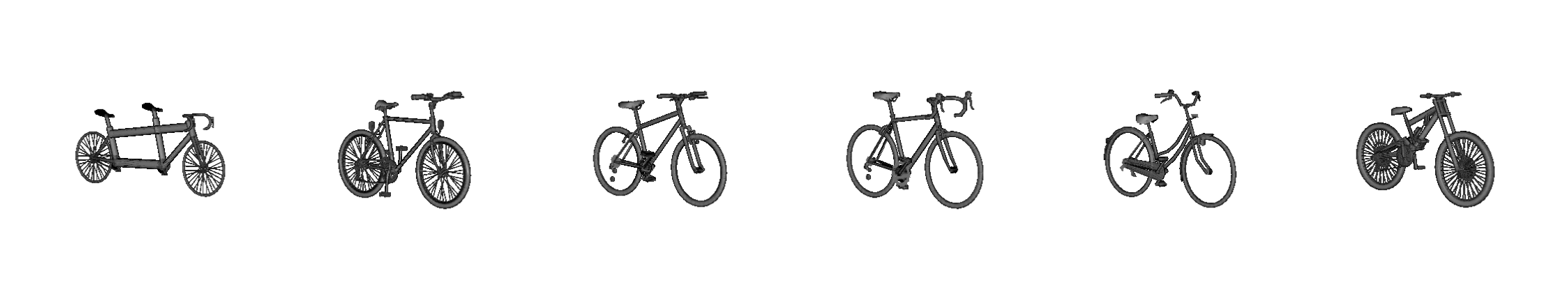}
\end{subfigure}\\[0.3cm]
\begin{subfigure}{\textwidth}
\centering
\includegraphics[height=1cm]{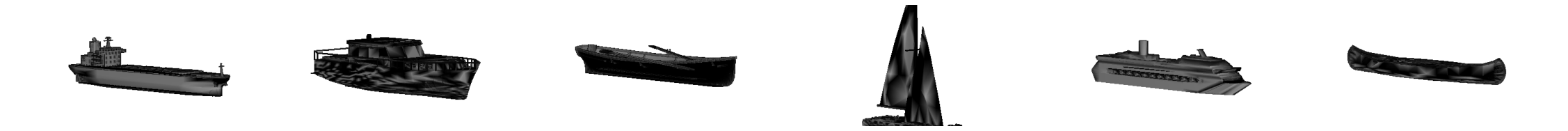}
\end{subfigure}\\[0.3cm]
\begin{subfigure}{\textwidth}
\centering
\includegraphics[height=1cm]{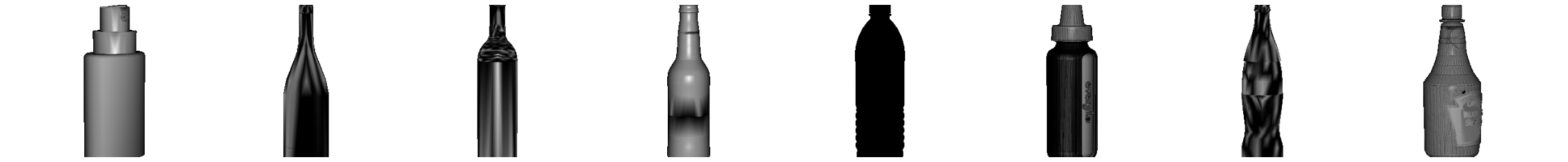}
\end{subfigure}\\[0.3cm]
\begin{subfigure}{\textwidth}
\centering
\includegraphics[height=0.8cm]{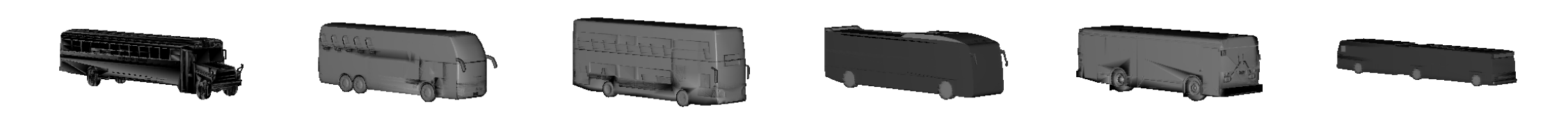}
\end{subfigure}\\[0.3cm]
\begin{subfigure}{\textwidth}
\centering
\includegraphics[height=0.6cm]{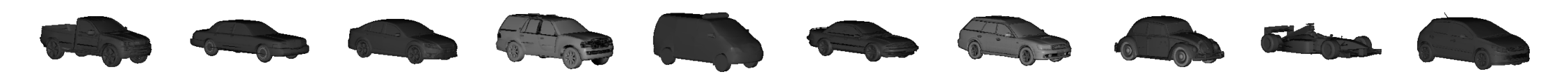}
\end{subfigure}\\[0.3cm]
\begin{subfigure}{\textwidth}
\centering
\includegraphics[height=1.6cm]{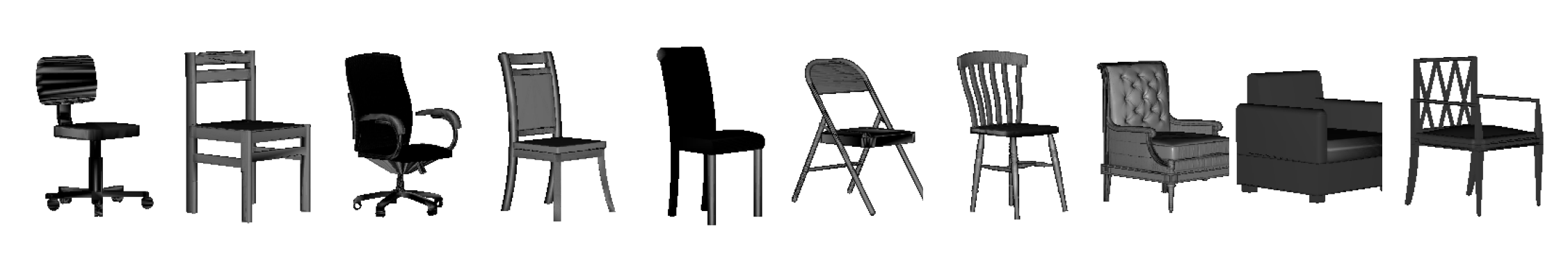}
\end{subfigure}\\[0.3cm]
\begin{subfigure}{\textwidth}
\centering
\includegraphics[height=0.8cm]{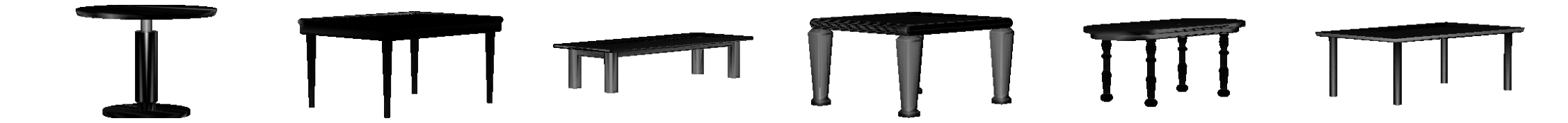}
\end{subfigure}\\[0.3cm]
\begin{subfigure}{\textwidth}
\centering
\includegraphics[height=2.2cm]{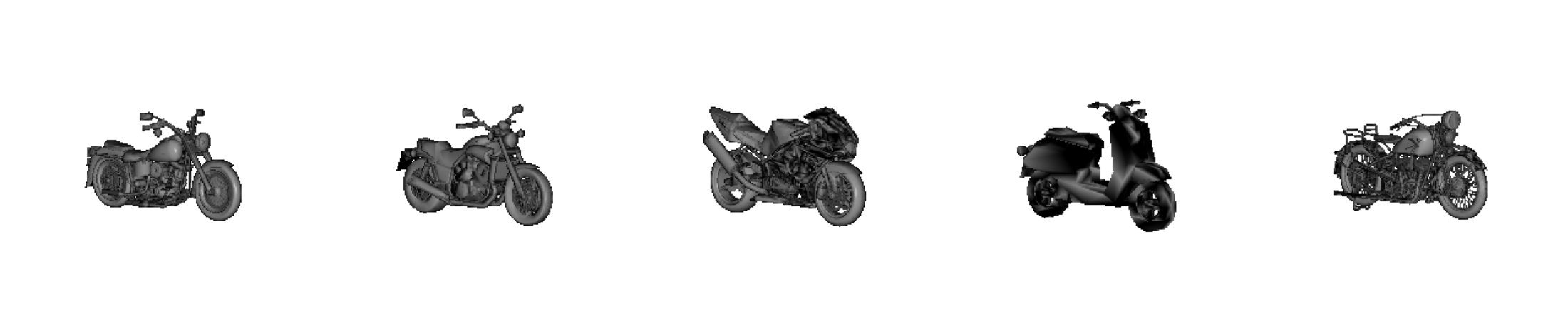}
\end{subfigure}\\[0.3cm]
\begin{subfigure}{\textwidth}
\centering
\includegraphics[height=1cm]{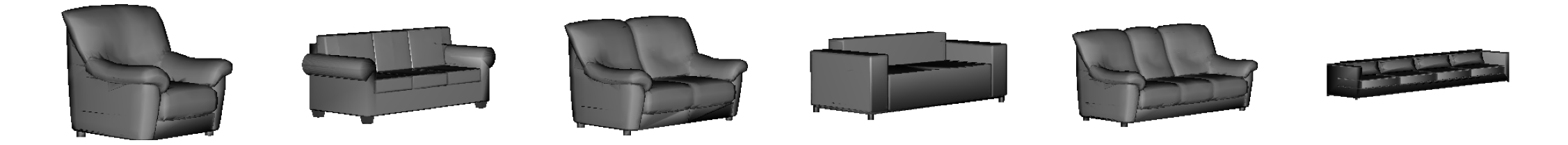}
\end{subfigure}\\[0.3cm]
\begin{subfigure}{\textwidth}
\centering
\includegraphics[height=1.4cm]{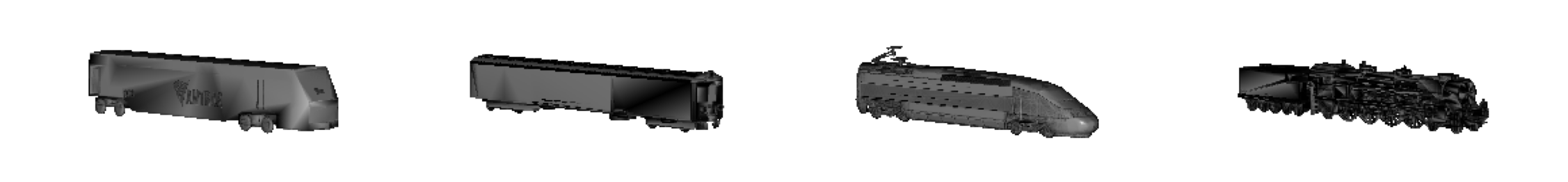}
\end{subfigure}\\[0.3cm]
\begin{subfigure}{\textwidth}
\centering
\includegraphics[height=1.6cm]{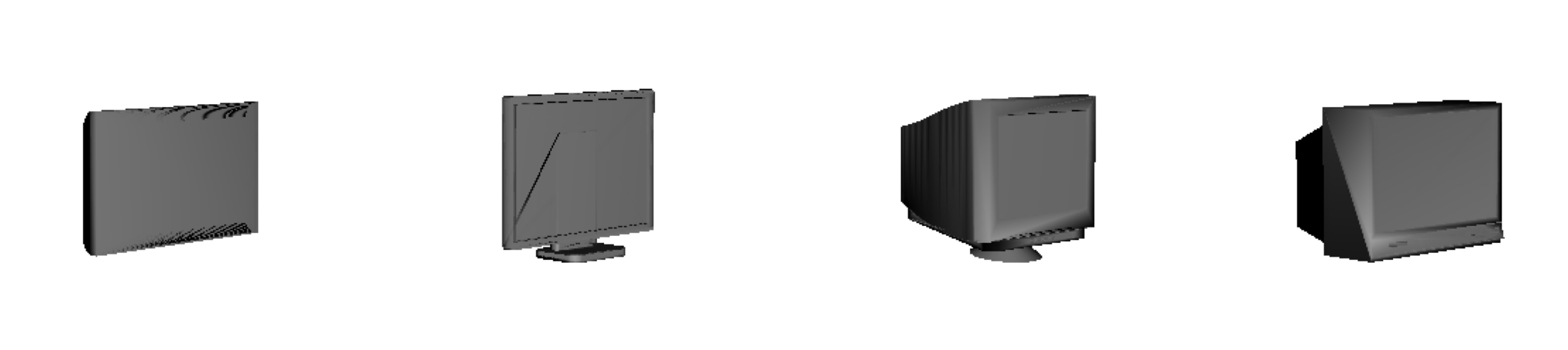}
\end{subfigure}
\caption{CAD Models per object category of PASCAL3D, from top to bottom, aeroplane, bicycle, boat, bottle, bus, car, chair, diningtable, motorbike, sofa, train, tvmonitor.}
\label{fig:supplementary:data-overview:pascal3d:cad-models}
\end{figure}

\begin{figure}[t]
\begin{subfigure}{\textwidth}
\centering
\includegraphics[width=\textwidth]{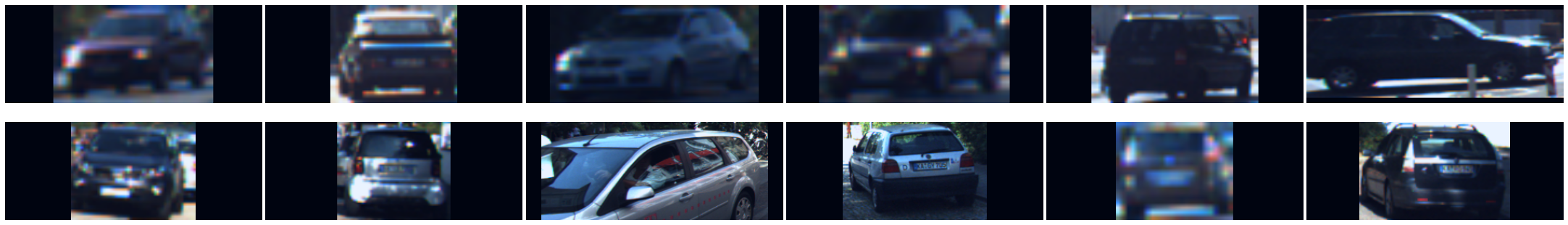}
\caption{Fully Visible}
\end{subfigure}\\
\begin{subfigure}{\textwidth}
\centering
\includegraphics[width=\textwidth]{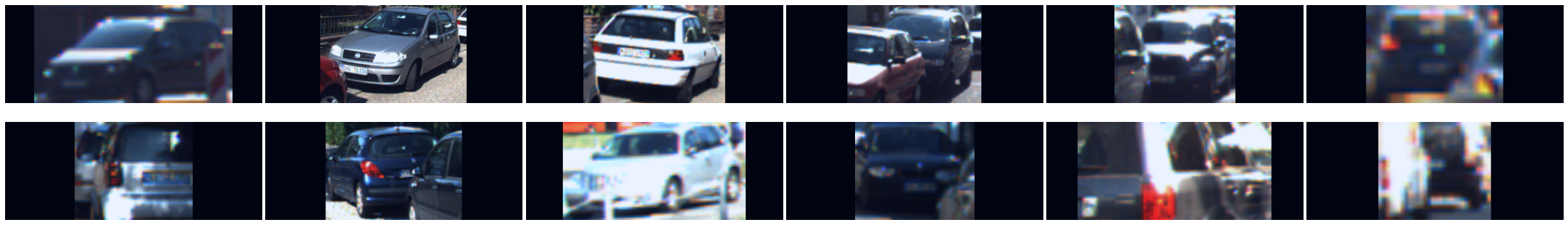}
\caption{Partly Occluded}
\end{subfigure}\\
\begin{subfigure}{\textwidth}
\centering
\includegraphics[width=\textwidth]{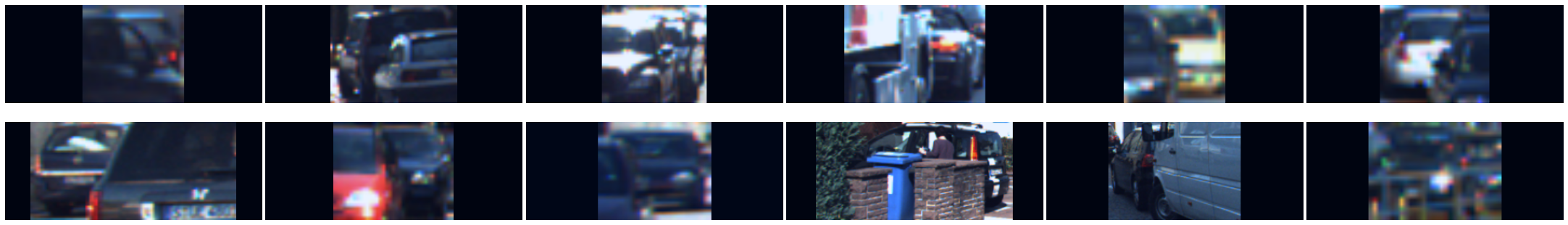}
\caption{Largely Occluded}
\end{subfigure}\\
\begin{subfigure}{\textwidth}
\centering
\end{subfigure}
\caption{Sample images from KITTI3D demonstrating the level of occlusion per occlusion category.}
\label{fig:supplementary:data-overview:kitti3d:images}
\end{figure}

In this section we present an overview of the data in PASCAL3D and OccludedPASCAL3D starting with Table~\ref{tab:supplementary:data-overview:pascal3d:data-distribution} that contains the number of samples and number of CAD models per object category and per occlusion category of PASCAL3D and OccludedPASCAL3D. Then we present some example images from all object categories and all occlusion categories in Figure~\ref{fig:supplementary:data-overview:pascal3d:images}, as well as renderings for all CAD models in Figure~\ref{fig:supplementary:data-overview:pascal3d:cad-models}. Furthermore, we include a similar overview of KITTI3D in Table~\ref{tab:supplementary:data-overview:kitti3d:data-distribution} and Figure~\ref{fig:supplementary:data-overview:kitti3d:images}.

\begin{table}[ht]
\centering
\setlength{\tabcolsep}{2.9pt}
\begin{tabular}{|c c | c c c c c c c c c c c c | c |}
\hline
\multicolumn{2}{|c|}{Samples} & aero & bike & boat & bottle & bus & car & chair & table & mbike & sofa & train & tv & Total \\
\hline
\multicolumn{2}{|c|}{Train} & 986 & 661 & 1099 & 745 & 548 & 2763 & 526 & 1160 & 624 & 642 & 662 & 629 & 11068 \\
\hline
\multirow{4}{*}{\rotatebox[origin=c]{90}{Test}} & L0 & 969 & 645 & 1059 & 747 & 532 & 2712 & 507 & 1153 & 596 & 624 & 646 & 622 & 10812 \\
& L1 & 951 & 634 & 921 & 732 & 527 & 2646 & 496 & 1145 & 578 & 542 & 633 & 616 & 10421 \\
& L2 & 937 & 619 & 905 & 725 & 525 & 2612 & 483 & 1107 & 570 & 521 & 613 & 602 & 10219 \\
& L3 & 903 & 601 & 887 & 720 & 521 & 2573 & 472 & 1075 & 555 & 507 & 583 & 586 & 9983 \\
\hline\hline
\multicolumn{2}{|c|}{CADs} & 8 & 6 & 6 & 8 & 6 & 10 & 10 & 6 & 5 & 6 & 4 & 4 & 79 \\
\hline
\end{tabular}
\caption{Number of samples and number of CAD models per object category of test sets L0-L3 from PASCAL3D and OccludedPASCAL3D}
\label{tab:supplementary:data-overview:pascal3d:data-distribution}
\end{table}

\begin{table}[ht]
\centering
\setlength{\tabcolsep}{6pt}
\begin{tabular}{|l | c | c c c c |}
\hline
\multirow{2}{*}{Split} & \multirow{2}{*}{Frames} & \multicolumn{4}{|c|}{Object Instances by Occlusion Level} \\
& & FullyVis. & PartlyOcc. & MostlyOcc. & Total \\
\hline
Train & 3712 & 3036 & 1861 & 1541 & 6438 \\
Val & 3769 & 2895 & 1950 & 2035 & 6880 \\
Test & 7518 & - & - & - & -\\
\hline
\end{tabular}
\caption{Distribution of KITTI3D car instances (w,h > 40 px)}
\label{tab:supplementary:data-overview:kitti3d:data-distribution}
\end{table}

\section{Evaluation of Rendering Types} \label{appendix:rendering-types}
We evaluated five different rendering types, namely, RGB renderings, silhouettes, depth, normals, and triplets (concatenations of RGB, depth, and normals). According to the results in Table~\ref{tab:experiments:renderings}, we observe that surface normals are more robust to higher levels of occlusions, although, our approach achieves satisfactory performance in all cases. Overall, by observing the standard deviation, all renderings seem to perform similarly in $L0$, but as the difficulty increases towards $L3$ the performance varies more widely making the superiority of surface normal maps more easily observed.

\begin{table*}[t]
\setlength{\tabcolsep}{4.6pt}
\centering
\begin{tabular}{| l |  c c c c | c c c c | c c c c |}
\hline
 & \multicolumn{4}{c|}{$ACC_\frac{\pi}{6} \uparrow$} & \multicolumn{4}{c|}{$ACC_\frac{\pi}{18} \uparrow$} & \multicolumn{4}{c|}{$MedErr \downarrow$} \\
Rendering & L0 & L1 & L2 & L3 & L0 & L1 & L2 & L3 & L0 & L1 & L2 & L3\\
\hline
RGB & \textbf{99.3} & \textbf{97.7} & 90.1 & 67.7 & 95.9 & 86.6 & 64.1 & 30.4 & \textbf{3.0} & 4.5 & 7.5 & 17.1 \\
Silhouette & 99.2 & 97.1 & 89.0 & 62.0 & \textbf{96.3} & 86.7 & 65.5 & 30.0 & 3.2 & 4.4 & 7.1 & 18.4 \\
Depth & 99.1 & 97.4 & 90.0 & 64.9 & 96.0 & 87.0 & 62.2 & 25.1 & \textbf{3.0} & 4.4 & 7.1 & 18.4 \\
Normals & 99.2 & 97.4 & \textbf{91.5} & \textbf{69.6} & 95.9 & \textbf{88.3} & \textbf{68.8} & \textbf{32.6} & 3.1 & \textbf{4.2} & \textbf{6.7} & \textbf{16.1} \\
Triplet & 99.2 & 97.2 & 90.8 & 67.9 & 96.0 & 86.8 & 64.3 & 27.9 & 3.1 & 4.3 & 7.5 & 18.0 \\
\hline
\end{tabular}
\caption{Comparison of rendering types on cars of PASCAL3D L0-L3}
\label{tab:experiments:renderings}
\end{table*}

\section{Results per Object Category} \label{appendix:results}

In Tables \ref{tab:supplementary:baselines_per_cat_l0}-\ref{tab:supplementary:baselines_per_cat_l3}, we present the performance of our approach per object category against the competing methods, StarMap and NeMo, on PASCAL3D (L0) and OccludedPASCAL3D (L1-L3). To compute the average across all object categories we use a weighted average, where the weight of each category is its number of samples divided by the total number of samples, as shown in Table \ref{tab:supplementary:data-overview:pascal3d:data-distribution}. As observed in the aforementioned tables, we outperform the state-of-the-art methods in the vast majority of object categories and occlusion levels. In Table~\ref{tab:supplementary:results:consistency}, we demonstrate the consistent performance of our approach across five models, trained from scratch on PASCAL3D with surface normal renderings. The standard deviation is less than $3\%$ in all metrics and all five models outperform the state-of-the-art.

\begin{table*}[t]
\setlength{\tabcolsep}{2.2pt}
\centering
\begin{tabular}{|c l | c c c c c c c c c c c c | c |}
\hline
& & aero & bike & boat & bottle & bus & car & chair & table & mbike & sofa & train & tv & Mean\\
\hline
\multirow{5}{*}{\rotatebox[origin=c]{90}{$ACC_\frac{\pi}{6}$}} 
& Res50-A & 83.0 & 79.6 & 73.1 & 87.9 & 96.8 & 95.5 & 91.1 & 82.0 & 80.7 & 97.0 & 94.9 & 83.3 & 88.1 \\
& Res50-S & 79.5 & 75.8 & 73.5 & 90.3 & 93.5 & 95.6 & 89.1 & 82.4 & 79.7 & 96.3 & 96.0 & 84.6 & 87.6 \\
& StarMap & 85.5 & 84.4 & 65.0 & \textbf{93.0} & 98.0 & 97.8 & 94.4 & 82.7 & 85.3 & \textbf{97.5} & 93.8 & 89.4 & 88.1 \\
& NeMo & 73.3 & 66.4 & 65.5 & 83.0 & 87.4 & 98.8 & 82.8 & 81.9 & 74.6 & 94.7 & 87.0 & 85.5 & 84.1 \\
& NeMo-M & 76.9 & 82.2 & 66.5 & 87.1 & 93.0 & 98.0 & 90.1 & 80.5 & 81.8 & 96.0 & 89.3 & 87.1 & 86.7 \\
& NeMo-S & 82.2 & 78.4 & 68.1 & 88.0 & 91.7 & 98.2 & 87.0 & 76.9 & 85.0 & 95.0 & 83.0 & 82.2 & 86.1 \\
& PoseCon. & 83.7 & 84.0 & \textbf{82.5} & 88.9 & 97.7 & 96.7 & 95.3 & 86.9 & \textbf{87.2} & 97.1 & 96.7 & 87.8 & 90.8 \\
& Ours & \textbf{84.4} & \textbf{88.1} & \textbf{82.5} & 91.7 & \textbf{98.7} & \textbf{99.2} & \textbf{95.9} & \textbf{88.8} & 85.6 & 97.0 & \textbf{98.0} & \textbf{90.0} & \textbf{92.3}\\
\hline
\multirow{5}{*}{\rotatebox[origin=c]{90}{$ACC_\frac{\pi}{18}$}} 
& Res50-A & 31.3 & 25.7 & 23.9 & 35.9 & 67.2 & 63.5 & 37.0 & 40.2 & 18.9 & 62.5 & 51.2 & 24.9 & 44.6 \\
& Res50-S & 29.1 & 22.9 & 25.3 & 39.0 & 62.7 & 62.9 & 37.5 & 42.0 & 19.5 & 57.5 & 50.2  & 25.4 & 43.9 \\
& StarMap & 49.8 & 34.2 & 25.4 & 56.8 & 90.3 & 81.9 & 67.1 & 57.5 & 27.7 & \textbf{70.3} & 69.7 & 40.0 & 59.5\\
& NeMo & 39.0 & 31.3 & 29.6 & 38.6 & 83.1 & 94.8 & 46.9 & 58.1 & 29.3 & 61.1 & 71.1 & \textbf{66.4} & 60.4\\
& NeMo-M & 43.1 & 35.3 & 36.4 & 48.6 & 89.7 & 95.5 & 49.5& 56.5 & 33.8 & 68.8 & 75.9 & 56.8 & 63.2\\
& NeMo-S & 49.7 & 29.5 & 37.7 & 49.3 & 89.3 & 94.7 & 49.5 & 52.9 & 29.0 & 58.5 & 70.1 & 42.4 & 61.1\\
& PoseCon. & 53.3 & 40.0 & 50.0 & 56.1 & 93.2 & 88.6 & 67.3 & 71.7 & \textbf{37.7} & 64.7 & 82.5 & 50.2 & 67.2 \\
& Ours & \textbf{59.5} & \textbf{42.8} & \textbf{54.2} & \textbf{68.7} & \textbf{94.5} & \textbf{95.9} & \textbf{70.4} & \textbf{71.8} & 33.9 & 69.9 & \textbf{88.7} & 58.7 & \textbf{72.2}\\
\hline
\multirow{5}{*}{\rotatebox[origin=c]{90}{$MedErr$}}
& Res50-A & 13.3 & 15.9 & 15.6 & 12.1 & 8.9 & 8.8 & 11.5 & 11.4 & 16.6 & 8.7 & 9.9 & 15.8 & 11.7 \\
& Res50-S & 14.2 & 17.3 & 15.4 & 11.7 & 9.0 & 8.8 & 12.0 & 11.0 & 17.1 & 9.2 & 10.0 & 14.9 & 11.8 \\
& StarMap & 10.0 & 14.0 & 19.7 & 8.8 & 3.2 & 4.2 & 6.9 & 8.5 & 14.5 & 6.8 & 6.7 & 12.1 & 9.0\\
& NeMo & 13.8 & 17.5 & 18.3 & 12.8 & 3.4 & 2.7 & 10.7 & 8.2 & 16.1 & 8.0 & 5.6 & \textbf{6.6} & 9.3\\
& NeMo-M & 11.8 & 13.4 & 14.8 & 10.2 & \textbf{2.6} & \textbf{2.8} & 10.1 & 8.8 & 14.0 & 7.0 & 5.0 & 8.1 & 8.2\\
& NeMo-S & 10.1 & 16.3 & 14.9 & 10.2 & 3.2 & 3.2 & 10.1 & 9.3 & 14.1 & 8.6 & 5.4 & 12.2 & 8.8\\
& PoseCon. & 9.3 & 12.0 & 10.0 & 8.8 & 3.1 & 3.5 & 7.1 & \textbf{6.0} & \textbf{12.2} & 7.8 & 4.8 & 9.9 & 7.1 \\
& Ours & \textbf{8.2} & \textbf{11.6} & \textbf{9.4} & \textbf{7.1} & 3.0 & 3.1 & \textbf{6.7} & 6.3 & 13.5 & \textbf{6.5} & \textbf{3.9} & 8.4 & \textbf{6.6} \\
\hline
\end{tabular}
\caption{Comparison with competing methods on unoccluded PASCAL3D (L0) per category}
\label{tab:supplementary:baselines_per_cat_l0}
\end{table*}

\begin{table*}[t]
\setlength{\tabcolsep}{2.2pt}
\centering
\begin{tabular}{|c l |c c c c c c c c c c c c | c |}
\hline
& & aero & bike & boat & bottle & bus & car & chair & table & mbike & sofa & train & tv & Mean\\
\hline
\multirow{5}{*}{\rotatebox[origin=c]{90}{$ACC_\frac{\pi}{6}$}}
& Res50-A & 57.3 & 56.8 & 51.4 & 78.3 & 82.5 & 80.0 & 62.3 & 63.1 & 61.1 & 84.9 & 87.8 & 69.8 & 70.4 \\
& Res50-S & 54.0 & 59.5 & 48.9 & 84.4 & 86.1 & 84.4 & 67.1 & 64.9 & 65.9 & 87.8 & 92.4 & 74.5 & 73.2 \\
& StarMap & 52.6 & 65.3 & 42.0 & 81.8 & \textbf{87.9} & 86.1 & 64.5 & 66.5 & 62.8 & 76.9 & 85.2 & 59.7 & 71.1\\
& NeMo & 49.0 & 51.4 & 52.9 & 73.5 & 82.2 & 94.3 & 70.2 & 67.9 & 53.8 & 86.7 & 75.0 & 79.4 & 73.1 \\
& NeMo-M & 58.1 & 68.8 & 53.4 & 78.8 & 86.9 & 94.0 & 76.0 & 70.0 & 61.8 & 87.3 & 82.8 & 82.8 & 77.2\\
& NeMo-S & 61.9 & 63.4 & 52.9 & 81.3 & 84.8 & 92.7 & 78.4 & 68.2 & 68.9 & 87.1 & 80.3 & 76.9 & 76.0\\
& PoseCon & 57.7 & 66.6 & 56.9 & \textbf{86.7} & 87.1 & 83.6 & 66.9 & 74.2 & 72.3 & 90.6 & 89.4 & 78.2 & 76.2 \\
& Ours & \textbf{71.4} & \textbf{79.2} & \textbf{70.6} & 85.2 & 87.7 & \textbf{97.4} & \textbf{87.2} & \textbf{81.9} & \textbf{78.4} & \textbf{94.1} & \textbf{96.5} & \textbf{80.0} & \textbf{85.7}\\
\hline
\multirow{5}{*}{\rotatebox[origin=c]{90}{$ACC_\frac{\pi}{18}$}}
& Res50-A & 11.8 & 12.5 & 12.3 & 26.5 & 45.0 & 40.7 & 14.7 & 22.3 & 10.7 & 24.4 & 34.9 & 13.0 & 25.3 \\
& Res50-S & 12.4 & 10.7 & 13.8 & 30.2 & 46.9 & 44.8 & 21.2 & 24.0 & 10.4 & 28.0 & 40.6 & 17.9 & 28.1 \\
& StarMap & 15.6 & 15.1 & 10.8 & 36.2 & 66.6 & 58.1 & 26.6 & 32.0 & 14.4 & 23.8 & 47.4 & 13.0 & 34.4\\
& NeMo & 18.5 & 19.9 & 19.1 & 24.0 & 72.1 & 82.0 & 25.8 & 35.7 & 12.6 & 44.3 & 54.0 & 49.0 & 45.1\\
& NeMo-M & 25.4 & 23.3 & 22.9 & 36.7 & \textbf{86.9} & 84.8 & 33.1 & 36.8 & 20.8 & 46.5 & 61.0 & \textbf{46.3} & 49.9\\
& NeMo-S & 29.3 & 18.0 & 24.3 & 41.5 & 76.1 & 80.5 & 27.2 & 31.4 & 19.4 & 39.9 & 55.1 & 32.0 & 46.3\\
& PoseCon. & 23.4 & 24.9 & 27.7 & 48.8 & 73.1 & 67.4 & 30.8 & 48.5 & 21.8 & 45.2 & 65.7 & 29.2 & 46.4 \\
& Ours & \textbf{35.1} & \textbf{25.9} & \textbf{34.7} & \textbf{51.8} & 74.4 & \textbf{88.3} & \textbf{44.4} & \textbf{53.1} & \textbf{23.9} & \textbf{59.0} & \textbf{81.4} & 29.9 & \textbf{56.7}\\
\hline
\multirow{5}{*}{\rotatebox[origin=c]{90}{$MedErr$}}
& Res50-A & 25.3 & 24.5 & 29.0 & 14.9 & 10.6 & 11.2 & 22.4 & 18.1 & 23.3 & 15.5 & 11.7 & 21.1 & 17.9\\
& Res50-S & 26.8 & 23.7 & 31.0 & 13.8 & 10.5 & 10.6 & 18.2 & 16.7 & 21.8 & 13.6 & 10.9 & 19.3 & 17.3 \\
& StarMap & 27.3 & 22.1 & 38.9 & 12.9 & 7.0 & 8.2 & 19.1 & 17.2 & 21.7 & 16.8 & 10.6 & 24.1 & 17.6\\
& NeMo & 30.8 & 29.0 & 27.3 & 17.6 & 5.9 & 5.1 & 18.6 & 14.7 & 27.4 & 11.3 & 8.8 & \textbf{10.2} & 15.6\\
& NeMo-M & 22.6 & 18.6 & 25.8 & 14.1 & \textbf{4.7} & 4.6 & 15.1 & 13.8 & 21.2 & 11.0 & 8.0 & 11.3 & 13.0\\
& NeMo-S & 18.9 & 23.2 & 26.7 & 12.6 & 5.2 & 5.4 & 15.6 & 15.4 & 20.1 & 12.1 & 8.6 & 15.3 & 13.6\\
& PoseCon. & 22.0 & 18.0 & 21.6 & 10.2 & 6.0 & 6.6 & 16.8 & 10.4 & 18.1 & 10.9 & 7.2 & 16.7 & 12.6 \\
& Ours & \textbf{14.3} & \textbf{15.0} & \textbf{15.2} & \textbf{9.7} & 4.8 & \textbf{4.2} & \textbf{11.3} & \textbf{9.3} & \textbf{17.1} & \textbf{8.5} & \textbf{5.0} & 14.8 & \textbf{9.7} \\
\hline
\end{tabular}
\caption{Comparison with competing methods on OccludedPASCAL3D L1 per category}
\label{tab:supplementary:baselines_per_cat_l1}
\end{table*}

\begin{table*}[t]
\setlength{\tabcolsep}{2.2pt}
\centering
\begin{tabular}{|c l |c c c c c c c c c c c c | c |}
\hline
& & aero & bike & boat & bottle & bus & car & chair & table & mbike & sofa & train & tv & Mean\\
\hline
\multirow{5}{*}{\rotatebox[origin=c]{90}{$ACC_\frac{\pi}{6}$}} 
& Res50-A & 33.3 & 40.2 & 33.6 & 70.6 & 69.5 & 57.0 & 41.8 & 47.4 & 43.3 & 66.8 & 80.4 & 58.1 & 52.8 \\
& Res50-S & 36.3 & 44.9 & 36.1 & 76.1 & 73.1 & 65.5 & 53.2 & 49.5 & 45.4 & 72.7 & 88.3 & 65.0 & 58.4 \\
& StarMap & 28.5 & 38.9 & 21.3 & 65.0 & 61.7 & 59.3 & 37.5 & 44.7 & 43.2 & 55.1 & 56.4 & 36.2 & 47.2\\
& NeMo & 38.2 & 41.2 & 39.6 & 58.3 & 72.6 & 84.7 & 50.7 & 51.1 & 34.9 & 70.1 & 60.0 & 64.6 & 59.9\\
& NeMo-M & 43.1 & \textbf{55.7} & 43.3 & 69.1 &\textbf{ 79.8} & 84.5 & 58.8 & 58.4 & 43.9 & 76.4 & 64.3 & \textbf{70.3} & 65.2\\
& NeMo-S & 43.4 & 49.6 & 43.6 & 76.0 & 71.2 & 83.8 & 61.9 & 55.9 & 50.9 & 78.3 & 63.1 & 68.6 & 63.9\\
& PoseCon. & 38.5 & 51.2 & 39.2 & \textbf{81.8} & 69.5 & 61.8 & 49.3 & 57.6 & 56.1 & 74.1 & 82.4 & 61.0 & 59.3\\
& Ours & \textbf{54.6} & 54.6 & \textbf{55.4} & 68.8 & 71.0 & \textbf{91.5} & \textbf{66.5} & \textbf{67.8} &\textbf{ 57.9} & \textbf{84.4} & \textbf{93.1} & 67.3 & \textbf{72.7} \\
\hline
\multirow{5}{*}{\rotatebox[origin=c]{90}{$ACC_\frac{\pi}{18}$}}
& Res50-A & 6.1 & 4.5 & 7.2 & 20.1 & 25.9 & 21.4 & 9.5 & 13.2 & 6.1 & 14.0 & 23.0 & 8.6 & 14.5 \\
& Res50-S & 5.7 & 6.9 & 8.0 & 25.5 & 33.9 & 29.1 & 13.0 & 11.6 & 6.8 & 18.4 & 32.0 & 13.8 & 18.6 \\
& StarMap & 3.8 & 5.8 & 2.4 & 19.7 & 30.5 & 24.5 & 7.7 & 9.6 & 5.1 & 9.6 & 21.5 & 5.8 & 13.9\\
& NeMo & 10.7 & 10.5 & 11.3 & 13.9 & 55.8 & 60.6 & 9.3 & 20.3 & 6.3 & 26.1 & 34.6 & 32.1 & 30.2 \\
& NeMo-M & 12.8 & \textbf{16.6} & 16.8 & 21.9 & \textbf{62.3} & 64.6 & 17.2 & 20.3 & 12.3 & 32.4 & 38.2 & \textbf{32.7} & 34.5\\
& NeMo-S & 14.9 & 11.1 & 15.6 & 18.2 & 56.0 & 62.4 & 17.4 & 18.7 & 10.2 & 30.5 & 36.4 & 22.4 & 32.0 \\
& PoseCon. & 11.3 & 13.6 & 17.3 & \textbf{41.1} & 46.3 & 38.2 & 16.4 & 28.4 & 12.3 & 26.7 & 44.9 & 17.9 & 28.1 \\
& Ours & \textbf{19.0} & 14.4 & \textbf{22.3} & 32.1 & 50.5 & \textbf{68.8} & \textbf{22.2} & \textbf{29.5} & \textbf{13.3} & \textbf{35.7} & \textbf{67.5} & 17.6 & \textbf{38.9}\\
\hline
\multirow{5}{*}{\rotatebox[origin=c]{90}{$MedErr$}}
& Res50-A & 49.3 & 42.5 & 58.5 & 17.7 & 15.9 & 21.3 & 35.4 & 32.0 & 36.1 & 20.3 & 15.2 & 25.3 & 30.4 \\
& Res50-S & 45.8 & 33.9 & 52.8 & 16.3 & 12.4 & 15.1 & 27.1 & 30.9 & 32.4 & 18.3 & 12.3 & 24.1 & 26.1 \\
& StarMap & 55.2 & 37.1 & 69.1 & 20.6 & 19.0 & 21.3 & 39.2 & 34.0 & 35.5 & 27.0 & 24.8 & 40.3 & 34.1\\
& NeMo & 39.8 & 37.7 & 44.2 & 24.8 & 8.8 & 7.7 & 29.7 & 28.5 & 47.5 & 16.9 & 18.2 & 17.0 & 24.1 \\
& NeMo-M & 38.5 & \textbf{26.4} & 38.2 & 18.8 & \textbf{7.0} & 7.3 & 23.0 & 23.0 & 36.0 & 14.0 & 14.9 & \textbf{16.1} & 20.2 \\
& NeMo-S & 39.9 & 30.6 & 38.8 & 19.5 & 8.3 & 7.8 & 21.3 & 24.8 & 29.5 & 14.2 & 16.9 & 18.5 & 20.9\\
& PoseCon. & 44.8 & 29.1 & 43.8 & \textbf{12.0} & 10.7 & 14.9 & 30.3 & 20.9 & 25.5 & 16.7 & 11.3 & 24.7 & 23.1 \\
& Ours & \textbf{25.7} & 26.7 & \textbf{24.6} & 15.2 & 9.7 & \textbf{6.7} & \textbf{19.9} & \textbf{17.1} & \textbf{23.9} & \textbf{12.9} & \textbf{6.8} & 22.5 & \textbf{16} \\
\hline
\end{tabular}
\caption{Comparison with competing methods on OccludedPASCAL3D L2 per category}
\label{tab:supplementary:baselines_per_cat_l2}
\end{table*}

\begin{table*}[t]
\setlength{\tabcolsep}{2.2pt}
\centering
\begin{tabular}{|c l |c c c c c c c c c c c c | c |}
\hline
 & & aero & bike & boat & bottle & bus & car & chair & table & mbike & sofa & train & tv & Mean\\
\hline
\multirow{5}{*}{\rotatebox[origin=c]{90}{$ACC_\frac{\pi}{6}$}} 
& Res50-A & 18.3 & 20.8 & 21.2 & 62.1 & 57.0 & 36.9 & 31.1 & 32.2 & 24.3 & 56.2 & 64.5 & 53.4 & 37.8 \\
& Res50-S & 20.0 & 33.4 & 25.5 & 67.5 & 57.8 & 42.0 & 40.7 & 33.9 & 30.3 & 56.6 & 82.8 & 56.5 & 43.1 \\
& StarMap & 7.6 & 18.5 & 10.6 & 46.3 & 35.1 & 25.3 & 22.5 & 24.6 & 15.9 & 26.4 & 24.0 & 19.5 & 22.9\\
& NeMo & 24.0 & 31.3 & 27.4 & 43.3 & 48.8 & 62.8 & 31.8 & 29.7 & 18.4 & 44.2 & 34.5 & 51.4 & 41.3\\
& NeMo-M & 23.8 & \textbf{34.3} & 29.5 & 53.9 & \textbf{56.0} & 65.5 & 43.4 & 41.5 & 25.4 & 58.2 & 43.2 & \textbf{54.1} & 47.1\\
& NeMo-S & 20.6 & 33.8 & 27.6 & 61.7 & 49.9 & 61.8 & \textbf{44.7} & 41.2 & \textbf{35.3} & \textbf{62.9} & 47.9 & 50.2 & 46.8 \\
& PoseCon. & 19.2 & 30.6 & 27.4 & \textbf{73.5} & 47 & 35.2 & 33.3 & 38.0 & 33.3 & 52.1 & 70.7 & 44.4 & 39.7 \\
& Ours & \textbf{27.4} & 28.8 & \textbf{31.8} & 43.3 & 41.3 & \textbf{69.6} & 40.9 & \textbf{45.6} & 32.1 & 62.1 & \textbf{85.2} & 47.8 & \textbf{49.8}\\
\hline
\multirow{5}{*}{\rotatebox[origin=c]{90}{$ACC_\frac{\pi}{18}$}}
& Res50-A & 1.6 & 2.3 & 2.9 & 11.9 & 14.4 & 7.6 & 3.8 & 5.7 & 3.1 & 7.9 & 12.7 & 8.9 & 6.7 \\
& Res50-S & 2.0 & 5.5 & 4.8 & 16.7 & 21.1 & 13.1 & 5.9 & 5.7 & 4.3 & 9.9 & 22.5 & 6.0 & 9.9 \\
& StarMap & 0.8 & 1.7 & 1.1 & 11.8 & 8.3 & 4.8 & 2.1 & 2.6 & 1.6 & 2.8 & 5.2 & 0.7 & 3.7\\
& NeMo & 4.4 & 6.2 & 6.7 & 6.8 & 26.5 & 31.1 & 3.4 & 6.7 & 2.0 & 9.3 & 13.0 & \textbf{16.7} & 14.5 \\
& NeMo-M & 5.5 & 5.2 & 7.9 & 10.8 & \textbf{34.2} & \textbf{37.4} & 7.4 & 8.2 & 4.5 & 15.8 & 15.1 & 15.9 & 17.8 \\
& NeMo-S & 4.7 & \textbf{6.7} & 8.6 & 11.7 & 29.2 & 33.7 & \textbf{11.0} & 10.7 & 4.9 & \textbf{17.8} & 17.2 & 10.9 & 17.1\\
& PoseCon. & 4.0 & 5.3 & 7.1 & \textbf{26.8} & 18.4 & 13.0 & 9.1 & 13.0 & \textbf{5.0} & 15.2 & 27.4 & 11.4 & 12.7 \\
& Ours & \textbf{6.5} & 3.3 & \textbf{9.0} & 16.9 & 19.2 & 32.6 & 7.2 & \textbf{14.1} & 3.6 & 15.4 & \textbf{42.2} & 7.0 & \textbf{17.9}\\
\hline
\multirow{5}{*}{\rotatebox[origin=c]{90}{$MedErr$}}
& Res50-A & 69.8 & 70.9 & 73.2 & 22.7 & 24.9 & 46.7 & 41.5 & 44.4 & 59.8 & 26.3 & 21.3 & 28.4 & 46.4 \\
& Res50-S & 65.8 & 47.1 & 75.8 & 20.9 & 18.5 & 46.6 & 35.9 & 49.9 & 56.3 & 26.4 & 15.3 & 26.5 & 44.0 \\
& StarMap & 87.0 & 67.6 & 90.2 & 32.6 & 51.3 & 64.0 & 60.7 & 53.2 & 73.4 & 51.0 & 52.7 & 54.7 & 63.0\\
& NeMo & \textbf{65.3} & 48.4 & 65.2 & 34.5 & 34.9 & 17.2 & 44.6 & 55.7 & 74.3 & 33.7 & 47.6 & 29.3 & 41.8 \\
& NeMo-M & 69.8 & 49.6 & 63.0 & 28.2 & \textbf{19.4} & \textbf{14.9} & 35.4 & 39.9 & 60.0 & 23.7 & 38.1 & \textbf{27.2} & \textbf{36.1} \\
& NeMo-S & 74.8 & \textbf{46.1} & 70.1 & 24.5 & 30.2 & 16.3 & \textbf{35.2} & 37.5 & 50.5 & \textbf{21.5} & 31.7 & 29.9 & 36.5 \\
& PoseCon. & 66.8 & 61 & 64.7 & \textbf{16.6} & 34.2 & 51.5 & 41.4 & 42.5 & 50.7 & 28.7 & 16.7 & 33.3 & 45.5 \\
& Ours & 75.2 & 54.5 & \textbf{61.9} & 48.8 & 53.8 & 16.1 & 36.8 & \textbf{34.2} & \textbf{49.2} & 23.8 & \textbf{11.8} & 31.1 & 37.9\\
\hline
\end{tabular}
\caption{Comparison with competing methods on OccludedPASCAL3D L3 per category}
\label{tab:supplementary:baselines_per_cat_l3}
\end{table*}

\begin{table}[t]
\setlength{\tabcolsep}{2.0pt}
\centering
\begin{tabular}{| c | c c c c | c c c c | c c c c |}
\hline
\multirow{2}{*}{Model} & \multicolumn{4}{|c|}{$ACC_\frac{\pi}{6} \uparrow$} & \multicolumn{4}{c|}{$ACC_\frac{\pi}{18} \uparrow$} & \multicolumn{4}{c|}{$MedErr \downarrow$}\\
& L0 & L1 & L2 & L3 & L0 & L1 & L2 & L3 & L0 & L1 & L2 & L3 \\
\hline
1 & 99.2 & 97.4 & 91.5 & 69.6 & 95.9 & 89.3 & 68.8 & 32.6 & 3.1 & 4.2 & 6.7 & 16.1 \\
2 & 99.2 & 97.0 & 90.2 & 69.4 & 96.2 & 87.2 & 66.0 & 29.7 & 3.2 & 4.5 & 7.2 & 16.5 \\
3 & 99.1 & 97.6 & 90.3 & 66.5 & 95.9 & 87.7 & 66.5 & 30.0 & 3.2 & 4.3 & 7.2 & 17.4 \\
4 & 99.2 & 96.6 & 90.7 & 69.9 & 96.4 & 88.4 & 70.2 & 35.2 & 3.2 & 4.1 & 6.7 & 14.9 \\
5 & 99.2 & 97.4 & 90.2 & 69.0 & 96.0 & 86.6 & 63.0 & 29.3 & 3.4 & 4.7 & 8.0 & 17.1 \\
\hline
Mean & 99.2 & 97.2 & 90.6 & 68.9 & 96.1 & 87.6 & 66.7 & 31.4 & 3.2 & 4.4 & 7.2 & 16.4\\
SD & 0.04 & 0.4 & 0.55 & 1.37 & 0.22 & 0.76 & 2.84 & 2.51 & 0.11 & 0.24 & 0.53 & 0.98\\
\hline
\end{tabular}
\caption{Evaluation of performance consistency on PASCAL3D Cars}
\label{tab:supplementary:results:consistency}
\end{table}

\section{Evaluation of Reference Set Designs} \label{appendix:set-designs}
To decide what is the best reference set design, we trained and evaluated on three distinct reference sets, namely TrainDB, CoarseDB, and FineDB. TrainDB contains the renderings that were generated based on the poses in the traininsg set, while CoarseDB and FineDB were generated using a discretization of the viewing sphere as shown in Table~\ref{tab:supplementary:sphere-discretization}. To train more efficiently we sample renderings from CoarseDB and FineDB for every object instance in a batch. 
To run inference, we examine all possible combinations to find the optimal setting, as presented in Table~\ref{tab:supplementary:reference-sets}. Out of the three designed reference sets, we found TrainDB to achieve the highest performance thanks to it being more representative of the data while also being the fastest, since it contains the least amount of renderings. Furthermore, we attribute the performance drop of CoarseDB and FineDB to our sampling scheme that does not utilize all the samples, but only the ones close to the training data. Therefore, the models are not trained on all available renderings in the database, so they are not able to project them correctly to the embedding space, thus resulting in the lower performance, seen in Table~\ref{tab:supplementary:reference-sets}.

\begin{table}[t]
\setlength{\tabcolsep}{2pt}
\centering
\begin{tabular}{| l | c | c | c c c c | c c c c | c c c c |}
\hline
\multirow{2}{*}{Training} & \multirow{2}{*}{Inference} & \multirow{2}{*}{fps $\uparrow$} & \multicolumn{4}{c|}{$ACC_\frac{\pi}{6} \uparrow$} & \multicolumn{4}{c|}{$ACC_\frac{\pi}{18} \uparrow$} & \multicolumn{4}{c|}{$MedErr \downarrow$}\\
&&& L0 & L1 & L2 & L3 & L0 & L1 & L2 & L3 & L0 & L1 & L2 & L3 \\
\hline
TrainDB & TrainDB & 35 & 99.2 & \textbf{97.4} & \textbf{91.5} & \textbf{69.6} & \textbf{95.9} & \textbf{89.3} & \textbf{68.8} & \textbf{32.6} & \textbf{3.1} & \textbf{4.2} & \textbf{6.7} & \textbf{16.1} \\
\hline
CoarseDB & TrainDB & 35 & 99.2 & \textbf{97.4} & 89.1 & 65.4 & 95.6 & 85.2 & 60.0 & 25.6 & 3.2 & 4.7 & 8.1 & 19.9 \\ 
CoarseDB & CoarseDB & 2.5 & 99.0 & 95.7 & 84.5 & 56.0 & 91.0 & 74.0 & 44.2 & 14.4 & 4.2 & 6.1 & 11.2 & 26.2 \\
CoarseDB & Both & 0.5 &  99.0 & 95.6 & 84.5 & 56.0 & 91.6 & 74.9 & 45.2 & 14.8 & 3.5 & 5.8 & 11.1 & 26.2 \\
\hline
FineDB & TrainDB & 35 & \textbf{99.3} & 96.7 & 90.9 & 68.9 & 95.8 & 86.6 & 64.3 & 30.3 & 3.2 & 4.6 & 7.4 & 17.4 \\
FineDB & FineDB & 0.5 & 97.6 & 92.5 & 81.5 & 52.0 & 90.7 & 75.7 & 48.0 & 15.7 & 4.3 & 6.2 & 10.4 & 28.3 \\
FineDB & Both & 0.5 & 98.0 & 92.9 & 82.0 & 52.3 & 91.7 & 76.6 & 48.9 & 16.1 & 3.6 & 5.8 & 10.3 & 28.1 \\
\hline
\end{tabular}
\caption{Evaluation of the three designed reference sets}
\label{tab:supplementary:reference-sets}
\end{table}

\section{Occlusion Augmentation Results} \label{appendix:occlusion-augmentation-results}
\begin{figure}[t]
\centering
\begin{subfigure}{0.5\textwidth}
    \includegraphics[width=\textwidth]{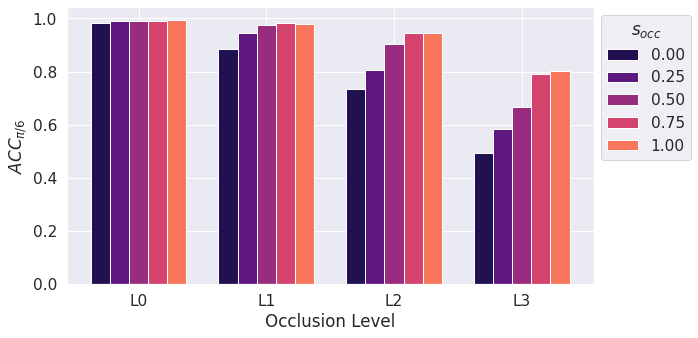}
    \caption{$ACC_\frac{\pi}{6}$}
\end{subfigure}%
\begin{subfigure}{0.5\textwidth}
    \includegraphics[width=\textwidth]{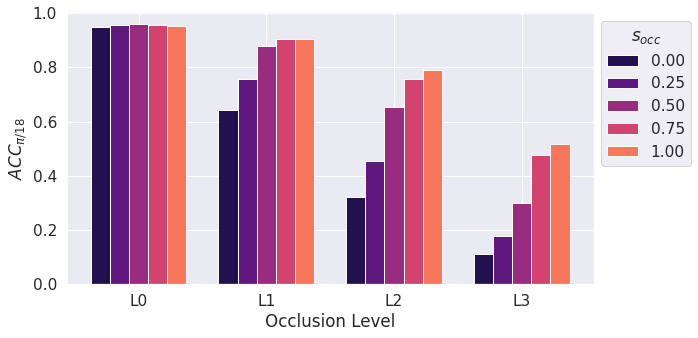}
    \caption{$ACC_\frac{\pi}{18}$}
\end{subfigure}\\[0.5cm]
\begin{subfigure}{0.5\textwidth}
    \includegraphics[width=\textwidth]{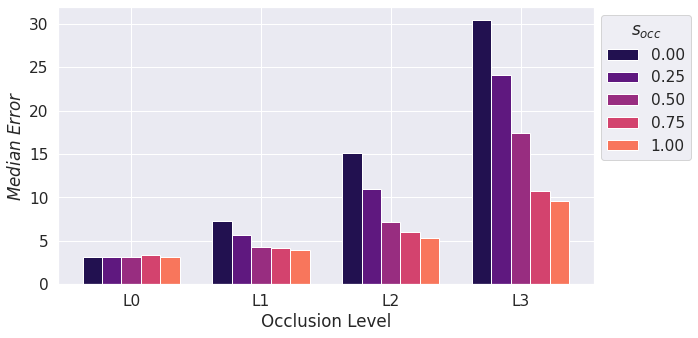}
    \caption{\textit{Median Error}}
\end{subfigure}
\caption{Evaluation of models trained on various occlusion scales $s_{occ}$ on Cars of PASCAL3D and OccludedPASCAL3D.}
\label{fig:supplementary:occlusion-augmentation}
\end{figure}

In Figure~\ref{fig:supplementary:occlusion-augmentation}, we provide graphs of all three evaluation metrics, not just $ACC_\frac{\pi}{6}$, for the experiments on our occlusion augmentation scheme. All three metrics follow the same trend indicating that the higher the occlusion scale $s_{occ}$ during training, the higher the robustness across the increasingly occluded sets L0-L3. However, since the occlusion augmentation scheme is not sophisticated enough to avoid fully occluding the object of interest, using high a occlusion scale can actually make training harder and require many more epochs to converge. Due to this effect, we observe a slight drop in performance on L0, which could be alleviated by increasing the number of training epochs or by progressively decreasing the occlusion scale. 

\begin{table}[t]
\setlength{\tabcolsep}{4pt}
\begin{center}
\begin{tabular}{ l c c c c c}
\hline
Set &  Azimuth & Elevation & In-plane Rotation & Renderings \\
\hline
TrainDB & - & - & - & 2.7k \\
CoarseDB & $5^\circ$ & $5^\circ$ & $5^\circ$ & 178k \\
FineDB & $1^\circ$ & $5^\circ$ & $5^\circ$ & 889k \\
\hline
Range & $0^\circ$:$360^\circ$ & $-30^\circ$:$60^\circ$ & $-30^\circ$:$30^\circ$ & - \\
\hline
\end{tabular}
\end{center}
\caption{Discretization of pose space per rendering set}
\label{tab:supplementary:sphere-discretization}
\end{table}

\section{Inference Speed} \label{appendix:inference-speed}
Our approach can run inference on an object instance in approximately 30ms. Out of those, roughly $60\%$ are spent on embedding the object instance and the rest $40\%$ are spent on calculating the distances and finding the nearest neighbour. Further increasing the inference speed can be accomplished by training a smaller backbone (e.g. ResNet18), using an embedding size lower than $512$, removing very similar poses from TrainDB, or using renderings of only one CAD model. In case of a larger database, in particular, it would also be beneficial to employ a kd-tree to speed up the nearest neighbour search. This delay can further be reduced or even eliminated by incorporating a regression head and a regression loss term.

\section{Bounding Box Augmentation Results} \label{appendix:bbox-augmentation-results}
\begin{figure}[t]
\centering
\begin{subfigure}{0.5\textwidth}
    \includegraphics[width=\textwidth]{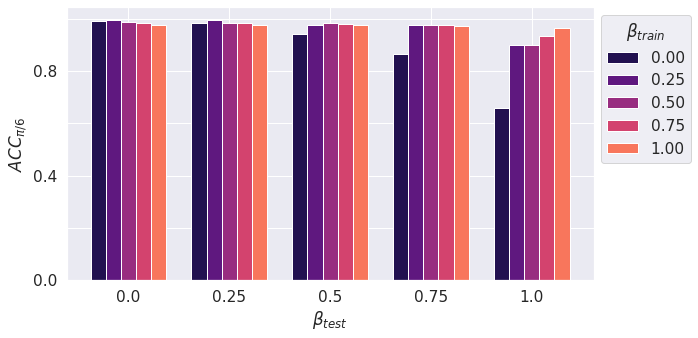}
    \caption{$ACC_\frac{\pi}{6}$}
\end{subfigure}%
\begin{subfigure}{0.5\textwidth}
    \includegraphics[width=\textwidth]{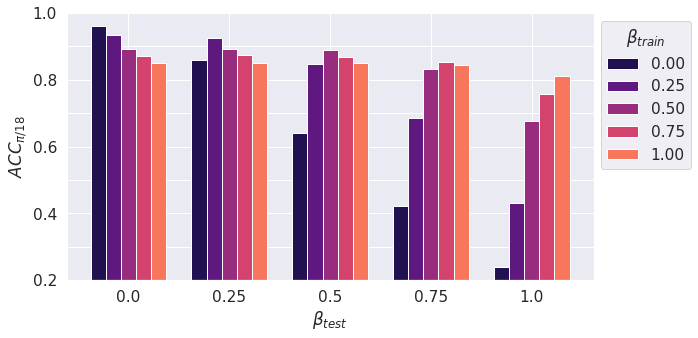}
    \caption{$ACC_\frac{\pi}{18}$}
\end{subfigure}\\[0.5cm]
\begin{subfigure}{0.5\textwidth}
    \includegraphics[width=\textwidth]{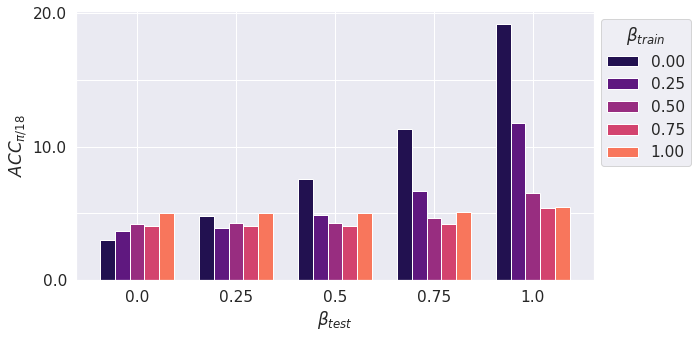}
    \caption{\textit{Median Error}}
\end{subfigure}
\caption{Evaluation of models trained on various levels $\beta_{train}$ of bounding box augmentation on PASCAL3D L0 Cars with various levels $\beta_{test}$ of test-time bounding box augmentation.}
\label{fig:supplementary:bbox-augmentation}
\end{figure}

In Figure~\ref{fig:supplementary:bbox-augmentation}, we provide graphs of all three evaluation metrics for the experiments on our bounding box augmentation scheme. The three metrics follow a similar trend which denotes that training with higher bounding box noise results in more robustness to test time noise. However, training with too high $\beta_{train}$ can result in a performance drop in cases of minimal test time noise, which could be potentially alleviated by training for more epochs. Nevertheless, training with $\beta_{train} \in [0.1, 0.25]$ seems a good trade-off between robustness to noise and accuracy in absence of noise. Better cross-dataset performance can be achieved by training with higher bounding box noise which leads to better generalization to cases without perfect center/scale alignment. For example training with $\beta_{train}=0.75$ on PASCAL3D lead to better cross-dataset performance on KITTI3D.

\section{Qualitative Results} \label{appendix:qualitative-results}
\begin{figure}[t]
\begin{subfigure}{\textwidth}
\centering
\includegraphics[width=\textwidth]{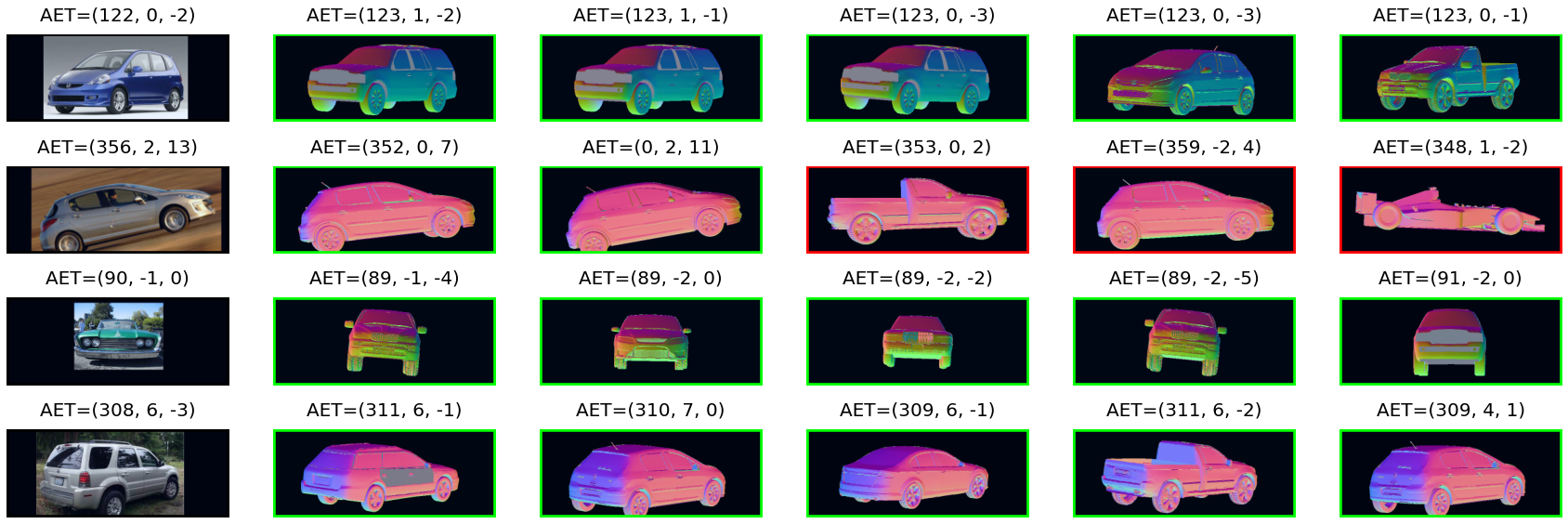}
\caption{L0}
\end{subfigure}\\[-0.1cm]
\begin{subfigure}{\textwidth}
\centering
\includegraphics[width=\textwidth]{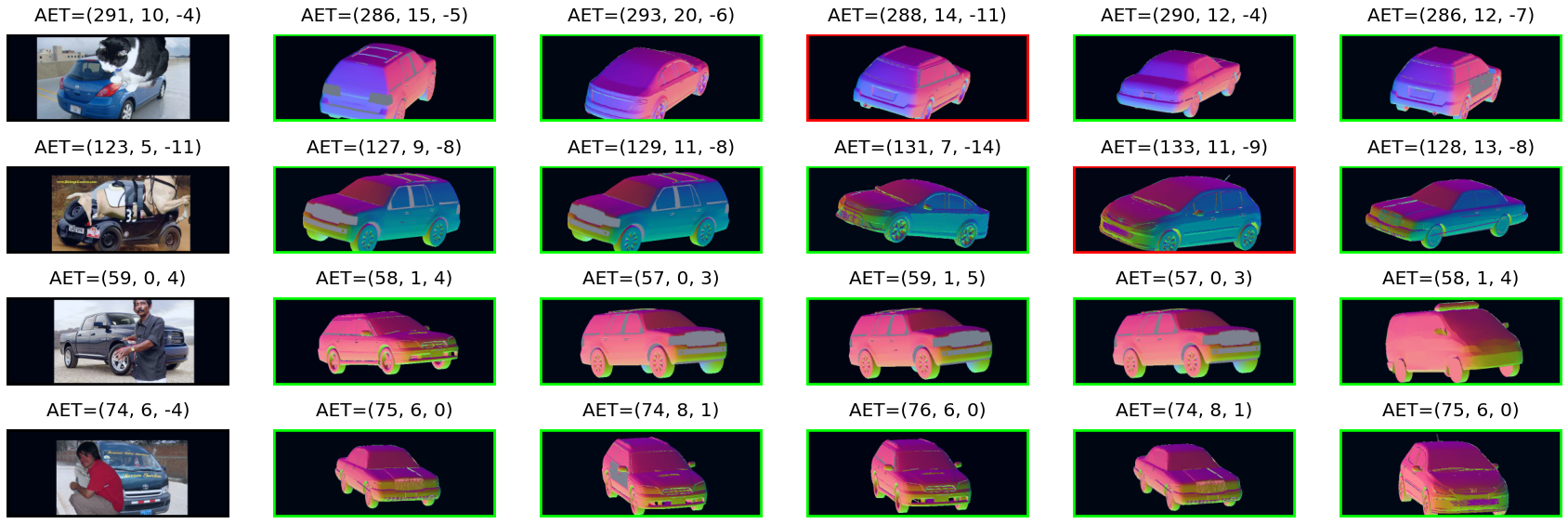}
\caption{L1}
\end{subfigure}\\[-0.1cm]
\begin{subfigure}{\textwidth}
\centering
\includegraphics[width=\textwidth]{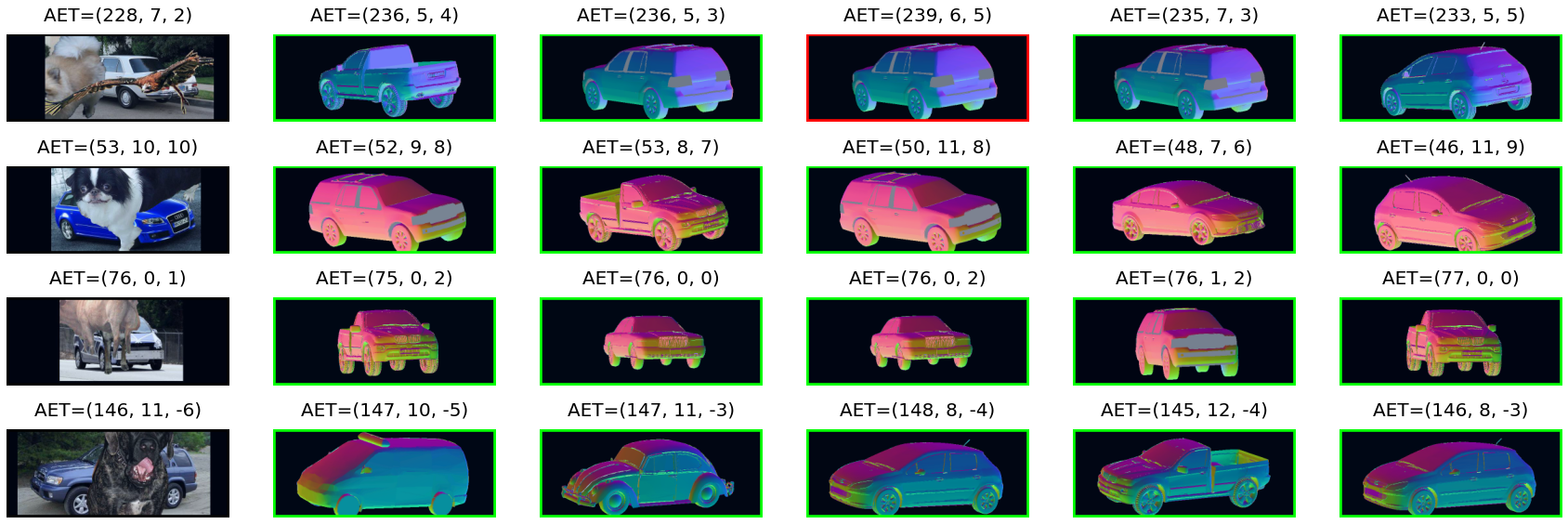}
\caption{L2}
\end{subfigure}\\[-0.1cm]
\begin{subfigure}{\textwidth}
\centering
\includegraphics[width=\textwidth]{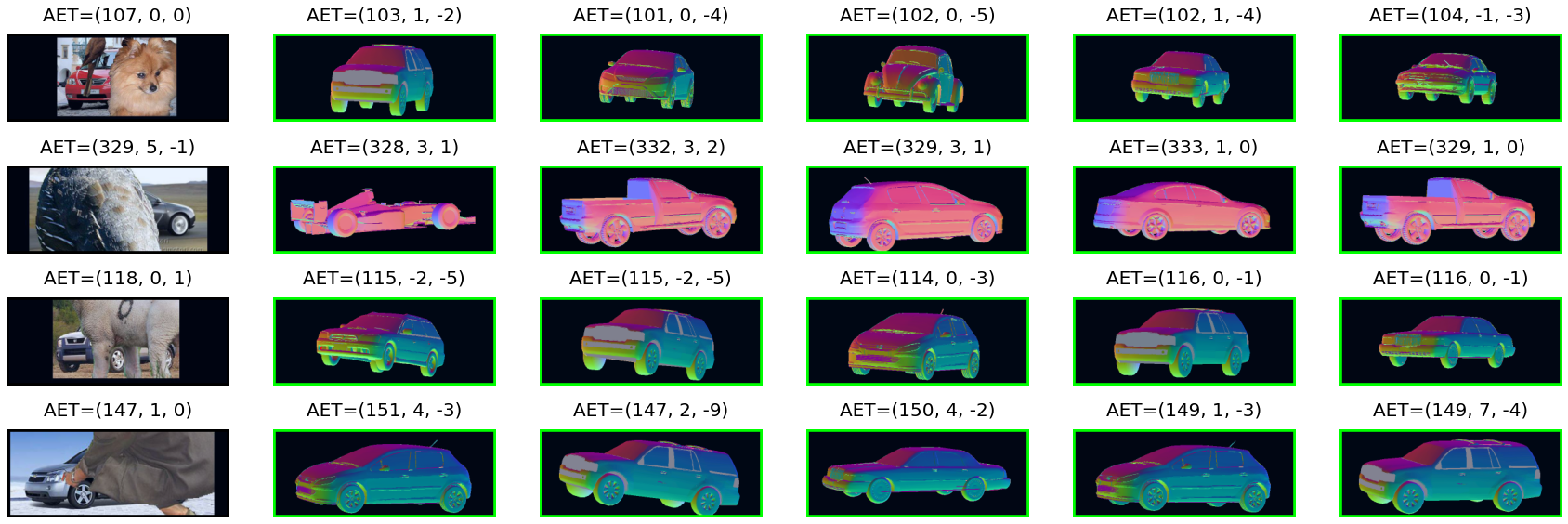}
\caption{L3}
\end{subfigure}
\caption{Retrieval of nearest neighbours for occlusion levels L0-L3 in PASCAL3D Cars.}
\label{fig:supplementary:retrievals:pascal3d}
\end{figure}

\begin{figure}[t]
\begin{subfigure}{\textwidth}
\centering
\includegraphics[width=\textwidth]{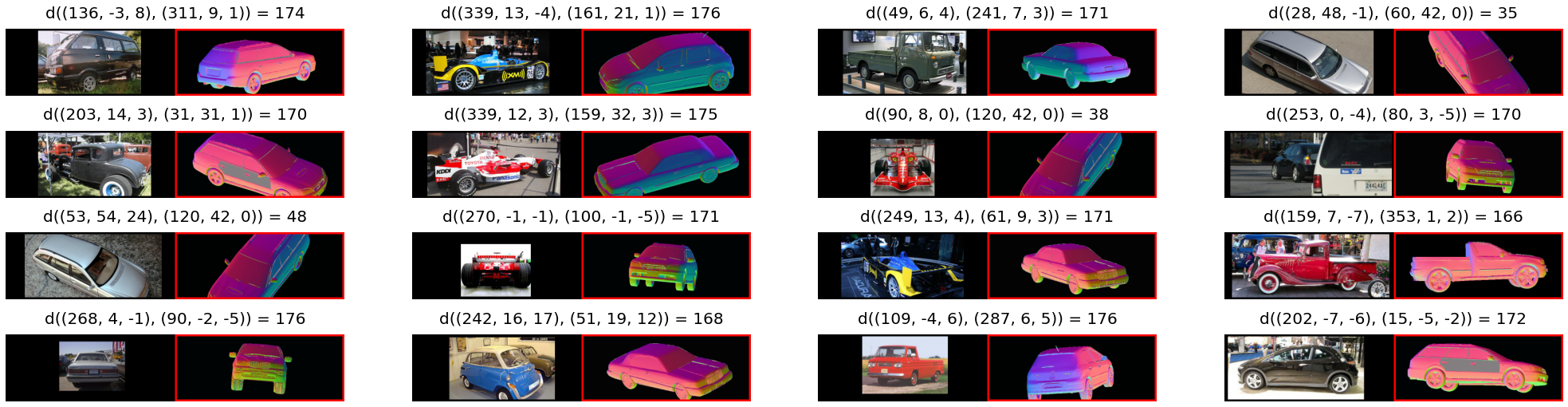}
\caption{L0}
\end{subfigure}\\
\begin{subfigure}{\textwidth}
\centering
\includegraphics[width=\textwidth]{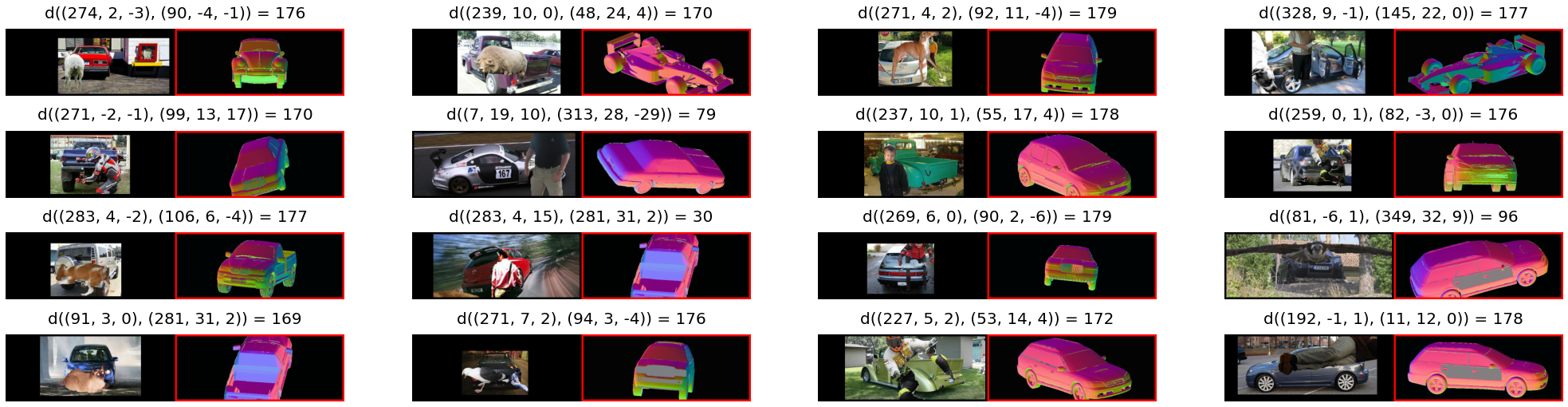}
\caption{L1}
\end{subfigure}\\
\begin{subfigure}{\textwidth}
\centering
\includegraphics[width=\textwidth]{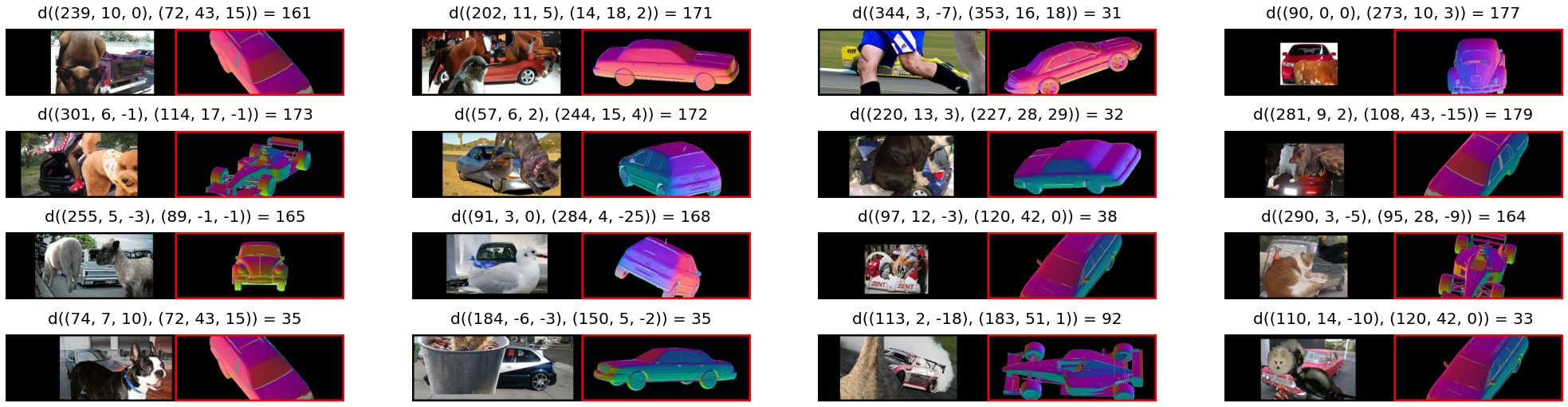}
\caption{L2}
\end{subfigure}
\begin{subfigure}{\textwidth}
\centering
\includegraphics[width=\textwidth]{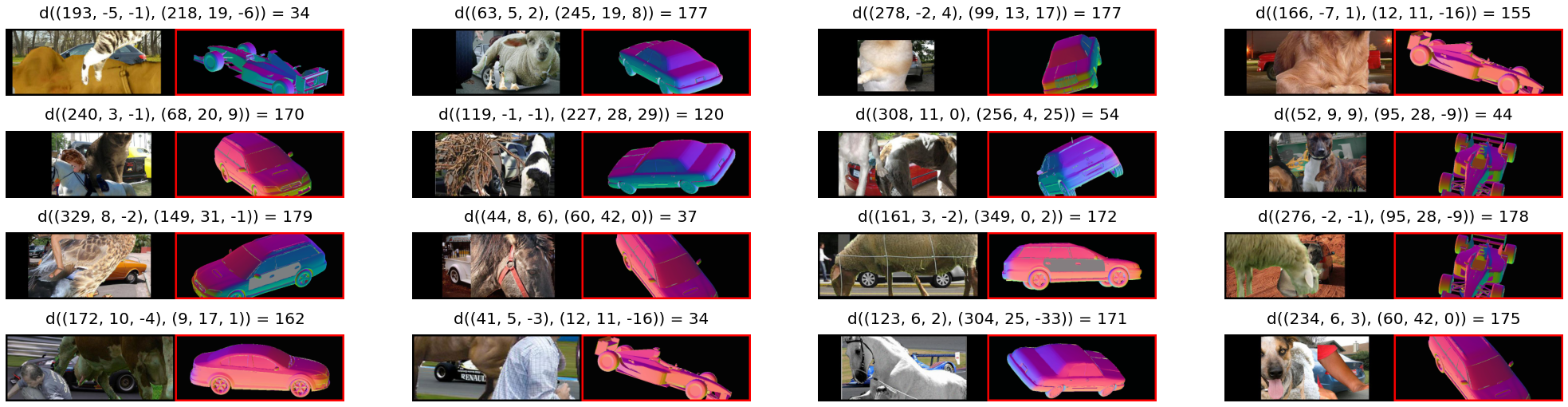}
\caption{L3}
\end{subfigure}
\caption{Failure cases for four different levels of occlusion L0-L3 in PASCAL3D Cars.}
\label{fig:supplementary:retrievals:pascal3d-failures}
\end{figure}

\begin{figure}[t]
\begin{subfigure}{\textwidth}
\centering
\includegraphics[width=\textwidth]{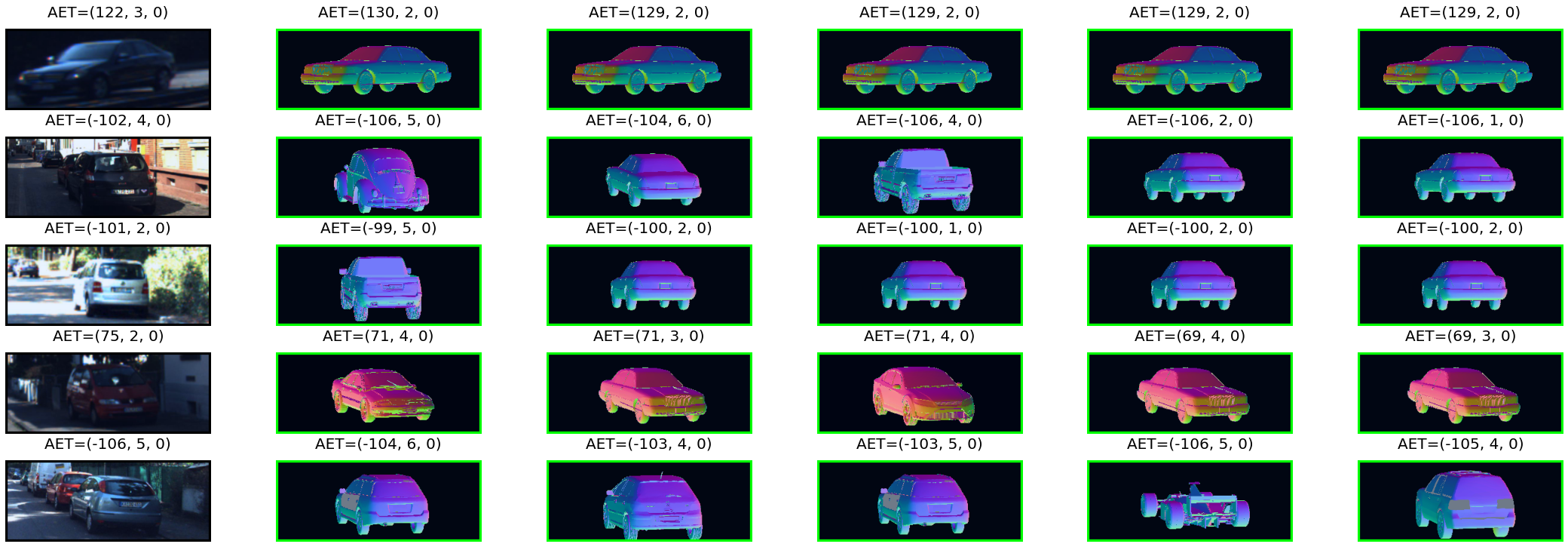}
\caption{Fully Visible Objects}
\end{subfigure}\\[0.4cm]
\begin{subfigure}{\textwidth}
\centering
\includegraphics[width=\textwidth]{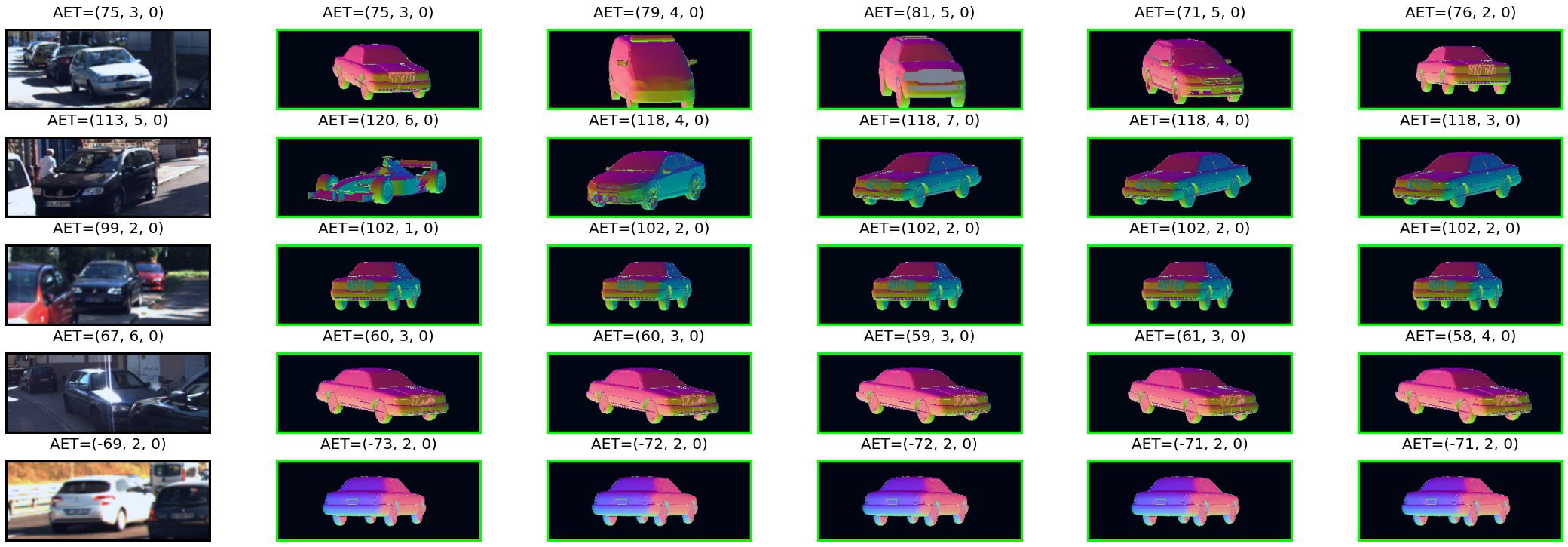}
\caption{Partly Occluded Objects}
\end{subfigure}\\[0.4cm]
\begin{subfigure}{\textwidth}
\centering
\includegraphics[width=\textwidth]{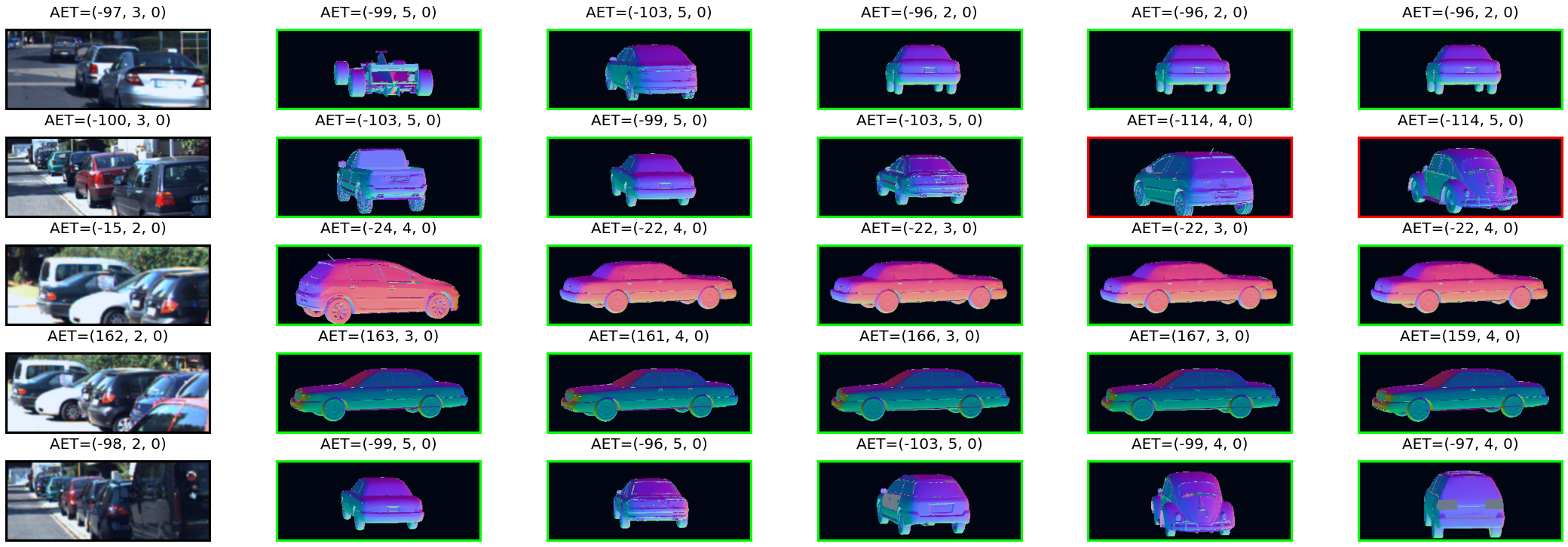}
\caption{Largely Occluded Objects}
\end{subfigure}
\caption{Retrieval of nearest neighbours for three different levels of occlusion in KITTI3D Cars.}
\label{fig:supplementary:retrievals:kitti3d}
\end{figure}

\begin{figure}[t]
\begin{subfigure}{\textwidth}
\centering
\includegraphics[width=\textwidth]{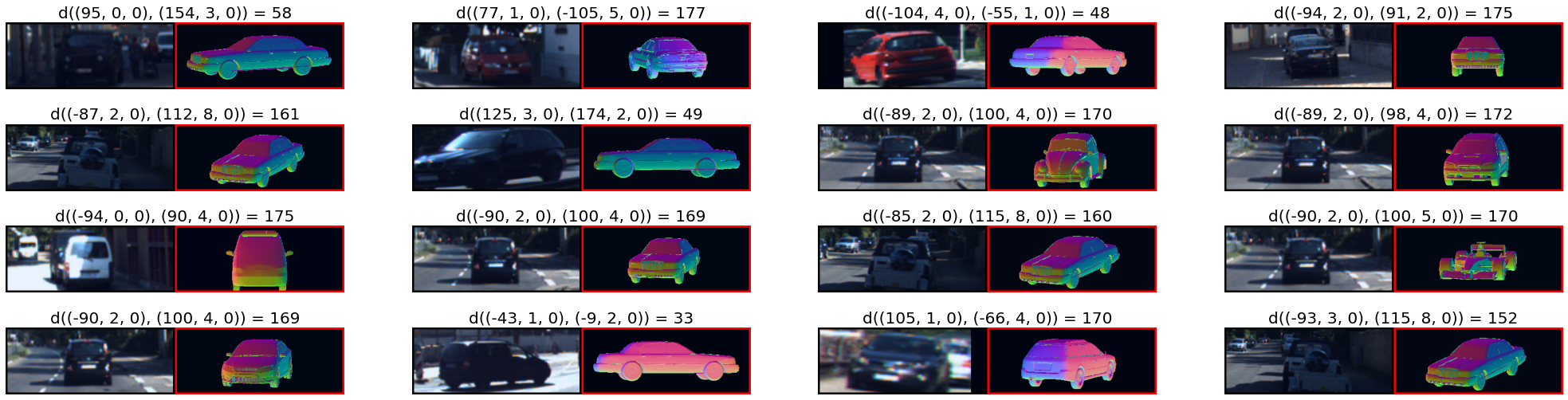}
\caption{Fully Visible}
\end{subfigure}\\[0.4cm]
\begin{subfigure}{\textwidth}
\centering
\includegraphics[width=\textwidth]{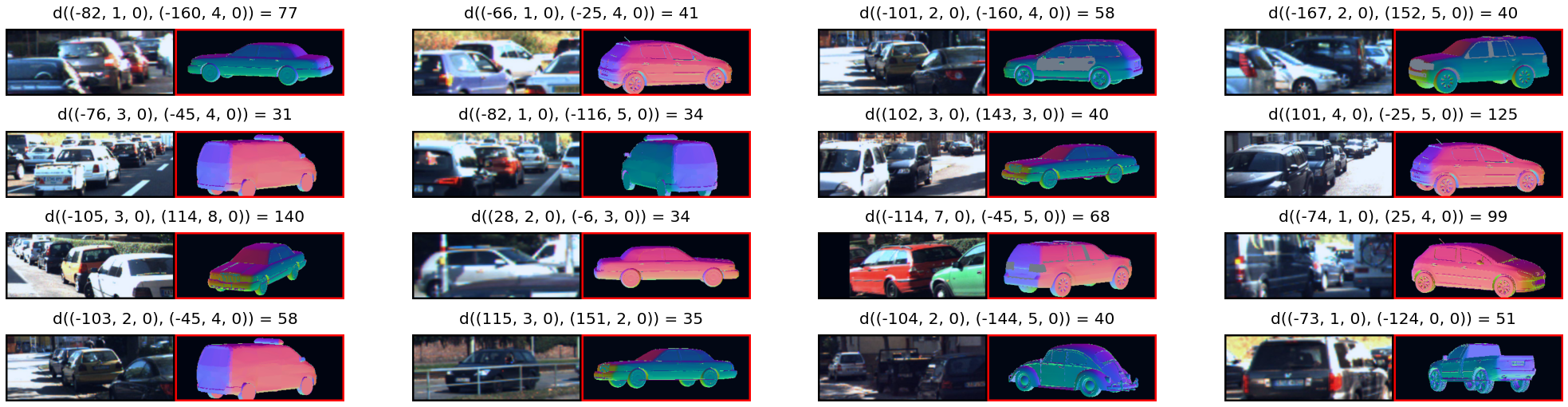}
\caption{Partly Occluded}
\end{subfigure}\\[0.4cm]
\begin{subfigure}{\textwidth}
\centering
\includegraphics[width=\textwidth]{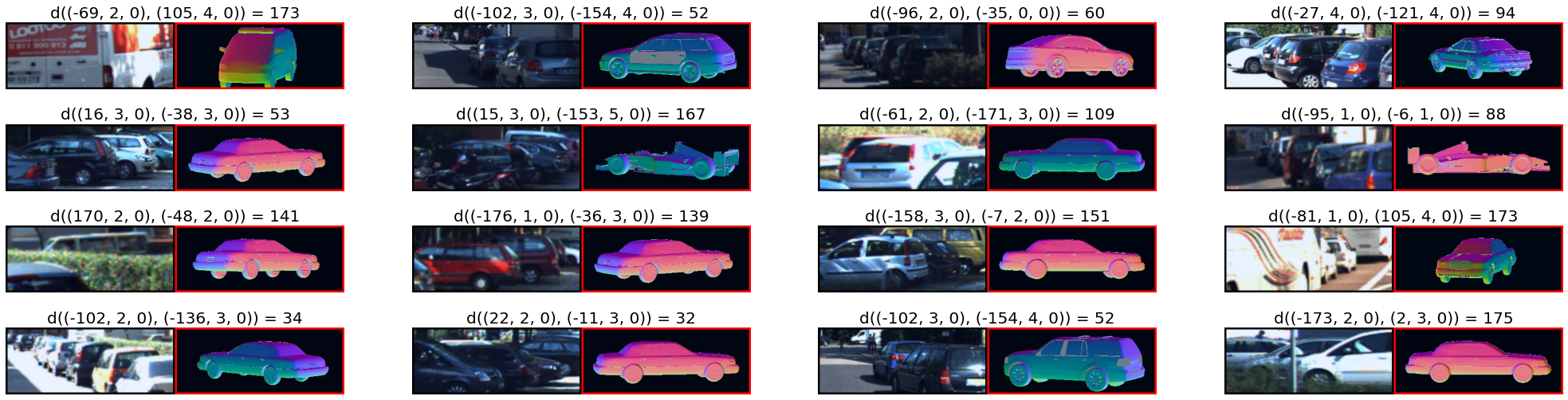}
\caption{Largely Occluded}
\end{subfigure}
\caption{Failure cases for three different levels of occlusion in KITTI3D Cars.}
\label{fig:supplementary:retrievals:kitti3d-failures}
\end{figure}

In this section, we present some additional qualitative results. In particular, Figures~\ref{fig:supplementary:retrievals:pascal3d} and \ref{fig:supplementary:retrievals:pascal3d-failures} present successful pose retrievals and failures cases for PASCSAL3D (L0) and OccludedPASCAL3D (L1-L3), while  Figures~\ref{fig:supplementary:retrievals:kitti3d} and \ref{fig:supplementary:retrievals:kitti3d-failures} present similar cases for KITTI3D in all its occlusion levels, namely fully-visible, partly-occluded, and fully-occluded. We observe that the model has learnt to disregard the specific type of object type and focus more on its pose. In addition, thanks to the occlusion augmentation scheme, the model does not need to see the whole object to estimate its pose, but instead, it can estimate it adequately well even when seeing a small part of the object. On the other hand, observing the failure cases reveals that the model struggles with highly atypical cars, vastly different unseen poses, distinguishing between opposite directions, and large or same-category occlusions.

\end{document}